%% file: main.tex
\documentclass[10pt]{article} 
\usepackage[preprint]{tmlr}

\input{math_commands.tex}

\usepackage{hyperref}
\usepackage{url}
\usepackage{subfigure}
\usepackage{wrapfig}
\usepackage[edges]{forest}
\usepackage{booktabs}
\usepackage[affil-it]{authblk}

\include{settings}

\title{Improving PINNs By Algebraic Inclusion of Boundary and Initial Conditions}

\author{%
      \name Mohan Ren \email mohan.ren@student.manchester.ac.uk
      \AND
      \name Zhihao Fang \email zhihao.fang@manchester.student.ac.uk
      \AND
      \name Keren Li \email keren.li-2@student.manchester.ac.uk
      \AND
      \name Anirbit Mukherjee\thanks{corresponding author} \email anirbit.mukherjee@manchester.ac.uk \\
      \addr Department of Computer Science\\
      The University of Manchester
      }

\def\month{MM}  
\def\year{YYYY} 
\def\openreview{\url{https://openreview.net/forum?id=XXXX}} 

\begin{document}

\maketitle
\begin{abstract} 
``AI for Science'' aims to solve fundamental scientific problems using AI techniques. As most physical phenomena can be described as Partial Differential Equations (PDEs) , approximating their solutions using neural networks has evolved as a central component of scientific-ML. Physics-Informed Neural Networks (PINNs) is the general method that has evolved for this task but its training is well-known to be very unstable. In this work we explore the possibility of changing the model being trained from being just a neural network to being a non-linear transformation of it - one that algebraically includes the boundary/initial conditions. This reduces the number of terms in the loss function than the standard PINN losses. We demonstrate that our modification leads to significant performance gains across a range of benchmark tasks, in various dimensions and without having to tweak the training algorithm. Our conclusions are based on conducting hundreds of experiments, in the fully unsupervised setting, over multiple linear and non-linear PDEs set to exactly solvable scenarios, which lends to a concrete measurement of our performance gains in terms of order(s) of magnitude lower fractional errors being achieved, than by standard PINNs. The code accompanying this manuscript is publicly available at \url{https://github.com/MorganREN/Improving-PINNs-By-Algebraic-Inclusion-of-Boundary-and-Initial-Conditions}
\end{abstract} 






\include{Sections/Intro}


\bibliography{tmlr}
\bibliographystyle{tmlr}

\clearpage  
\appendix

\clearpage 
\tableofcontents 

\include{Computational_Resource_Appendix}

\include{Sections/Poisson}

\include{Sections/Burger}

\include{Sections/KdV}

\end{document}

%% file: math_commands.tex
\usepackage{amsmath,amsfonts,bm}

\def\eqref#1{equation~\ref{#1}}

\def\1{\bm{1}}

\def\ve{{\bm{e}}}

\def\vg{{\bm{g}}}

\def\vu{{\bm{u}}}

\def\vx{{\bm{x}}}
\def\vy{{\bm{y}}}

\DeclareMathAlphabet{\mathsfit}{\encodingdefault}{\sfdefault}{m}{sl}
\SetMathAlphabet{\mathsfit}{bold}{\encodingdefault}{\sfdefault}{bx}{n}

\def\gN{{\mathcal{N}}}

\def\gR{{\mathcal{R}}}

\newcommand{\R}{\mathbb{R}}



%% file: settings.tex
\usepackage{amsmath,amsthm,amssymb,amsxtra,mathtools,bm}
\usepackage{array}
\usepackage{graphicx}
\usepackage{algorithm}
\usepackage{algpseudocode}
\usepackage{physics}
\usepackage{MnSymbol,wasysym}
\usepackage{tikzsymbols}
\usepackage{array,tabularx,multirow}
\usepackage{makecell}
\usepackage{array}

\usepackage{array}
\newcolumntype{P}[1]{>{\centering\arraybackslash}p{#1}}

\usepackage{verbatim}
\usepackage{enumerate}
\usepackage{parskip}

\usepackage{color,soul}
\definecolor{clemson-orange}{RGB}{234,106,32}
\definecolor{highlight-orange}{RGB}{255,150,150}
\definecolor{chicago-maroon}{RGB}{128,0,0}
\definecolor{cincinnati-red}{RGB}{190,0,0}
\definecolor{soft-cyan}{RGB}{68,85,90}
\definecolor{firebrick}{RGB}{178,34,34}
\definecolor{crimson}{RGB}{220,20,60}
\definecolor{cerrulean}{rgb}{0.165,0.322,0.745}
\definecolor{jaam}{rgb}{0.45,0.0,0.45}

\usepackage{hyperref}
\hypersetup{
  colorlinks   = true,    
  urlcolor     = jaam,    
  linkcolor    = firebrick,    
  citecolor    = blue      
}

\usepackage{parskip}

\usepackage{thmtools}
\declaretheoremstyle[
    headfont=\bfseries, 
    bodyfont=\normalfont\itshape, spaceabove=10pt,
    spacebelow=10pt]{mystyle}
\theoremstyle{mystyle}

\newif\ifsolutions \solutionstrue

\def\final{0}
\newcommand{\reviewer}[3]{
  \expandafter\newcommand\csname #1\endcsname[1]{
    \ifthenelse{\equal{\final}{1}} {
      \textcolor{#3}{}
    } {
      \textcolor{#3}{\begin{center} \textbf{#2} ##1 \end{center}}
    }
  }
}

\reviewer{anirbit}{}{jaam}

\def\ve#1{\mathchoice{\mbox{\boldmath$\displaystyle\bf#1$}}
{\mbox{\boldmath$\textstyle\bf#1$}}
{\mbox{\boldmath$\scriptstyle\bf#1$}}
{\mbox{\boldmath$\scriptscriptstyle\bf#1$}}}

\newcommand{\x}{{\ve x}}

\renewcommand{\u}{{\ve u}}

\renewcommand{\t}{{\ve t}}

\def\1{\bm{1}}

\def\ve{{\bm{e}}}

\def\vg{{\bm{g}}}

\def\vu{{\bm{u}}}

\def\vx{{\bm{x}}}
\def\vy{{\bm{y}}}

\DeclareMathAlphabet{\mathsfit}{\encodingdefault}{\sfdefault}{m}{sl}
\SetMathAlphabet{\mathsfit}{bold}{\encodingdefault}{\sfdefault}{bx}{n}

\def\gN{{\mathcal{N}}}

\def\gR{{\mathcal{R}}}



%% file: Sections/Intro.tex
\section{Introduction} 
    The pursuit of first principles of working of nature has led to uncovering various fundamental principles of the natural world, like quantum mechanics, general relativity, fluid dynamics. These principles often apply at different space-time scales and often we do not have an unifying picture across them or even a way to derive one from the other even when they apply to the same system at different scales of description. But somehwat surprisingly, the foundational principles for nearly all natural sciences and engineering disciplines, predominantly manifest as partial differential equations (PDEs). Hence, devising ways to find solutions of PDEs stands as a pivotal subject in the realm of computational science and engineering.

    Throughout the history of PDEs, the various techniques that have been developed to solve them range from the plethora of analytic methods to various discretization-based approaches. Significant progress has been made in providing analytic and numerical solutions to various specific PDEs - but general methods do not exist. A major challenge is to approximately solve PDEs without facing the curse of dimensionality \citep{E_2021, han2018solving} which poses significant difficulties for even very structured classes of PDEs \citep{Beck_2021, berner2020numerically}. Over the past years, deep learning has emerged as a powerful tool across various domains - including for solving PDEs. Some of the earliest attempts\citep{Lagaris_1998} at leveraging neural networks for PDE directly inspired this current work. Since then, there has been a growing trend in utilizing deep learning techniques for solving complex scientific problems, giving rise to the field now being known as ``AI for Science''. The potential that this direction holds can be glimpsed from the rapid developments that are underway in the particular area of using AI techniques in advancing climate science \citep{2021ComEE...2..159G, irrgang2021, Yuval_2020}.

    
    Currently, there are emerging a plethora of different ways \citep{karniadakis2021piml} in which one can write a loss function involving a partial differential equation (or its available approximate solutions) and neural net(s)  such that the net(s) obtained at the end of training can be used to infer approximate solutions to the intended PDEs. These data-driven methods of solving the PDEs can broadly be classified into two kinds, {\bf (1)} ones which train a single neural net to solve a specific PDE and {\bf (2)} operator methods -- which train multiple nets in tandem to be able to solve a family of PDEs in one shot \citep{Lu_2021, LU2022114778, wang2021pdeOnet}. The operator methods are particularly of interest when the underlying physics is not known. While there have evolved many kinds of methods \citep{kaiser2021datadriven, erichson2019physicsinformed, Wandel_2021, li2022learning, salvi2022neural} that can solve the PDEs in a deep learning way, the Physics Informed Neural Networks (PINNs) from \citep{raissi2019physics} have emerged as an umbrella framework. PINNs have been shown to be good for simulating the Navier-Stokes PDE \citep{ARTHURS2021110364, wang2020physicsinformed, Eivazi_2022}, inviscid, incompressible fluid flows modeled by the Euler PDE \citep{wang2022asymptotic} and shallow water waves as modeled by the Korteweg-De Vries PDEs \citep{hu2022XPINNs} - which shall be a focal point of this work.

    The principle of PINNs is to employ neural networks to approximate solutions of PDEs by minimizing a loss function that has multiple terms separately penalizing the neural net being trained for failing to satisfy the initial and boundary conditions as well as the partial differential operator. Each of these terms would be on a different domain. Thus, PINNs minimize a multi-distribution loss which is a departure from various conventional ML implementations. The prototype idea of PINNs can be found in meshless numerical methods \citep{lawal2022prototype, PANG2015280kansa, broomhead1988RBF}. 

    A theoretical understanding of the PINN's generalization properties has only begun to be explored \citep{kutyniok2020theoretical, gühring2019error, geist2020numerical}. Work in \citep{Mishra_PINN_generalization_error} has  provided certain preliminary error bounds but there are significant gaps between the loss function used therein (which use quadrature weighting terms) and standard PINNs - like what we explore in this work. Recent works by some of the current authors \citep{Kumar_2024} has provided ways to establish relationships between the PINN loss and the error of approximating the PDE solution, focussing on Burgers' PDE near finite-time blow-up conditions.

    We note, that despite the efforts, as outlined above, it remains unclear,  why minimization of the standard PINN loss function would lead to neural nets which are a good approximation of the true PDE solution. 

    The growing interest in this technology has also brought to light the challenges that PINNs can face in solving even some basic PDEs, as highlighted by \citep{krishnapriyan2021characterizing, wang2022pinns}. Attempts have been made to address these limitations by exploring alternative gradient descent approaches \citep{wang2020understanding} or by patching together multiple solutions \citep{hu2022XPINNs} etc. 
    
    Except for a few instances (discussed in Section \ref{section_disc}), most of the current improvement strategies focus on tuning the regularization terms in the loss or the training algorithm. In contrast, \emph{we attempt an expansive study on modifying the PINN loss function to make their training better - in the fully unsupervised setup.}

    In this study, we investigate the potential of employing as the predictive model a non-linear transformation of the neural network output. This transformation is designed to incorporate as much prior knowledge as possible. Our primary strategy is to leverage the simplicity of prevalent boundary and initial conditions, enabling their algebraic integration into the models under training. We argue that this modification of the predictor being trained yields performance gains on widely varying PDEs such as Poisson PDE and Burgers' PDE in different dimensions, and the Korteweg-de Vries (KdV) PDE. 

\paragraph{Review of the Deep Ritz Method} The Deep Ritz Method (DRM) is a computational technique that combines deep learning and the Ritz variational principle to solve elliptic PDEs effectively \citep{e2017deep}. This method relies on the parameterization of neural networks to implement a search to find the variational minima and it can be an alternative to PINNs when a variational formulation of the PDE is available. 


Works like \citep{lu2021priori} have delved into an analysis of the generalization bounds for the DRM with a focus on the Poisson equation and the static Schrödinger equation. Complementing this \citep{muller2022error} have made advances in estimating the error of DRM, \citep{dondl2022uniform} have presented a study of the convergence guarantees for the empirical risk of DRM and \citep{wu2021understanding} have explored the landscapes of the different loss functions within the DRM framework. 

In the context of the Poisson PDE, we will show how our modifications to PINNs can supersede DRM too.

\paragraph{Motivations for Studying the Poisson PDE} \label{motivation_poisson}


For $\Omega = {(0, 1)}^d$ and a appropriate choice of a real valued function $f$ on $\Omega$, we consider the Poisson PDE specified as,
\begin{gather}\label{eqn: poisson equation 1}
\nonumber - \Delta u = f(\vx ) ~\forall \vx \in \Omega, ~u(\vx ) = 0 ~\forall \vx \in \partial \Omega.
\end{gather}
Specifically for the Poisson PDE, moving away from conventional PINN and DRM, works like \citep{wu2021understanding} have already initiated ways to improve deep-learning performance for this PDE by incorporating the boundary conditions into the model's architecture. This approach circumvents complications associated with boundary sampling in high-dimensional problems.


A goal of our studies would be to deeply investigate the relative performance of the above methods in solving the Poisson PDE and demonstrate new ways of setting up the loss function for it which would outperform all existing methods. Poisson PDE emerges as a lucrative test case as it allows for both variational energy methods and residual methods to be applied. Furthermore, the Poisson equation is inherently defined in every dimension, rendering it an excellent benchmark for investigating the trade-offs because of the multi-dimensionality.


We recall that the incompressible Navier-Stokes equations are given by,
$\frac{\partial \mathbf{u}}{\partial t} + (\mathbf{u} \cdot \nabla) \mathbf{u} = -\frac{1}{\rho} \nabla p + \nu \nabla^2 \mathbf{u} + \mathbf{f}
~\& ~
\nabla \cdot \mathbf{u} = 0$, 
where $\mathbf{u}$ is the fluid velocity field, $t$ is time, $p$ is pressure, $\rho$ is fluid density, $\nu$ is the kinematic viscosity, and $\mathbf{f}$ represents external body forces. The first equation represents the momentum equation, while the second enforces the incompressibility constraint (the continuity equation). To satisfy the incompressibility constraint, the pressure field must be determined. This is achieved by taking the divergence of the momentum equation and rearranging terms to obtain the Poisson equation for the pressure field, $\nabla^2 p = \nabla \cdot (\rho (\frac{\partial \mathbf{u}}{\partial t} + (\mathbf{u} \cdot \nabla) \mathbf{u}))$. Here we have assumed the viscosity to be a non-dynamical variable. Once the pressure field is known, it can be incorporated back into the momentum equation to update the velocity field \citep{pletcher2012computational}. Navier-Stokes PDE are a common model for modeling physically consistent fluid behavior \citep{Schneiderbauer_2014} and the above argument illustrates how understanding the Poisson PDE is a fundamental part of that scheme. 

\paragraph{Motivations for Studying the Burgers' PDE} \label{motivation_burger} In recent years, the Burgers' PDE has frequently been used as a benchmark problem for testing machine learning algorithms, particularly physics-informed neural networks (PINNs) \citep{raissi2019physics}, \citep{blechschmidt2021three}. In light of the discussion above we can see Burgers' PDE as being the zero-pressure (``dust'') special case of Navier-Stokes PDE. Using the same variables as above, the generic dimensional Burgers' PDE is given as,

\begin{equation}
\label{BurgerPDE}
\begin{array}{c}
\partial_t \vu + (\vu \cdot \nabla)\vu = \nu \nabla^{2} \vu,  (\vx,t)\in [0,1]^d \times [0,1]\\
\vu(\vx, 0)=F(\vx), \vx \in [0,1]^{d}\\
\vu(\vx, t) = \vg(\vx,t) , \vx \in \partial ([0,1]^{d})
\end{array}
\end{equation}

In above, $F$ and $\vg$ are appropriately chosen real and $\R^d$ valued functions on the domains $[0,1]^d$ and its boundary, respectively. 
When expanded out in coordinates
\footnote{
\begin{equation}
\label{Burger}
\partial_t
\begin{bmatrix}
 u_1\\u_2\\.\\.\\.\\u_d\\
\end{bmatrix}
=
\begin{bmatrix}
 \partial_t u_1\\ \partial_t u_2\\.\\.\\.\\\partial_t u_d\\
\end{bmatrix}
\ , \
(\sum_{i=1}^{d} u_i\partial_{x_i})
\begin{bmatrix}
 u_1\\u_2\\.\\.\\.\\u_d\\
\end{bmatrix}
=
\begin{bmatrix}
 \sum_{i=1}^{d} u_i\partial_{x_i} u_1\\ \ \sum_{i=1}^{d} u_i\partial_{x_i} u_2\\.\\.\\.\\ \sum_{i=1}^{d} u_i\partial_{x_i} u_d\\
\end{bmatrix}
\ , \
\nu
\begin{bmatrix}
 \nabla^{2}u_1\\ \nabla^{2}u_2\\.\\.\\.\\ \nabla^{2}u_d\\
\end{bmatrix}
=
\nu
\begin{bmatrix}
 \sum_{i=1}^{d} \partial_{x_i}^2 u_1\\ \ \sum_{i=1}^{d} \partial_{x_i}^2 u_2\\.\\.\\.\\ \sum_{i=1}^{d} \partial_{x_i}^2 u_d\\
\end{bmatrix}
\end{equation}
}
the above represents a PDE for $d$ coupled scalar fields $u_1$,$u_2$......$u_d$. Note that we have not imposed the divergenceless-ness condition in equation \ref{BurgerPDE} and this allows us to have solutions that blow-up in finite-time despite starting from smooth initial conditions - which constitute a hard test case for deep-learning methods as we consider here. Recently theoretical progress has been made to understand the guarantees for standard PINN methods to solve the above,\citep{Kumar_2024}. But in contrast, here we shall extensively modify the loss function to get state-of-the-art performances. 

\paragraph{Motivations for Studying the KdV PDE} \label{motivation_KdV}

The Korteweg–de Vries (KdV) equation is among the most significant non-linear PDEs in one spatial dimension. It's a very educational benchmark, since despite KdV being a third-order PDE, it allows for exact solutions that are also interpretable as models of shallow water waves that maintain their shape over time. KdV's history dates back to Scott Russell's experiments in 1834 \citep{kdv_origin}. 


It was only much later that numerical investigations led to the discovery \citep{solitons} that its solutions decompose into a collection of ``solitons'' at large times. Solitons can be imagined as waves that maintain their shape and velocity as they propagate through a medium \citep{LEWIS2022329}. Specifically, a bi-variate real-valued function $u$ would be said to satisfy the KdV PDE if it satisfies the following equations, 


\begin{equation}
\label{KdV}
\begin{aligned}
    \pdv{u}{t} +\pdv[3]{u}{x} + 6\,u \,\pdv{u}{x} =0, (x, t) \in \Omega \times T, \\
    u(x, 0) = u_0(x), x\in \Omega, \\
    u(x, t) = g(x, t), (x, t) \in \partial \Omega \times T,  \\
    \Omega, T \subset {\R}\ {\rm both\ bounded}.
\end{aligned}
\end{equation}

where $u_0(x)$ and $g(x, t)$ are the appropriate initial and boundary conditions, respectively. In this research, a particular parametric class of initial conditions \citep{soliton_cambridge} of the above shall be considered,
\begin{equation}\label{initial_condition}
    u(x, 0) = N(N+1){\rm sech}^2(x),
\end{equation}
where $N \in \{1,2,3,\ldots\}$ would be said to be the ``soliton number'' of the solution that evolves from above. Thus, the corresponding $1-$soliton solution of the KdV PDE becomes,
\begin{equation}\label{1_soliton_sol}
    \begin{aligned}
        u(x, t) = 2 \sech^2(x-4t),
    \end{aligned}
\end{equation}
while multiple soliton solutions can all be written as, $ u(x, t) = 2 \partial_x^2 (\log f)$ for appropriate $f$. For example, the $2-$soliton solution arises from, 
\begin{equation} \label{2_soliton_sol}
\begin{aligned}
    & f=1+e^{\eta_1}+e^{\eta_2} + e^{\eta_1 + \eta_2 + A_{12}}, ~\eta_i = k_ix - k_i^3t + \eta_i^{0}, ~e^{A_{ij}} = (\frac{k_i-k_j}{k_i+k_j})^2,
\end{aligned}
\end{equation}
And the $3-$soliton solution arises from,
\begin{equation} \label{3_soliton_sol}
\begin{aligned}
    f &=1+e^{\eta_1}+e^{\eta_2}+e^{\eta_3} + e^{\eta_1 + \eta_2 + A_{12}}+ e^{\eta_2 + \eta_3 + A_{23}}+ e^{\eta_3 + \eta_1 + A_{31}}+ e^{\eta_1 + \eta_2+ \eta_3 + A_{12}+ A_{23}+ A_{31}} \\
    \eta_i &= k_ix - k_i^3t + \eta_i^{0}, ~e^{A_{ij}} = (\frac{k_i-k_j}{k_i+k_j})^2,
\end{aligned}
\end{equation}
By using the fixed initial condition, equation \ref{initial_condition}, the parameters in the corresponding solutions can be calculated. For the $2-$soliton solution in equation \ref{2_soliton_sol}, thus we determine that $k_1=-2, k_2=4, \eta_1^0=\ln{\frac{1}{3}}, \eta_2^0=\ln{\frac{1}{3}}$. Similarly for the $3-$soliton solution in equation \ref{3_soliton_sol}, we determine that $k_1=2, k_2=4, k_3=-6, \eta_1^0=\ln{\frac{3}{2}}, \eta_2^0=\ln{\frac{3}{5}}, \eta_3^0=\ln{\frac{1}{10}}$.


\paragraph{Notations} Given a point in $p$ dimensions, $\vx \in \R^p$, its image in $q-$dimensions via a depth $m$ standard feed forward neural net, $\gN : \mathbb R^p \xrightarrow{} \mathbb R^q$ is to be evaluated as, 
\begin{equation}\label{Net}
    {\gN}(\vx) = {\rm A}_m(\sigma({\rm A}_{m-1}(...(\sigma ({\rm A}_1(\vx)))...))),
\end{equation}
where, ${\rm A}_1 : \mathbb{R}^p \xrightarrow{} \mathbb{R}^n,
{\rm A}_2 : \mathbb{R}^n \xrightarrow{} \mathbb{R}^n, ...,  {\rm A}_{m-1}: \mathbb{R}^n \xrightarrow{} \mathbb{R}^n,
{\rm A}_m : \mathbb{R}^n \xrightarrow{} \mathbb{R}^q$ are affine transforms and $n$ is a fixed inner-dimension and hence the size of every layer of gates with activation $\sigma$ -- a univariate non-linear function acting coordinate wise. One can read off that this neural net has $(p \times n + n) + (m-2) \times (n^2 + n) + (n \times q + q)$ parameters - which we shall often denote collectively as $\theta$. When needed to emphasize the dependency of the net on the parameters $\gN(\vx)$ will be replaced by $\gN(\vx;\theta)$.


To compare a trained model $g$ with any exact solution of a PDE $u$, we would measure the fractional error as $\frac{\lVert g - u \lVert _2^2}{\lVert u \lVert _2^2}$ where the norms are to be understood in the $L_2$ space. 

\paragraph{Organization}  Our results are summarized in Section \ref{summary_section}. Section \ref{Best_Model_with_Boundary} presents our study on high-dimensional Poisson PDEs, and one and two-dimensional Burgers' PDEs, as examples of PDE solution targets where the boundary-inclusive model yields optimal performance. Section \ref{Best_Model_with_Initial} presents a study on three-dimensional Burgers' PDEs and the KdV equation where models incorporating initial conditions provide the most accurate approximations. We conclude in Section \ref{section_disc} with a discussion of possible future directions.  
~\\ \\
\qquad Appendix \ref{appendix_computational_resource} documents the computational resources utilized. Appendix \ref{Sec_Poisson_exp} details all the variants of PINN and DRM strategies tried on the Poisson PDE. Appendix \ref{Sec_Burger_exp} presents all the details of the PINN variants tried on the Burgers' PDE, inclusive of both viscid and inviscid solutions. Appendix \ref{Sec_KdV_exp} includes all the details on the KdV experiments organized as one subsection for each soliton type considered. 

\section{Summary of Results} \label{summary_section}


With respect to the PDEs reviewed above, we discuss our findings in two segments : ones among those for which the optimal models found correspond to algebraically including the boundary conditions or the initial conditions into the predictor. 
\subsection{PDE Solution Targets Where the Best Performance is of Models with Boundary Conditions Included} \label{Best_Model_with_Boundary}
\paragraph{Poisson PDE}
For both, energy \citep{e2017deep} and residual minimization methods, we incorporated the boundary condition via two distinct methodologies : via encoding it within the model, and via adding penalty terms to the loss function. Furthermore, in the former approach we also incorporated a novel dimension dependent multiplicative regularizer. Consequently, we have trained a total of six settings in this experiment.

\begin{figure}[htb!]
  \centering
  \subfigure{
    \includegraphics[width=2.6in, height = 1.8in]{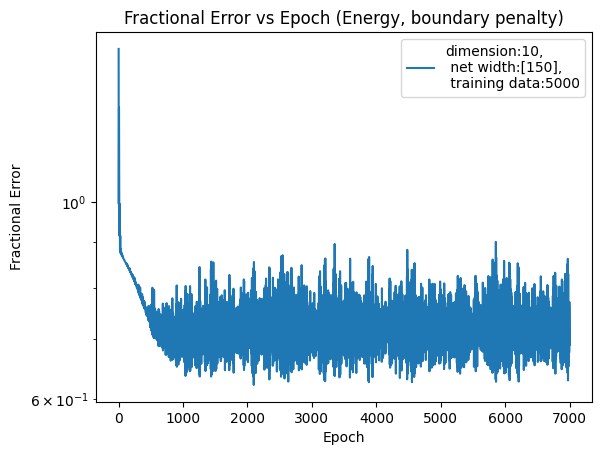}
    \fbox{\includegraphics[width=2.6in, height = 1.8in]{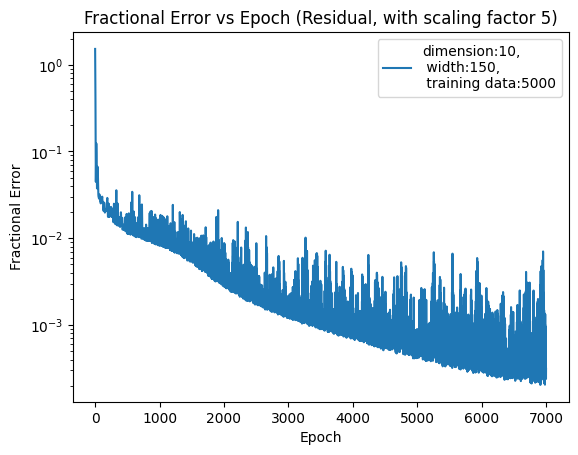}}
    }
  \caption{Fractional Error of Different Loss Functions with the Best Performance on the Right (\(d= 10\)).
  {\it Left}  : $\mathcal{\hat{\gR}}_{\rm energy, loss-with-boundary-penalty}$ (\ref{compare_loss}) i.e the ``Deep Ritz Method''.  {\it Right}:  $\mathcal{\hat{\gR}}_{\rm residual, boundary-included, \lambda = 5}$ (\ref{best_empirical_risk}).}
  \label{fig:Poisson_best_results}
\end{figure}

In Figure \ref{fig:Poisson_best_results}, in its two panels we present a comparative analysis of change of fractional error with epochs between DRM with a boundary penalty -- and the best performance we have identified, as shown on the right. The combinations we discovered demonstrated lower errors than DRM, across all dimensions. Nevertheless, in the setup of DRM,  the convergence speed was superior across all configurations. 

Explicitly, we recall the boundary penalized DRM training loss  \citep{e2017deep}, as the following equation in terms of a neural net $\gN$,
\begin{equation}\label{compare_loss}
\mathcal{\hat{\gR}}_{\rm energy, loss-with-boundary-penalty}(\gN) = \frac{1}{N} \sum_{i=1}^{N} \left[ \frac{1}{2} \left\vert \nabla  \gN(\vx_i;\theta) \right\vert^2 - f(\vx_i)  \gN(\vx_i;\theta) \right] + \frac{\beta}{M} \sum_{j=1}^{M} \gN(\vy_j; \theta)^2.
\end{equation}

And for all settings of the Poisson PDE, equation \ref{eqn: poisson equation 1}, that we tested, the best performance was always the residual minimization with a boundary condition included model and having a multiplicative regularizer,
\begin{equation}\label{best_poisson_model}
    {\rm model} \coloneqq \hat{u}(\vx;\theta) \coloneqq \lambda^d \cdot \left (\prod_{i = 1}^{d} (1 - x_i) x_i \right ) \cdot \gN(\vx;\theta)
\end{equation}
which was being trained on the following residual empirical risk, 
\begin{equation}\label{best_empirical_risk}
    \mathcal{\hat{\gR}}_{\rm residual, boundary-included, \lambda = 5}(\hat{u}) = \frac{1}{N} \sum_{i=1}^{N} \left(-\Delta \hat{u}({\vx}_i; \theta) - f({\vx}_i)\right)^2, \quad \text{where}\ \vx_i \in [0,1]^d
\end{equation}

For the above model in dimension $1$, the minimum fractional error found was $3.63 \times 10^{-6}$,   $1.02 \times 10^{-5}$ in dimension $2$, $1.76 \times 10^{-5}$ in dimension $3$ and $4.00 \times 10^{-4}$ in dimension $10$.\\

 Full experiment can be found in Appendix \ref{Sec_Poisson_exp}, where section \ref{loss_settings} goes into detail about the various settings. And the summary of our finding can be found in section \ref{sum_poisson}.

\paragraph{Burgers' PDE}
In the context of both one-dimensional and two-dimensional Burgers' PDE, all three models, the standard PINN model, an initial-condition-included model and a boundary-condition-included model (which we will describe in further detailed below) were all trained at $4$ different sizes of neural networks spanning various scales of under and overparameterization. This stands as testimony to the robustness of our conclusions. 

In one spatial dimension, we test against an exact solution of the Burgers' PDE, equation \ref{BurgerPDE}, that corresponds to initial condition $F(x)=x$ and boundary condition $g(x,t) = \frac{x}{t+1}, x \in \{0, 1\}$. We find that irrespective of the size of the model the best performance comes from a boundary-condition-included model of the fluid velocity,
\begin{equation}\label{best_Burger1_Model}
 {\rm model_{boundary-included}}({x,t}) \coloneqq \hat{u}_{b} \coloneqq {\gN}(x, t) \cdot x \cdot (1-x)  +  (1-x) \cdot {g}_{x,0}(t) + x{\cdot} {g}_{x,1}(t)
\end{equation}

where ${\gN(x, t)}$ is a neural net of the form as in equation \ref{Net}. The corresponding population risk function (that is discretized to a training loss) is as follows,
\[ \operatorname{{\gR} _{\rm boundary-included}}({u}_{b}) \coloneqq \left\|\frac{\partial {u}_{b}}{\partial t}+{u}_{b} \frac{\partial {u}_{b}}{\partial x}\right\|_{[0,1] \times[0,1], \nu_{1}}^{2}+\|{u}_{b}-F(x)\|_{t=0,[0,1], \nu_{2}}^{2}. \]

where $\nu_{1}$, $\nu_{2}$ are the uniform measures from which the collocation points are sampled in the respective domains. Further, for Burgers' PDE, equation \ref{BurgerPDE}, for $2$ spatial dimensions, we let $u$ and $v$ be defined to represent $u_1$ and $u_2$ respectively. We let ${g}_{x,0}(y, t)$ and ${g}_{x,1}(y, t)$ be the boundary conditions for $u$ at $x=0$ and $x=1$, and ${g}_{y,0}(x, t)$ and ${g}_{y,1}(x, t)$ be the boundary conditions for $v$ at $y=0$ and $y=1$ and $u_{0}$ and $v_{0}$ be the corresponding initial conditions. 

Thus the specific (and particularly challenging) exactly solvable $2-$dimensional Burgers PDE target we consider is, 
\begin{equation}
\label{burger2.inviscid}
\left\{\begin{array}{l}
u_{t}+u u_{x}+v u_{y}=0 \\
v_{t}+u v_{x}+v v_{y}=0
\end{array}\right.,
\left\{\begin{array}{l}
u_{0}= x+y \\
v_{0}= x-y
\end{array}\right.,
\left\{\begin{array}{l}
{g}_{x,0}(y, t)=\frac{y}{1-2\cdot t^2}\\
{g}_{x,1}(y, t)=\frac{1-y-2\cdot t}{1-2\cdot t^2}\\
{g}_{y,0}(x, t)=\frac{x}{1-2\cdot t^2}\\
{g}_{y,1}(x, t)=\frac{1-x-2\cdot t}{1-2\cdot t^2}
\end{array}\right.
\end{equation}
The computational domain $\Omega$ for the above solution is $\Omega=\{(x, y): 0 \leq x \leq 1, 0 \leq y \leq 1\}$. This two-dimensional inviscid Burgers' PDE is known to possess an exact solution characterized by a finite-time blow-up, a phenomenon whereby the solution becomes unbounded within a finite time despite starting from smooth initial conditions. This interesting attribute renders it a formidable test case for numerical methodologies as they have to not only be accurate but also robust enough to manage the steep gradients that emerge in the proximity to the blow-up time.


For solving the above $2-$dimensional Burgers' PDE, the model with the best performance turns out to be the following boundary-condition-included model,
\begin{align*} 
\hat{u}_{b} &\coloneqq {\gN_{u}}(x,y, t) \cdot x \cdot (1-x)  +  (1-x) \cdot {g}_{x,0}(y,t) + x \cdot {g}_{x,1}(y,t)\\
\hat{v}_{b} &\coloneqq {\gN_{v}}(x,y, t) \cdot y \cdot (1-y)  +  (1-y) \cdot {g}_{y,0}(x,t) + y \cdot {g}_{y,1}(x,t)
\end{align*}

where ${\gN(x, t)} : \R^3 \rightarrow \R^2$ is a neural net being trained whose two output coordinates have been named $\gN_u$ and $\gN_v$ above. Correspondingly, we define the PDE population risks, ${\gR}_{1}$  and $ {\gR}_{2}$ as follows, 

\begin{equation}
\label{Burger2.loss1}
\begin{aligned}
{\gR}_{1} &= \left\|\frac{\partial {\hat u}_{b}}{\partial t}+{\hat u}_{b} \frac{\partial {\hat u}_{b}}{\partial x} + {\hat v}_{b} \frac{\partial {\hat u}_{b}}{\partial y}\right\|_{[0,1]^2 \times[0,1], \nu_{1}}^{2} ~~\& ~~{\gR}_{2} &= \left\|\frac{\partial {\hat v}_{b}}{\partial t}+{\hat u}_{b} \frac{\partial {\hat v}_{b}}{\partial x} + {\hat v}_{b} \frac{\partial {\hat v}_{b}}{\partial y}\right\|_{[0,1]^2 \times[0,1], \nu_{1}}^{2}
\end{aligned}
\end{equation}
And residuals ${\gR}_{5} $ and $ {\gR}_{6} $ are defined as measuring the risk of non-satisfaction of the initial conditions,

\begin{equation}
\label{Burger2.loss3}
\begin{aligned}
{\gR} _{5} &= \left\| {\hat u}_{b} - u_{0}\right\|_{[0,1]^2 ,t=0, \nu_{3}}^{2} ~~\& ~~{\gR} _{6} &= \left\|{\hat v}_{b} - v_{0}\right\|_{[0,1]^2 ,t=0, \nu_{3}}^{2}
\end{aligned}
\end{equation}

As earlier, the $\nu$s above continue to denote uniform measures on the respective domains. Thus, combining the risks above, we obtain the full population risk function (whose discretization we train on) as follows that leads to the best performing model for Burgers' PDE in two dimensions, 
\[{\gR} _{\rm boundary-included}=\left({\gR}_{1}({\hat u}_{b})+{\gR}_{2}({\hat v}_{b})\right)+\left({\gR}_{5}({\hat u}_{b})+{\gR}_{6}({\hat v}_{b})\right) \]
For this $1-$dimensional inviscid Burgers' PDE considered above, the minimal fractional error we obtain is $6.42 \times 10^{-14}$. In the case of a $1-$dimensional viscid Burgers' PDE (described in Appendix \ref{Sec_1burger}), this value is $7.14 \times 10^{-13}$. The minimum fractional error for the $2-$dimensional inviscid Burgers' PDE is $5.95 \times 10^{-10}$. 

\begin{figure}[htbp!]
    \centering  %
    \subfigure[{Comparison Between Predicted and True Solutions for $t =0.6$}   ]{
        \centering        \includegraphics[width=\textwidth]{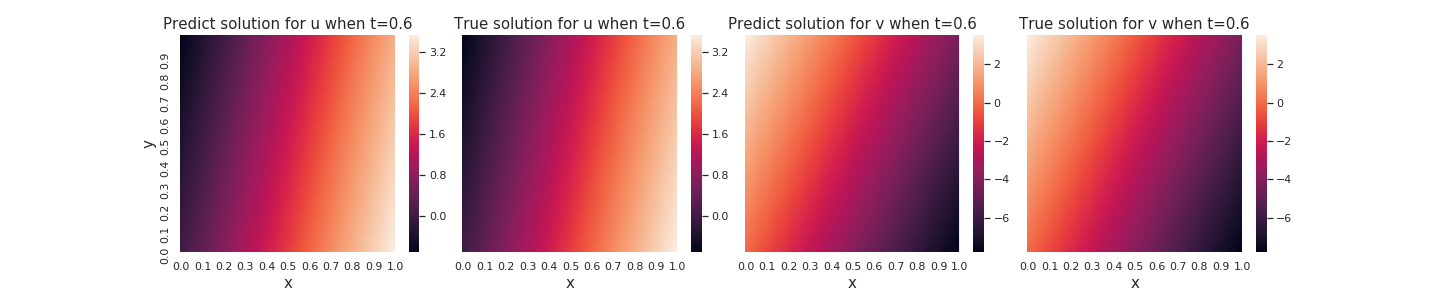}}
    \label{Burger2.heatmap.sub.1}
    \subfigure[{Comparison Between Predicted and True Solutions for $t = 0.705$}  ]{
        \label{Burger2.heatmap.sub.2}
        \includegraphics[width=\textwidth]{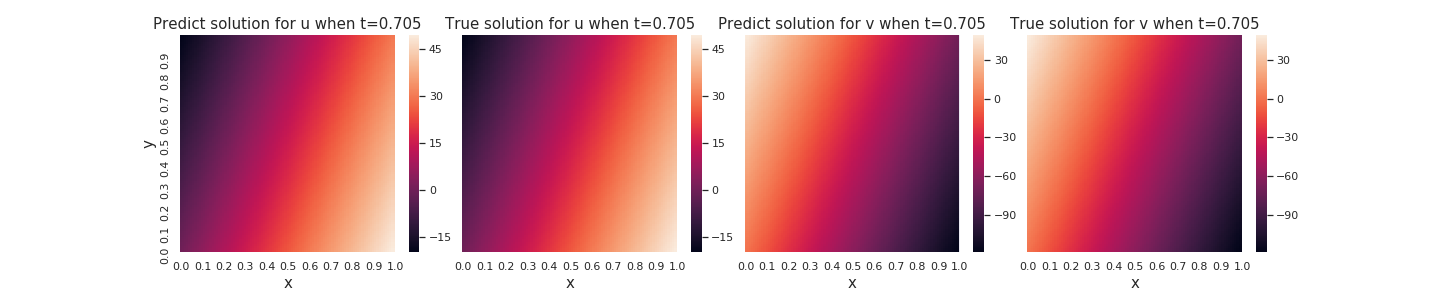}}
    \caption{Heatmap Representations for the True Solution and the Neural Approximant (Obtained By Minimizing the Boundary-Included Model) of the $2-$Dimensional Burgers' PDE Solved on Different Time Intervals Getting Close to the Blow-up Time Instant.}
    \label{Fig.Burger2.heatmap}
\end{figure}

Additionally, Figure \ref{Fig.Burger2.heatmap} shows comparative heat-map plots of the neurally obtained solution and the true blow-up solution of the two-dimensional Burgers' PDE at different time instants close to the blow-up.

Appendix \ref{Sec_1burger} discusses the experiments for the one-dimensional inviscid and viscid Burgers' PDE. Appendix \ref{Sec_2burger} gives further details on the experiments on two-dimensional Burgers' PDE. 

\subsection{PDE Solution Targets Where The Best Performance is of Models with Initial Conditions Included} \label{Best_Model_with_Initial}

In this segment we shall summarize our studies with two famous PDEs where we rigorously experimentally establish the advantages of having the initial conditions be algebraically included in the model being trained.  

\paragraph{Burgers' PDE} 
We trained $9$ models for solving the $3-$dimensional viscid Burgers' PDE case. The models account for trying $3$ neural net sizes for each of the three models, the standard PINN model, the boundary-condition-included model, and the initial-condition-included model.

For Burgers' PDE as stated in equation \ref{BurgerPDE}, but specialized to three dimensions we denote $u$ for $u_{1}$, $v$ for $u_{2}$, and $w$ for $u_{3}$ and $\nu=\frac{1}{\operatorname{Re}}$ is the kinematic viscosity. The following exact solution \citep{shukla2016modified} is used as the reference solution to test the results in the computational domain $x, y, z, t \in [0,1]$.
 \begin{gather*}
        {u}(x, y, z, t)=-\frac{2}{\mathrm{Re}} \left(\frac{1+ \cos (x) \sin (y) \sin (z) \exp (-t)}{1+x+\sin (x) \sin (y) \sin (z) \exp (-t)}\right), {v}(x, y, z, t)=-\frac{2}{\mathrm{Re}} \left(\frac{\sin (x) \cos (y) \sin (z) \exp (-t)}{1+x+\sin (x) \sin (y) \sin (z) \exp (-t)}\right)\\
        {w}(x, y, z, t)=-\frac{2}{\mathrm{Re}} \left(\frac{\sin (x) \sin (y) \cos (z) \exp (-t)}{1+x+\sin (x) \sin (y) \sin (z) \exp (-t)}\right)
\end{gather*}
Consequently, on the same computational domain as above $u_{0}$, $v_{0}$ and $w_{0}$ are the initial conditions obtained by by evaluating the above exact solutions at $t=0$. 

For the boundary conditions used in this experiment, we denote as ${g}_{x,0}(y,z,t)$ and ${g}_{x,1}(y,z,t)$ the boundary conditions for $u$ at $x=0,1$, ${g}_{y,0}(x,z,t)$ and ${g}_{y,1}(x,z,t)$ as the boundary conditions for $v$ at $y=0,1$, ${g}_{z,0}(x,y,t)$ and ${g}_{z,1}(x,y,t)$ as the boundary conditions for $w$ at $z=0,1$. Stated explicitly,
\begin{align*}
{g}_{x,0}(y, z, t)=-\frac{2}{\mathrm{Re}} \left(1+  \sin (y) \sin (z) \exp (-t)\right), &~{g}_{x,1}(y, z, t)=-\frac{2}{\mathrm{Re}} \left(\frac{1}{2+\sin (1) \sin (y) \sin (z) \exp (-t)}\right)\\
{g}_{y,0}(x, z, t)=-\frac{2}{\mathrm{Re}} \left(\frac{\sin (x) \sin (z) \exp (-t)}{1+x}\right), &~{g}_{y,1}(x, z, t)=-\frac{2}{\mathrm{Re}} \left(\frac{\sin (x) \cos (1) \sin (z) \exp (-t)}{1+x+\sin (x) \sin (1) \sin (z) \exp (-t)}\right)\\
{g}_{z,0}(x, y, t)=-\frac{2}{\mathrm{Re}} \left(\frac{\sin (x) \sin (y)  \exp (-t)}{1+x}\right), &~{g}_{z,1}(x, y, t)=-\frac{2}{\mathrm{Re}} \left(\frac{\sin (x) \sin (y) \cos (1) \exp (-t)}{1+x+\sin (x) \sin (y) \sin (1) \exp (-t)}\right)
\end{align*}
Towards solving for the above three-dimensional solution of the  Burgers' PDE, the model with the best performance turns out to be the one where the initial condition is included in the model as a parametric (in $t \in [0,1]$) convex combination with the neural net being trained. 
\begin{gather*}
        \hat{u}_{i} \coloneqq {\gN_{u}}(x,y,z, t) \cdot t  +  \cdot (1-t) \cdot u_{0}(x,y,z),
         \hat{v}_{i} \coloneqq {\gN_{v}}(x,y,z, t) \cdot t  +  \cdot (1-t) \cdot v_{0}(x,y,z), \\
        \hat{w}_{i} \coloneqq {\gN_{w}}(x,y,z, t) \cdot t  +  \cdot (1-t) \cdot w_{0}(x,y,z).
\end{gather*}
In above ${\cal N} : \R^4 \rightarrow \R^3$ is a neural net, whose three output coordinates are denoted as ${\cal N}_u, {\cal N}_v$ and ${\cal N}_w$, which is being trained to obtained ${u}$, ${v}$ and ${w}$ as specified above. We define the risk of not satisfying the PDE in the computational domain separately for each of the coordinates as ${\gR}{1}$,$~{\gR}{2}$ and ${\gR}{3}$,
\begin{align*}\label{burger3.loss1}
    {\gR}_{1} &= \left\|\frac{\partial {\hat u}_i}{\partial t}+{\hat u}_i \frac{\partial {\hat u}_i}{\partial x} + {\hat v}_i \frac{\partial {\hat u}_i}{\partial y} + {\hat w}_i \frac{\partial {\hat u}_i}{\partial z} - \nu\left(\frac{\partial^{2} {\hat u}_i}{\partial x^{2}}+\frac{\partial^{2} {\hat u}_i}{\partial y^{2}}+\frac{\partial^{2} {\hat u}_i}{\partial z^{2}}\right) \right\|_{[0,1]^3 \times[0,1], \nu_{1}}^{2},
    \\
    {\gR}_{2} &= \left\|\frac{\partial {\hat v}_i}{\partial t}+{\hat u}_i \frac{\partial {\hat v}_i}{\partial x} + {\hat v}_i \frac{\partial {\hat v}_i}{\partial y} + {\hat w}_i \frac{\partial {\hat v}_i}{\partial z} - \nu\left(\frac{\partial^{2} {\hat v}_i}{\partial x^{2}}+\frac{\partial^{2} {\hat v}_i}{\partial y^{2}}+\frac{\partial^{2} {\hat v}_i}{\partial z^{2}}\right) \right\|_{[0,1]^3 \times[0,1], \nu_{1}}^{2},
    \\
    {\gR}_{3} &= \left\|\frac{\partial {\hat w}_i}{\partial t}+ {\hat u}_i \frac{\partial {\hat w}_i}{\partial x} + {\hat v}_i \frac{\partial {\hat w}_i}{\partial y} + {\hat w}_i \frac{\partial {\hat w}_i}{\partial z} - \nu\left(\frac{\partial^{2} {\hat w}_i}{\partial x^{2}}+\frac{\partial^{2} {\hat w}_i}{\partial y^{2}}+\frac{\partial^{2} {\hat w}_i}{\partial z^{2}}\right) \right\|_{[0,1]^3 \times[0,1], \nu_{1}}^{2}.
\end{align*}
Similarly, we let ${\gR}_{4}$, ${\gR}_{5}$ and ${\gR}_{6}$ denote the risk evaluations at the boundary of the computational domain,
\begin{align*}
    {\gR}_{4} &= \left\| {\hat u}_i - {g}_{x,0}(y,z,t)\right\|_{\{0\} \times[0,1]^2 \times[0,1], \nu_{2}}^{2} + \left\| {\hat u}_i - {g}_{x,1}(y,z,t)\right\|_{\{1\} \times[0,1]^2 \times[0,1], \nu_{2}}^{2}
    \\
    {\gR}_{5} &= \left\|{\hat v}_i - {g}_{y,0}(x,z,t)\right\|_{[0,1] \times\{0\} \times[0,1] \times[0,1], \nu_{2}}^{2} + \left\|{\hat v}_i - {g}_{y,1}(x,z,t)\right\|_{[0,1] \times\{1\} \times[0,1] \times[0,1], \nu_{2}}^{2}
    \\
    {\gR}_{6} &= \left\| {\hat w}_i - {g}_{z,0}(x,y,t)\right\|_{[0,1]^2 \times \{0\} \times[0,1], \nu_{2}}^{2} + \left\| {\hat w}_i - {g}_{z,1}(x,y,t)\right\|_{[0,1]^2 \times \{1\} \times[0,1], \nu_{2}}^{2}
\end{align*}
Thus the full population risk function would be, 
\[ {\gR}_{\rm initial-included}=\left({\gR}_{1}+{\gR}_{2}+{\gR}_{3}\right)+\left({\gR}_{4}+{\gR}_{5}+{\gR}_{6}\right)
\]
In above the $\nu$s continue to denote the uniform measures on the respective domains. Minimizing the empirical risk corresponding to the above, for the $3-$dimensional viscid Burgers' PDE, the minimal fractional error we obtain is $\sim 10^{-2}$ over all values of $\nu$ tried. 

\begin{figure}[H]
    \centering  %
    \subfigure{
    \label{Burger3.nu.sub.40.1}\includegraphics[width=0.48\textwidth]{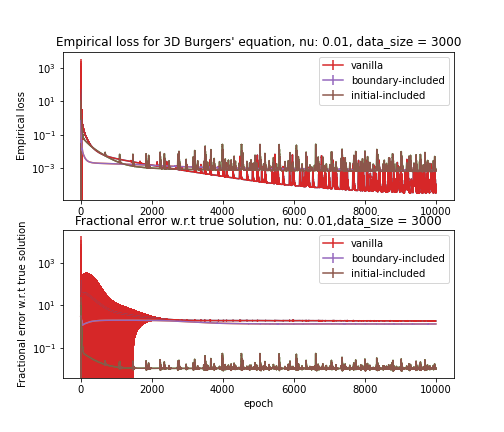}}
    \subfigure{
        \label{Burger3.nu.sub.40.4}
        \includegraphics[width=0.48\textwidth]{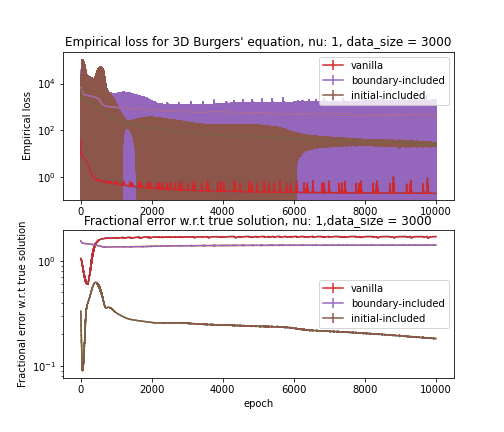}}
    \caption{Change of Empirical and Population Risk with Epochs of Training for Solving for the Target 3D Burgers' PDE at $\nu=0.01$ (Left Column) and $\nu=1$ (Right Column). The Three Lines in Each Plot Correspond to The Three Models Tried for Each $\nu$. Note That, in the Figures Above, the Legend of ``vanilla model'' Refers to the Standard PINN Method.}
    \label{Fig.Burger3.nu.40}
\end{figure}

In Figure \ref{Fig.Burger3.nu.40}, one can see the demonstration of how the initial condition inclusive model instantiated above can be trained to obtain the best values of fractional error at widely different values of the viscosity parameter. Full experimental details corresponding to this setup can be found in Appendix \ref{Sec_3burger}. 


\paragraph{KdV PDE} For the KdV PDE, $27$ models have been trained split as $12$ for solving the single-soliton solution, $12$ for solving the double-soliton solution case, and $3$ for solving the three-soliton. For single and double-soliton solutions, the standard PINN model, the boundary-condition-included model, and the initial-condition-included model were each trained at $4$ different sizes of neural networks.  

\begin{figure}[htbp!]
    \includegraphics[width=\textwidth,height = 4cm]{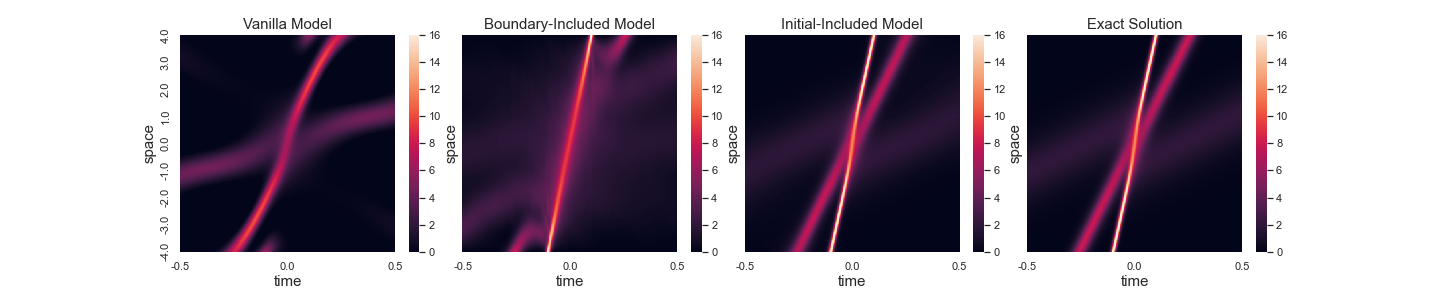}
    \caption{In the Figures Above We See a Comparison of the Outputs Between The Different Models Trying to Solve for the 3-Soliton Solution of the KdV PDE. Note That, in the Figures Above, the Legend of ``vanilla model'' Refers to the Standard PINN Method.}
    \label{fig:KdV3_heatmap}
\end{figure}

A fixed neural network was deployed for solving the three-soliton solution - which outperforms the SOTA PINN methods for solving the same. Figure \ref{fig:KdV3_heatmap} shows as a heatmap how the three models approximate the three-soliton solution of the KdV PDE compared with the exact solution.

For all solutions considered of the KdV PDE, equation \ref{KdV}, the model with the best performance is always the one which includes the initial condition as a certain parametric (in $t$ and $q$) convex combination with the neural net being trained, 

\begin{equation}\label{best_KdV_Model}
    f_{\rm {initial-included}}(x, t) := {\gN(x, t)} \cdot \frac{t^2}{t^2 + q} + \frac{q}{t^2 + q} \cdot u_0(x, 0), 
\end{equation}
where ${\gN(x, t)}$ is as defined in the equation \ref{Net}. The corresponding empirical risk function is (where the subscript on $f$ above is dropped for brevity),
\begin{equation}
    {\rm \hat{\gR}_{initial-included}} = \frac{1}{|\mathcal{D}_u|}\sum_{i=1}^{|\mathcal{D}_u|}\left |\pdv{f_{}}{t} + \pdv[3]{f_{}}{x} + 6f_{}\pdv{f_{}}{x} \right |^2_{i} + \frac{1}{|\mathcal{D}_b|}\sum_{j=1}^{|\mathcal{D}_b|} \left |f_{} - g \right |^2_{j}
\end{equation}
where $\mathcal{D}_u$ and $\mathcal{D}_b$ denote the points sets sampled from the bulk and the boundary condition and the notation of $\abs{\cdot}_i$ used above is a shorthand for denoting that the function in the argument of the absolute value function is being evaluated at the $i^{\rm th}$ point of the corresponding set.  

For the single-soliton solution case, the lowest fractional error achieved is $1.39 \times 10^{-6}$ by setting $q=10^{-9}$ in equation \ref{best_KdV_Model} and $500$ points being sampled in the set $\mathcal{D}_u$ while $250$ points being chosen in the set $\mathcal{D}_b$. Using the same $q$, $6.56 \times 10^{-1}$ is the best fractional error obtained for approximating the two-soliton solution for sizes of $\mathcal{D}_u$ and $\mathcal{D}_b$ being $2000$ and $1000$ respectively. 

\paragraph{Comparison to \citep{hu2022XPINNs}} For the $3-$soliton solution, which is among the most complicated benchmarks, a fractional error of $4.31 \times 10^{-2}$ is obtained by using $q=10^{-4}$ in the model as specifided in equation \ref{best_KdV_Model}, $18000$ points from $\mathcal{D}_u$ and 914 points from $\mathcal{D}_b$ which is the same configuration \citep{hu2022XPINNs} did. Notably, this performance is one order of magnitude better fractional error than in \citep{hu2022XPINNs} while using $12\%$ less parameters than them.

Further details on the experiments with the single-soliton can be seen in Appendix \ref{Sec_1Soliton}, for the double-soliton solution in Appendix \ref{Sec_2Soliton} and for the triple-soliton solution in Appendix \ref{Sec_3Soliton}. 

\section{Discussion} \label{section_disc}

As this work was nearing completion we became aware of the work \citep{jiang2023global} which has shown a proof of global convergence for asymptotically wide PINNs for solving second-order linear PDEs where the model being trained uses the method of \citep{mcfall2009boundary} to automatically satisfy the boundary conditions. This is closely related to the {\rm boundary-condition-included} models as we studied above. From our demonstrations, it is clear that this modification of the model is (a) of advantage for PDEs more complicated than what is in the ambit of this theory and (b) that there are PDEs where it might be superseded by the inclusion of the initial condition in the model.  In empirical studies like \citep{hu2023bias,lu2021physics}, some similar ideas like here were explored of including the initial/boundary conditions in the model - but we go much further to test out both the options for each PDE and show how within the same family the optimal modification can change with dimensions. Also, specific to heat-PDE where the DRM alternative is available, previous studies had not benchmarked the model modifications in PINNs against doing the same in the DRM framework. Thus our tests are much more robust than previous studies we are aware of. 

The detailed investigations in our work can be seen to motivate three immediate directions of research. {\em Firstly,} it is intriguing that some PDEs are better solved by algebraic inclusion of the boundary condition and some for the initial condition - and this can change with dimensions for even the same PDE, as for the Burgers. Thus an important question arises to be able to theoretically prove as to what about a PDE decides this. Also, we note that in certain cases (like for the KdV PDE and for DRM solving the Poisson PDE) we have shown that if we are allowed only short times for training then the best performing model can often be very different from that which emerges as the best model eventually. Understanding this time-accuracy trade-off further - both from theoretical as well as practical perspectives - would be a very fruitful direction of future research.  

{\em Secondly,} the novel modifications of the PINN loss that we have experimentally motivated have no proofs of convergence known for them. Provable training on such losses - and particularly at parametric width of the net is a significant and exciting challenge to take up for the future.  Works like, \citep{wu2021understanding, wang2020understanding, sukumar2022exact,gopalani2024global} have opened up new proof techniques for addressing such questions - and it would be very exciting to unravel if the mathematical structures leveraged therein continue to the more complex setups as in here.   {\em Thirdly,} Deep Ritz Method (DRM) has application to any elliptic PDE and hence an interesting question arises particularly in its context if our use of dimension multiplicatively regularized boundary condition included PINN losses can continue to supersede DRM for other elliptic PDEs too.


%% file: Computational_Resource_Appendix.tex
\section{Computational Resource}\label{appendix_computational_resource}

The GPU used in this research is Nvidia V100 from the high-performance computing cluster from the University of Manchester. Table \ref{tab:computational_resource} records the total time spent on each experiment.

\begin{table}[H]
    \centering
    \begin{tabular}{|c|c|c|c|c|}
        \hline
        \textbf{Experiment} & \textbf{NN \#Parameters} & \textbf{vanilla} & \textbf{boundary-included} & \textbf{initial-included} \\
        \hline
        \multicolumn{5}{|c|}{\em Poisson Equation} \\
        \hline
        \multirow{3}*{Energy Solution} & 1376 & na & $1.6 \times 10^{-2}$h & na \\
        \cline{2-5}
         & 20501 & na & 0.3h & na \\
        \cline{2-5}
         & 47101 & na & 1.4h & na \\
        \hline
        \multirow{3}*{Residual Solution} & 1376 & na & $1.6 \times 10^{-2}$h & na \\
        \cline{2-5}
         & 20501 & na & 0.5h & na \\
        \cline{2-5}
         & 47101 & na & 2.7h & na \\
        \hline
        \multicolumn{5}{|c|}{\em Burgers' Equation} \\
        \hline
        \multirow{4}*{1D Inviscid Solution} & 57 & 0.16h & 0.21h & 0.20h \\
        \cline{2-5}
         & 3441 & 0.30h & 0.35h & 0.37h \\
        \cline{2-5}
         & 81201 & 2.72h & 3.12h & 3.19h \\
         \cline{2-5}
         & 181801 & 6.21h & 7.24h & 7.78h \\
        \hline
        \multirow{3}*{1D Viscid Solution} & 57 & 1.71h & 2.12h & 2.22h \\
        \cline{2-5}
         & 3441 & 2.16h & 2.41h & 2.57h \\
         \cline{2-5}
         & 181801 & 2.66h & 2.93h & 3.02h \\
        \hline
        \multirow{3}*{2D Inviscid Solution} & 66 & 0.53h & 0.63h & 0.69h \\
        \cline{2-5}
         & 1794 & 0.57h & 0.76h & 0.71h \\
         \cline{2-5}
         & 20802 & 0.72h & 0.88h & 0.78h \\
        \hline
        \multirow{2}*{3D Viscid Solution} & 75 & 2.93h & 3.98h & 3.78h \\
        \cline{2-5}
         & 3641 & 10.21h & 11.26h & 11.73h \\
        \hline
        \multicolumn{5}{|c|}{\em KdV Equation} \\
        \hline
        \multirow{4}*{1-Soliton Solution} & 57 & 0.20h & 0.38h & 0.39h \\
        \cline{2-5}
         & 541 & 0.26h & 0.45h & 0.47h \\
        \cline{2-5}
         & 1009 & 0.31h & 0.51h & 0.53h \\
         \cline{2-5}
         & 1981 & 0.40h & 0.61h & 0.63h \\
        \hline
        \multirow{4}*{2-Soliton Solution} & 417 & 0.49h & 0.87h & 0.66h \\
        \cline{2-5}
         & 2109 & 0.57h & 0.98h & 0.86h \\
        \cline{2-5}
         & 4137 & 0.64h & 1.02h & 0.92h \\
         \cline{2-5}
         & 10221 & 0.76h & 1.18h & 1.04h \\
        \hline
        \multirow{1}*{3-Soliton Solution} & 3500 & 0.70h & 2.24h & 1.02h \\
        \hline
        
    \end{tabular}
    \caption{Time Cost for Each Experiment Corresponding to Each Model}
    \label{tab:computational_resource}
\end{table}

During the experiments, all models have been trained 3 times to average the performance, and therefore the actual time cost is triple the time listed above.

%% file: Sections/Poisson.tex
\section{A Study of PINN Modifications for the Poisson PDE}
\label{Sec_Poisson_exp}

We start our investigations by considering the Poisson PDE on $\Omega = {(0, 1)}^d$, 

\begin{equation}\label{eqn: poisson equation}
\nonumber - \Delta u = f(\vx ) ~\forall \vx \in \Omega, ~u(\vx ) = 0 ~\forall \vx \in \partial \Omega.
\end{equation}  

In the experiments to follow we shall use a factorizable exact solution of the above, specified as, 

\begin{equation} 
u(\vx) = \prod_{i = 1}^{d} \sin (\pi x_i) ~\text{ for,} ~f(\vx ) = d \pi^2 \prod_{i = 1}^{d} \sin (\pi x_i)
\end{equation}

To fully investigate the relative performance of energy minimization and residual minimization approaches available for this specific problem, the experiments conducted in this section investigate $6$ different loss functions that we can envisage for neural nets attempting to solve this PDE.

\subsection{Variational Energy Loss Setups}\label{sec3}

 
We recall, that in \citep{e2017deep} the Deep Ritz Methods with penalized loss function was motivated for this PDE and that is a variational energy loss setup. In the following segments, we explore that and other novel variations of it that we instantiate - which were motivated by us realizing that penalized DRM on its own does underperform for the PDE target considered here. 
 
\paragraph{Variational Energy Loss with Boundary Condition Inclusive Model}

In this setup, the Dirichlet boundary condition is enforced by directly incorporating it into the model as follows, 

\begin{equation}\label{boundmodel-energy}
{\rm model} = \hat{u}(\vx;\theta) =   \lambda^d \cdot \left (\prod_{i = 1}^{d} (1 - x_i) x_i \right ) \cdot \gN(\vx;\theta) = \prod_{i = 1}^{d} (1 - x_i) x_i \cdot \gN(\vx;\theta),
\end{equation}

where $\gN(\vx;\theta)$ is a neural network mapping $\R^d \rightarrow \R$ parameterized by weights $\theta$ and $\lambda$ is a multiplicative regularizer that we have introduced - and set to $1$ for this case. It is easy to see that the simple polynomial function multiplying the neural output ensures that the model satisfies the boundary conditions $\hat{u}(\vx)=0$ when $x_i=0$ or $x_i=1$ for any $i=1,\ldots,d$. We shall train this model via the variational energy loss denoted as $\mathcal{\hat{\gR} }_{\rm energy, boundary-included, \lambda = 1}(\hat{u})$, given as,
\begin{equation}\label{Energyloss-outscale}
\mathcal{\hat{\gR}}_{\rm energy, boundary-included, \lambda = 1}(\hat{u}) = \frac{1}{N} \sum_{i=1}^{N} \left[ \frac{1}{2} \left\vert \nabla \hat{u}(\vx_i; \theta) \right\vert^2 - f(\vx_i) \hat{u}( \vx_i; \theta) \right], \quad \text{where}\ \vx_i \in [0,1]^d,
\end{equation}
where $N$ represents the total number of mesh points used in the training. 

\paragraph{Variational Energy Loss with Boundary Condition Inclusive Model and Non-Trivial Multiplicative Regularizer}

Doing experiments with the above lead us to note that to improve the model's performance in higher dimensions the value of the regularizer needs to be tuned up. By doing a hyperparameter search we are led to choose, $\lambda = 5$ in the above - which helps counter the tendency of the model's value from getting arbitrarily low for higher dimensions.  Thus we are led to consider training the following model, 
\begin{equation}\label{scale-model-energy}
{\rm model} = \hat{u}(\vx;\theta) =   5^d \cdot \left (\prod_{i = 1}^{d} (1 - x_i) x_i \right ) \cdot \gN(\vx;\theta),
\end{equation}
using the energy loss function in equation \ref{Energyloss-outscale} but the $\hat{u}$ being understood to be the one above, 

\paragraph{Variational Energy Loss with Penalty for Boundary Condition}

The last option we try with variational losses is the original idea from  \citep{e2017deep}. Here the experiment aims to enforce the boundary conditions by adding terms to the variational energy loss which penalizes for the predictor not satisfying the boundary conditions. The population risk function in this case is defined as follows,

 \begin{equation}
\mathcal{\hat{\gR}}_{\rm energy, loss-with-boundary-penalty}(\gN) = \int_{\Omega} \left(\frac{1}{2} \left\vert \nabla \gN(\vx; \theta) \right\vert^2 - f(\vx) \gN(\vx; \theta) \right) dx + \beta \int_{\partial\Omega}\gN(\vx; \theta)^2{ds},
\end{equation}

Then for a choice of $N$ bulk mesh points ${\vx}_i \in [0,1]^d, ~i=1,\ldots, N$ and $M$ boundary points and a hyperparameter $\beta$ that controls the strength of the penalty term, the corresponding empirical risk that the algorithm attempts to minimize would be, 

\begin{equation}\label{Energyloss-pen}
\mathcal{\hat{\gR}}_{\rm energy, loss-with-boundary-penalty}(\gN) = \frac{1}{N} \sum_{i=1}^{N} \left[ \frac{1}{2} \left\vert \nabla  \gN(\vx_i;\theta) \right\vert^2 - f(\vx_i)  \gN(\vx_i;\theta) \right] + \frac{\beta}{M} \sum_{j=1}^{M} \gN(\vy_j; \theta)^2,
\end{equation}

In these implementations, the boundary points are chosen by generating $M =200$ random points on each face of the hypercube domain. 

\subsection{Residual Loss Setups}\label{sec4}
Analogous to the above, the residual losses can be set up in three different ways, 

\paragraph{Residual Loss with Boundary Condition Inclusive Model}

In here we consider the same model as in Equation \ref{boundmodel-energy}, and the corresponding empirical residual loss, denoted by $\mathcal{\hat{\gR}}_{\rm residual, boundary-included-model, \lambda = 1}(\hat{u})$, corresponding to $N$ bulk points, $\vx_i \in [0,1]^d, i=1,\ldots,N$ would be given by,

\begin{equation}\label{Residualloss-outscale}
\mathcal{\hat{\gR}}_{\rm residual, boundary-included, \lambda = 1}(\hat{u}) = \frac{1}{N} \sum_{i=1}^{N} \left(-\Delta \hat{u}({\vx}_i; \theta) - f({\vx}_i)\right)^2.
\end{equation}

We recognize that the loss function is the mean squared difference between the Laplacian of the predicted function $\hat{u}$ and the source term $f$ at each domain point ${x}_i$.

\paragraph{Residual Loss with Boundary Condition Inclusive Model and Non-Trivial Multiplicative Regularizer}


Motivated similarly, here we consider the same model as in Equation \ref{scale-model-energy}, and the corresponding empirical residual loss, denoted by $\mathcal{\hat{\gR}}_{\rm residual, boundary-included, \lambda = 5}(\hat{u})$, for a choice of $N$ bulk points, $\vx_i \in [0,1]^d, i=1,\ldots,N$, would be given by,

\begin{equation}\label{Residualloss-scale}
\mathcal{\hat{\gR}}_{\rm residual, boundary-included, \lambda = 5}(\hat{u}) = \frac{1}{N} \sum_{i=1}^{N} \left(-\Delta \hat{u}({\vx}_i; \theta) - f({\vx}_i)\right)^2, \quad \text{where}\ \vx_i \in [0,1]^d.
\end{equation}

\paragraph{Residual Loss with Boundary Penalty}
Similarly, we also consider the residual loss approach that uses the penalty terms to enforce the boundary conditions of the PDE. The loss function in this case for a choice of $N$ bulk points, $\vx_i \in [0,1]^d, i=1,\ldots, N$ and $M$ boundary points and a hyperparameter $\beta$ that controls the strength of the penalty term, would be given by, 
\begin{equation}\label{Residualloss-pen}
\mathcal{\hat{\gR}}_{\rm residual, loss-with-boundary-penalty}(\gN) = \frac{1}{N} \sum_{i=1}^{N} \left(-\Delta \gN({\vx}_i; \theta) - f({\vx}_i)\right)^2 + \frac{\beta}{M} \sum_{j=1}^{M} \gN(\vy_j; \theta)^2
\end{equation}

The relationship between these six settings delineated above can be summarized as the following diagram - whereby we also define abbreviated notations for them to be used later, 

\begin{forest}
  for tree={
    grow'=0,
    parent anchor=east,
    child anchor=west,
    edge path={
      \noexpand\path[\forestoption{edge}]
      (!u.parent anchor) -- +(5mm,0) |- (.child anchor)\forestoption{edge label};
    },
    base=bottom,
    anchor=west,
    align=left,
    l sep=7mm,
    s sep=3mm,
  }
  [{Two Families of Losses \\ for the Poisson PDE}
    [Energy
      [{Loss with Boundary Penalty, ($\mathcal{\hat{\gR}}_{\rm energy-p}$, \ref{Energyloss-pen})}]
      [Boundary Included Model
        [{$\lambda = 1$, ($\mathcal{\hat{\gR}}_{\rm energy-b(\lambda=1)}$, \ref{Energyloss-outscale} + \ref{boundmodel-energy})}]
        [{$\lambda = 5$, ($\mathcal{\hat{\gR}}_{\rm energy-b(\lambda=5)}$, \ref{Energyloss-outscale} + \ref{scale-model-energy})}]
      ]
    ]
    [Residual
      [{Loss with Boundary Penalty, ($\mathcal{\hat{\gR}}_{\rm residual-p}$, \ref{Residualloss-pen})}]
      [Boundary Included Model
        [{$\lambda = 1$, ($\mathcal{\hat{\gR}}_{\rm residual-b(\lambda=1)}$, \ref{Residualloss-outscale})}]
        [{$\lambda = 5$, ($\mathcal{\hat{\gR}}_{\rm residual-b(\lambda=5)}$, \ref{Residualloss-scale})}]
      ]
    ]
  ]
\end{forest}\label{loss_settings}

\subsection{The Hyperparameter Settings}
In Table \ref{table:loss_configurations} below we record the hyperparameter settings that were chosen corresponding to the six loss function settings previously mentioned. The training samples were progressively increased based on dimensions, with an initial value of 1000 at dimensions \( 1, 2, 3\), increasing to 5000 at dimension $10$.  And for the penalized loss settings, the experiment samples $10,000$ mesh points in $\Omega$ and $200$ points on each hyperplane that constitutes $\partial \Omega$.

The way in which the net width hyperparameter was tuned would be discussed in Section \ref{dimension-dependent}. 

\begin{table}[h]
\centering
\caption{Parameter Configuration for Different Settings.}\label{table:loss_configurations}
\begin{tabular}{c | c c c c c}
\toprule
\text{Loss Function}  & $\beta$ & \text{Learning Rate} & \text{d} & \text{Epoch} & \text{Net Width} \\
\hline
\multirow{2}{*}{$\mathcal{\hat{\gR}}_{\rm energy-b(\lambda=1)}$(\ref{Energyloss-outscale} + \ref{boundmodel-energy})} & \multirow{2}{*}{0} & \multirow{2}{*}{0.0001} & 1, 2, 3 & 2000 & \multirow{2}{*}{25, 50, 75, 100, 125, 150, 175, 200, 225, 250} \\
& & & 10 & 7000 & \\
\cline{1-6}
\multirow{2}{*}{$\mathcal{\hat{\gR}}_{\rm energy-b(\lambda=5)}(\ref{Energyloss-outscale} + \ref{scale-model-energy})$} & \multirow{2}{*}{0} & \multirow{2}{*}{0.0001} & 1, 2, 3 & 2000 & \multirow{2}{*}{25, 50, 75, 100, 125, 150} \\
& & & 10 & 7000 & \\
\cline{1-6}
\multirow{2}{*}{$\mathcal{\hat{\gR}}_{\rm energy-p}(\ref{Energyloss-pen})$} & \multirow{2}{*}{50} & \multirow{2}{*}{0.0001} & 1, 2, 3 & 2000 & \multirow{2}{*}{25, 50, 75, 100, 125, 150} \\
& & & 10 & 7000 & \\
\cline{1-6}
\multirow{2}{*}{$\mathcal{\hat{\gR}}_{\rm residual-b(\lambda=1)}(\ref{Residualloss-outscale})$} & \multirow{2}{*}{0} & \multirow{2}{*}{0.0001} & 1, 2, 3 & 2000 & \multirow{2}{*}{25, 50, 75, 100, 125, 150, 175, 200, 225, 250} \\
& & & 10 & 7000 & \\
\cline{1-6}
\multirow{2}{*}{$\mathcal{\hat{\gR}}_{\rm residual-b(\lambda=5)}(\ref{Residualloss-scale})$} & \multirow{2}{*}{0} & \multirow{2}{*}{0.0001} & 1, 2, 3 & 2000 & \multirow{2}{*}{25, 50, 75, 100, 125, 150} \\
& & & 10 & 7000 & \\
\cline{1-6}
\multirow{2}{*}{$\mathcal{\hat{\gR}}_{\rm residual-p}(\ref{Residualloss-pen})$} & \multirow{2}{*}{200} & \multirow{2}{*}{0.0001} & 1, 2, 3 & 2000 & \multirow{2}{*}{25, 50, 75, 100, 125, 150} \\
& & & 10 & 7000 & \\
\bottomrule
\end{tabular}
\end{table}

\subsection{Results About Solving the Poisson PDE in 10-Dimensions}
In higher dimensions, specifically $d = 10$, it has been observed that the six settings exhibit varying performance. A comprehensive analysis of the performance of these settings in terms of both empirical loss and fractional error at ten dimensions is provided in the subsequent discussion. We note that in these experiments $47101$ parameters are being trained. 

\paragraph{Performance of the Boundary Condition Included Models for Poisson PDE in $10-$Dimensions  - Variational Energy and Residual Loss}

\begin{figure}[htb!]
  \centering
  \subfigure{
    \includegraphics[width=3.0in, height = 2.2in]{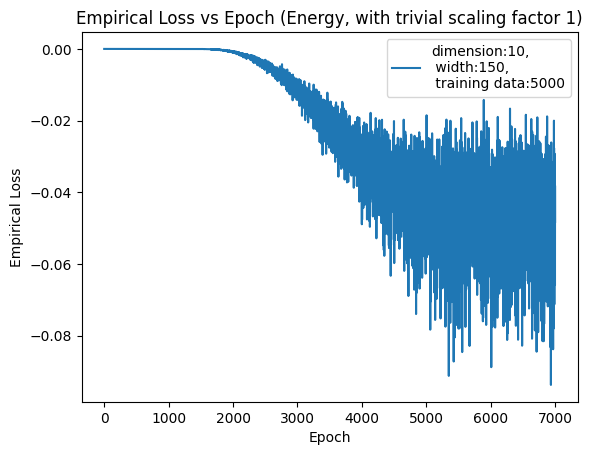}
    \includegraphics[width=3.0in, height = 2.2in]{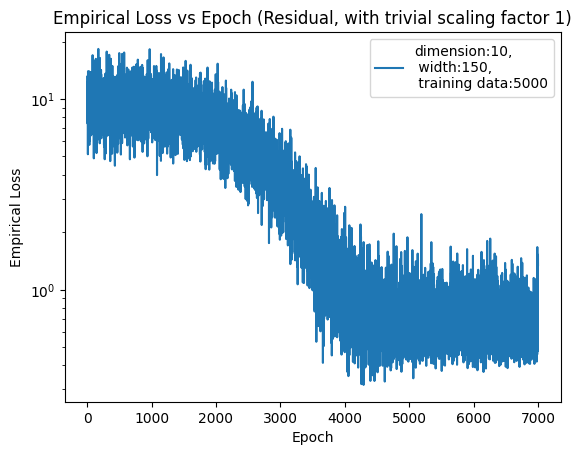}
    }
  \caption{Empirical Loss of Different Loss Function (\(d = 10 \)).
  {\it Left}  : $\mathcal{\hat{\gR}}_{\rm energy, boundary-included, \lambda = 1}$ (\ref{Energyloss-outscale}).  {\it Right}:  $\mathcal{\hat{\gR}}_{\rm residual, boundary-included, \lambda = 1}$ (\ref{Residualloss-outscale}).}
  \label{fig:empirical-loss-d10}
\end{figure}

\begin{figure}[htb!]
  \centering
  \subfigure{
    \includegraphics[width=3.0in, height = 2.2in]{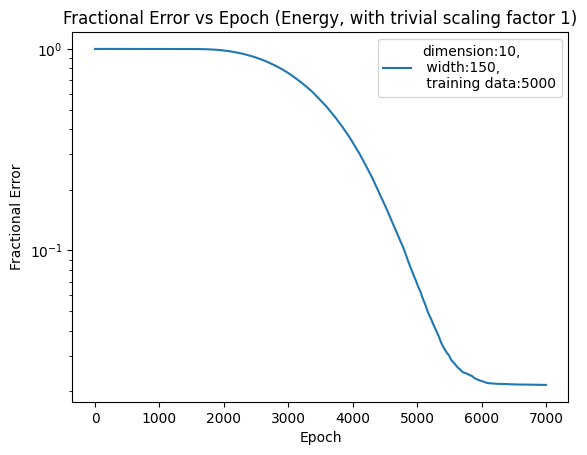}
    \includegraphics[width=3.0in, height = 2.2in]{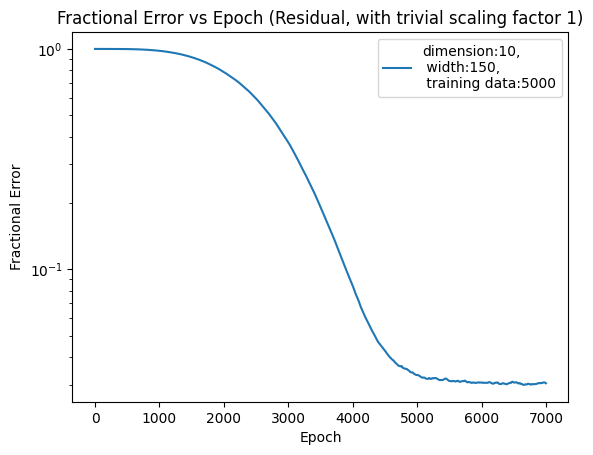}
    }
  \caption{Fractional Error of Different Loss Function (\(d= 10\)).
  {\it Left}  : $\mathcal{\hat{\gR}}_{\rm energy, boundary-included, \lambda = 1}$ (\ref{Energyloss-outscale}).  {\it Right}:  $\mathcal{\hat{\gR}}_{\rm residual, boundary-included, \lambda = 1}$ (\ref{Residualloss-outscale}).}
  \label{fig:fractional-error-d10}
\end{figure}

Initially, the investigation focuses on exploring the performance of variational energy loss (Equation \ref{Energyloss-outscale}) and residual loss (Equation \ref{Residualloss-outscale}) with boundary included model (trivial-multiplicative regularizer, $\lambda = 1$). Based on the results obtained from the initial settings, it can be observed from Figure \ref{fig:empirical-loss-d10} that the variational energy loss and the residual loss attain convergence after about 4000 epochs. Further, Figure \ref{fig:fractional-error-d10} indicates that both these methods yield similar performance in fractional error, with the error being around $10^{-1}$. \textit{In this preliminary configuration, the behavior of the two methods appear to be quite similar.}


\begin{figure}[htb!]
  \centering
  \subfigure{
    \includegraphics[width=3.0in, height = 2.2in]{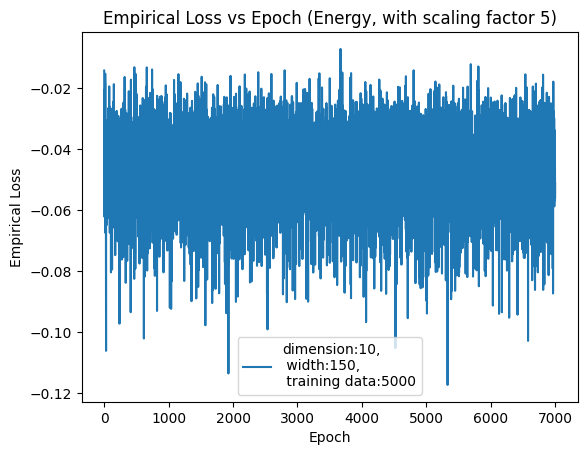}
    \includegraphics[width=3.0in, height = 2.2in]{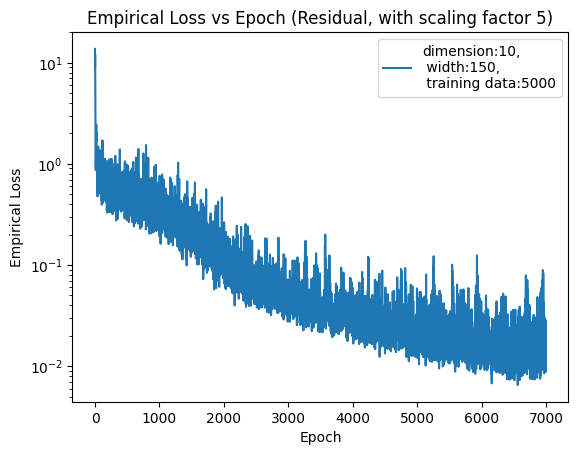}
    }
  \caption{Empirical Loss of Different Loss Function (\(d= 10\)).
  {\it Left}  : ${\rm (inadequately-trained)}\mathcal{\hat{\gR}}_{\rm energy, boundary-included, \lambda = 5}$ (\ref{Energyloss-outscale}).  {\it Right}:  $\mathcal{\hat{\gR}}_{\rm residual, boundary-included, \lambda = 5}$ (\ref{Residualloss-scale}).}
  \label{fig:empirical-loss-d10(s5)}
\end{figure}

\begin{figure}[htb!]
  \centering
  \subfigure{
    \includegraphics[width=3.0in, height = 2.2in]{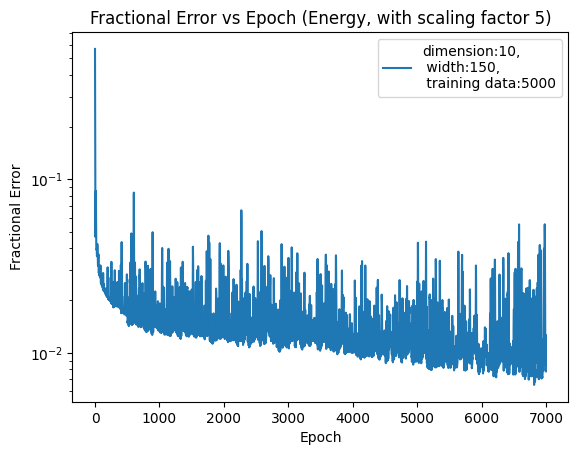}
    \fbox{\includegraphics[width=3.0in, height = 2.2in]{Images/Poisson-Images/d10_r_s5_error.png}}
    }
  \caption{Fractional Error of Different Loss Functions with the Best Performance on the Right (\(d= 10\)).
  {\it Left}  : $\mathcal{\hat{\gR}}_{\rm energy, boundary-included, \lambda = 5}$ (\ref{Energyloss-outscale}).  {\it Right}:  $\mathcal{\hat{\gR}}_{\rm residual, boundary-included, \lambda = 5}$ (\ref{Residualloss-scale}).}
  \label{fig:fractional-error-d10(s5)}
\end{figure}

\paragraph{Performance of the Boundary Condition Included Models for Poisson PDE in $10-$Dimensions  - with Multiplicative Regularizer for Variational Energy and Residual Loss}
In this section, the experiment compares the performance of variational energy loss (Equation \ref{Energyloss-outscale}) and residual loss (Equation \ref{Residualloss-scale}) with the boundary-included model (with the non-trivial multiplicative regularizer, $\lambda = 5$). The introduction of the multiplicative regularizer resulted in a significant improvement in the performance of both methods. However, Figure \ref{fig:empirical-loss-d10(s5)} illustrates that the variational energy method (on the left) appears to be inadequately trained, while the residual method (on the right) yields better results. As can be seen from Figure \ref{fig:fractional-error-d10(s5)} on the left, the fractional error for the variational energy method is found to be around $10^{-2}$, while that of the residual method on the right is about $10^{-3}$, indicating the superior performance of the residual-based approach. In this particular setting, the empirical loss derived from variational energy demonstrates inferior performance compared to the residual-based method. \textit{Having the multiplicative regularizer $\lambda = 5$ is equivalent to expanding the output of the function, which facilitates the fitting of the neural network and enables it to achieve higher fitting accuracy.}

\begin{figure}[htb!]
  \centering
  \subfigure{
    \includegraphics[width=3.0in, height = 2.2in]{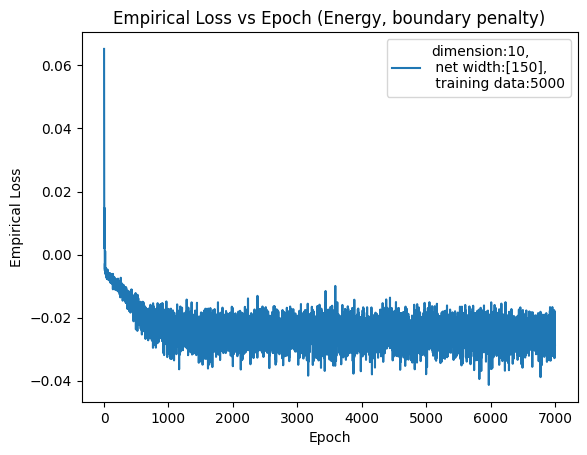}
    \includegraphics[width=3.0in, height = 2.2in]{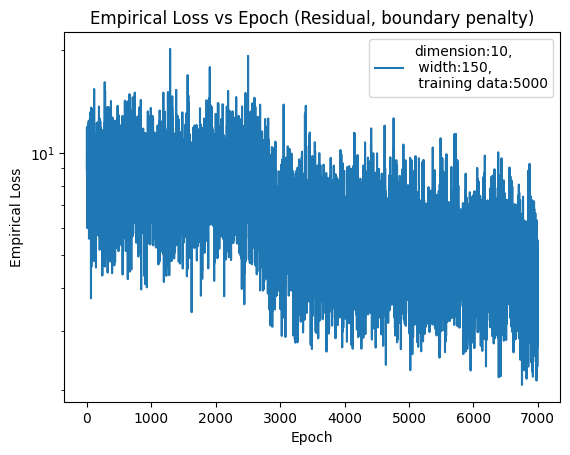}
    }
  \caption{Empirical Loss of Different Loss Function (\(d = 10\)).
  {\it Left}  : $\mathcal{\hat{\gR}}_{\rm energy, 
  loss-with-boundary-penalty}$ (\ref{Energyloss-pen}).  {\it Right}:  $\mathcal{\hat{\gR}}_{\rm residual, loss-with-boundary-penalty}$ (\ref{Residualloss-pen}).}
  \label{fig:empirical-loss-d10(pen)}
\end{figure}

\begin{figure}[H]
  \centering
  \subfigure{
    \includegraphics[width=3.0in, height = 2.2in]{Images/Poisson-Images/d10_e_pen_error.png}
    \includegraphics[width=3.0in, height = 2.2in]{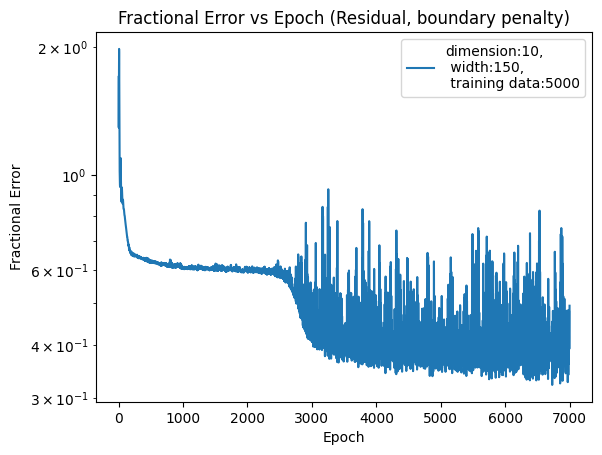}
    }
  \caption{Fractional Error of Different Loss Function (\(d = 10\)).
  {\it Left}  : $\mathcal{\hat{\gR}}_{\rm energy, loss-with-boundary-penalty}$ (\ref{Energyloss-pen}).  {\it Right}:  $\mathcal{\hat{\gR}}_{\rm residual, loss-with-boundary-penalty}$ (\ref{Residualloss-pen}).}
  \label{fig:fractional-error-d10(pen)}
\end{figure}

\paragraph{Performance of the Boundary Penalized Models for Poisson PDE in $10-$Dimensions  -  Variational Energy and Residual Loss}
The experiment further explored the performance of the approach in satisfying boundary conditions in the variational energy method by incorporating a penalty term into the loss function (Equation \ref{Energyloss-pen}), following the approach as described in \citep{e2017deep}. The same setting was then applied to residual methods to evaluate their performance (Equation \ref{Residualloss-pen}). In this specific problem, Figure \ref{fig:empirical-loss-d10(pen)} reveals that the variational methods (on the left) achieve a faster training speed, requiring only about 1000 epochs, whereas the residual method (on the right) necessitates 3000 epochs. From Figure \ref{fig:fractional-error-d10(pen)}, the left figure shows that the fractional error for variational energy methods oscillates around $60\%$, while the right figure indicates that the residual method hovers around $40\%$. \textit{It is observed that the residual loss exhibits slightly improved performance compared to the energy loss when a penalty term is added to the loss function.}

\subsection{A Study of the Dimension Dependence of Solving The Poisson PDE} 
To facilitate a clearer comparison of the performance of the different settings across different dimensions, the upcoming comparisons will illustrate the performance of the various settings as the network width is varied and the performances at different dimensions would be overlaid. We give the comparisons between $3$ pairs of models in the paragraphs to follow. 

\paragraph{Dimension Dependence of Performance of Boundary Condition Inclusive Models - Variational Energy and Residual Loss}\label{dimension-dependent}

In this segment, the investigation focuses on exploring the performance of variational energy loss (Equation \ref{Energyloss-outscale}) and residual loss (Equation \ref{Residualloss-outscale}) with boundary included model (with trivial multiplicative regularizer, $\lambda = 1$). The experiment commenced by employing equivalent epochs and training samples for dimensions $1, 2$ and $3$ by when the performances had saturated. Subsequently, the number of epochs and training samples were augmented for dimension $10$.

\begin{figure}[htp]
	\centering
	\subfigure{
    \includegraphics[width=0.5\textwidth,height = 2.5in]{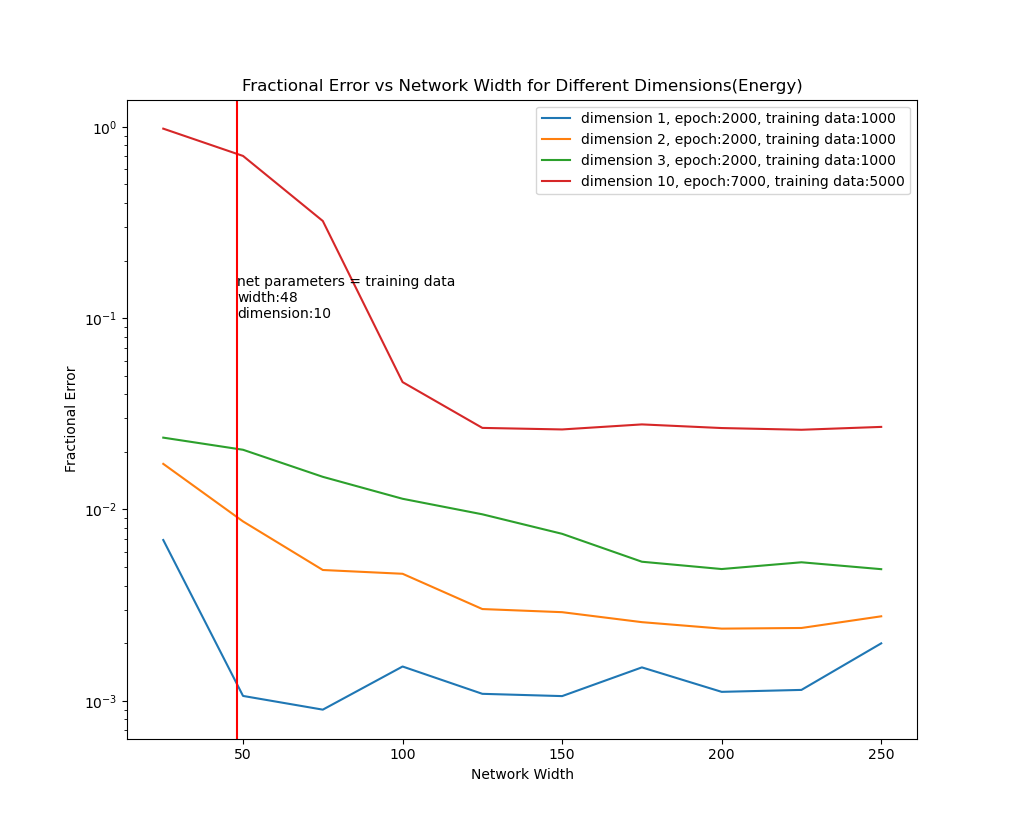}
    \includegraphics[width=0.5\textwidth,height = 2.5in]{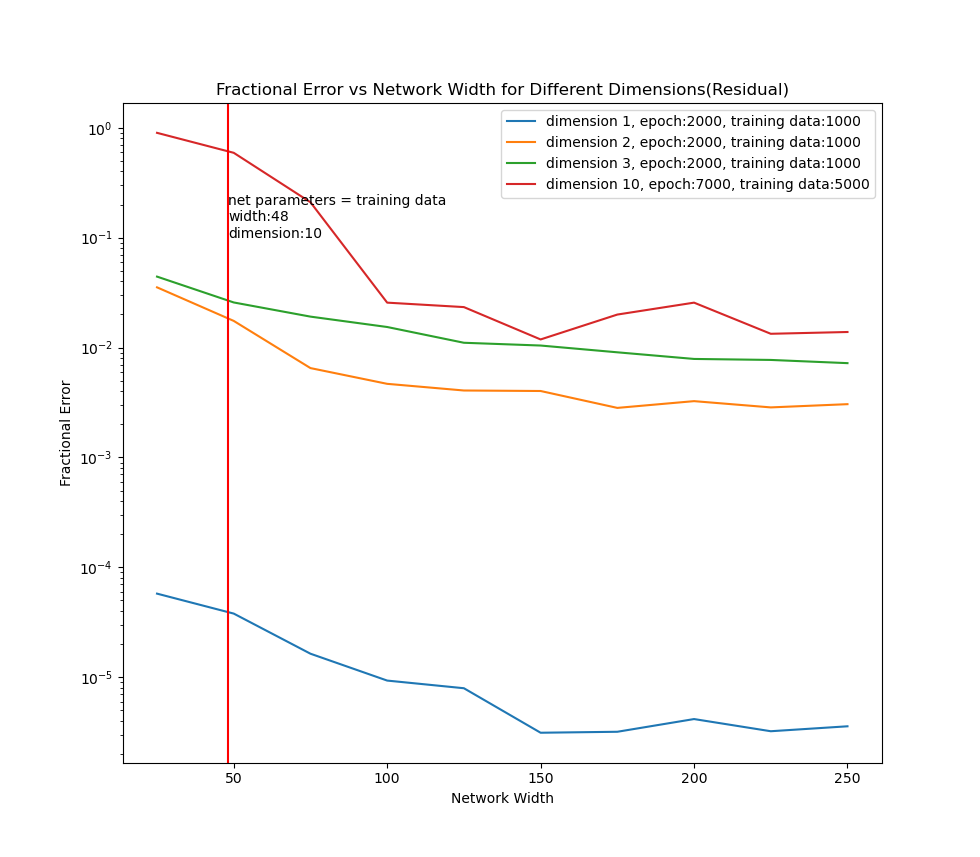}
    }
	\caption{Degradation Performance with Increasing Dimension with Net Width 250 (\(d = 1, 2, 3, 10\)).
    {\it Left}  : $\mathcal{\hat{\gR}}_{\rm energy, boundary-included, \lambda = 1}$ (\ref{Energyloss-outscale}) Lowest $\mathcal{Error}=1.50\times 10^{-3}(d=1)$ Highest $\mathcal{Error}=2.12 \times 10^{-2}(d=10)$.  {\it Right}:  $\mathcal{\hat{\gR}}_{\rm residual, boundary-included, \lambda = 1}$ (\ref{Residualloss-outscale}) Lowest $\mathcal{Error}=3.63 \times 10^{-6}(d=1)$ Highest $\mathcal{Error}=2.23 \times 10^{-2}(d=10)$.}
	\label{fig: Fractional error-width250}
\end{figure} 

The experiment first attempts to explore the network width range from 25 to 150, which indicates that in any dimension, the residual loss function exhibits superior performance compared to the energy loss function. And then the network width was raised to 250 while keeping all other parameters unchanged to investigate whether the fractional error in ten dimensions can be reduced to a similar level as in the lower dimensions by increasing the network width. 
As can be seen from Figure \ref{fig: Fractional error-width250}, the width is varying along the x-axis, and the results revealed that the issue cannot be resolved solely by increasing the network width, the performance of the variational energy loss and the residual loss was found to decrease significantly with increasing dimensionality. Figure \ref{fig: Fractional error-width250} demonstrates the absence of any discernible trend for fractional error reduction beyond a width of $150$ and hence an upper limit of $150$ for width has been selected for future experiments.

As can be seen from Figure \ref{fig: Fractional error-width250} (left), the variational energy method demonstrates a gap in performance between 10 and lower dimensions. And the right figure shows a similar phenomenon for the residual method. Overall, this indicates a similar trend in degradation performance as the dimension increases, regardless of whether it is the variational energy or residual loss being considered. 

To address this trend, our work introduces a multiplicative regularizer in an attempt to improve the performance of this setting where the boundary conditions are already satisfied by the model.

\paragraph{Dimension Dependence of Performance of Boundary Condition Inclusive Models with Multiplicative Regularizer - Variational Energy and Residual Loss}

In this segment, the experiment aims to compare the performance of variational energy loss (Equation \ref{Energyloss-outscale}) and residual loss (Equation \ref{Residualloss-scale}) in the context of a boundary-inclusive model with non-trivial multiplicative regularizer ($\lambda = 5$). Including a multiplicative regularizer resulted in considerable enhancement of the performance of both energy and residual loss. 

\begin{figure}[htb!]
  \centering
  \subfigure{
    \includegraphics[width=3.2in,height = 2.8in]{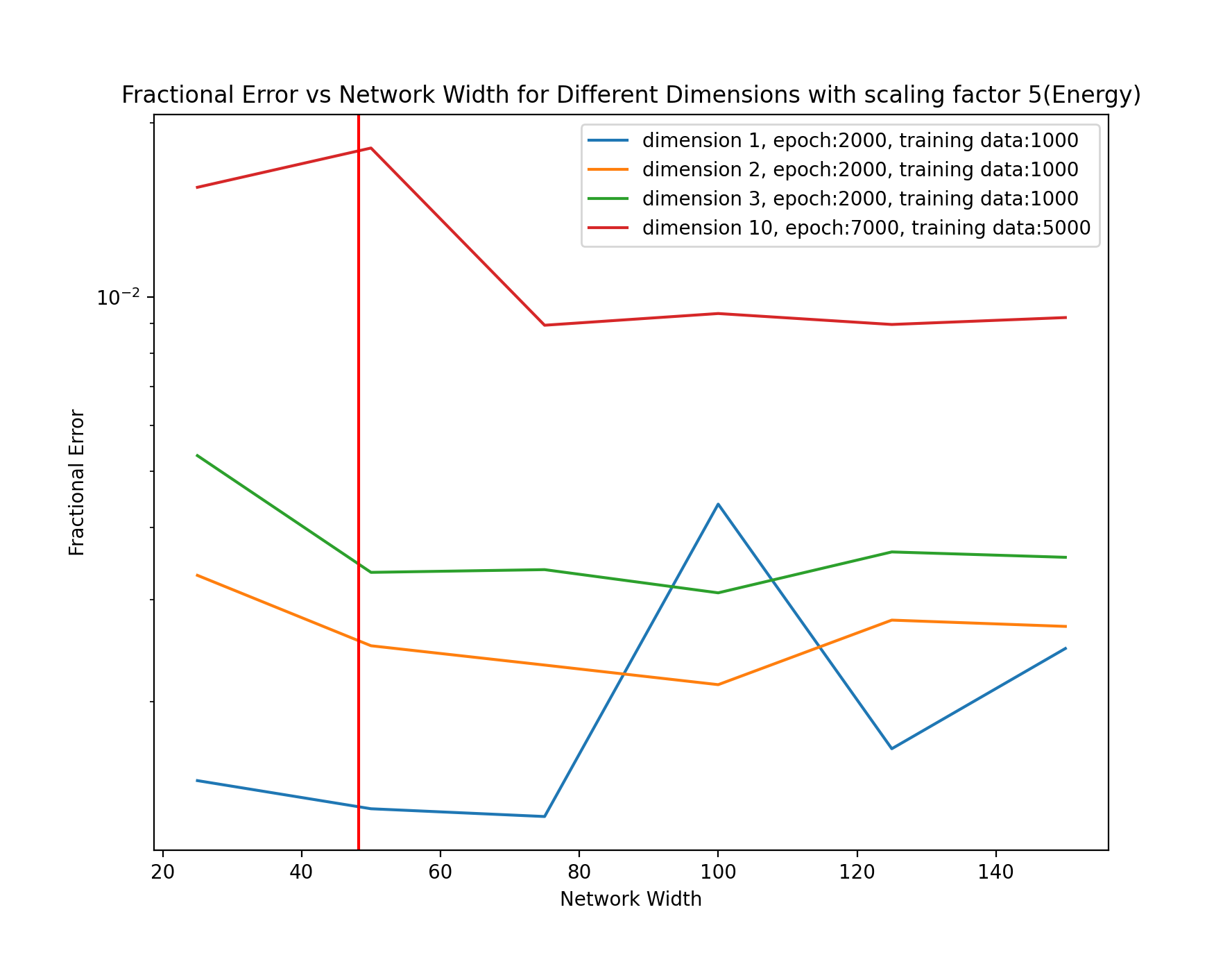}
    \fbox{\includegraphics[width=3.2in,height = 2.8in]{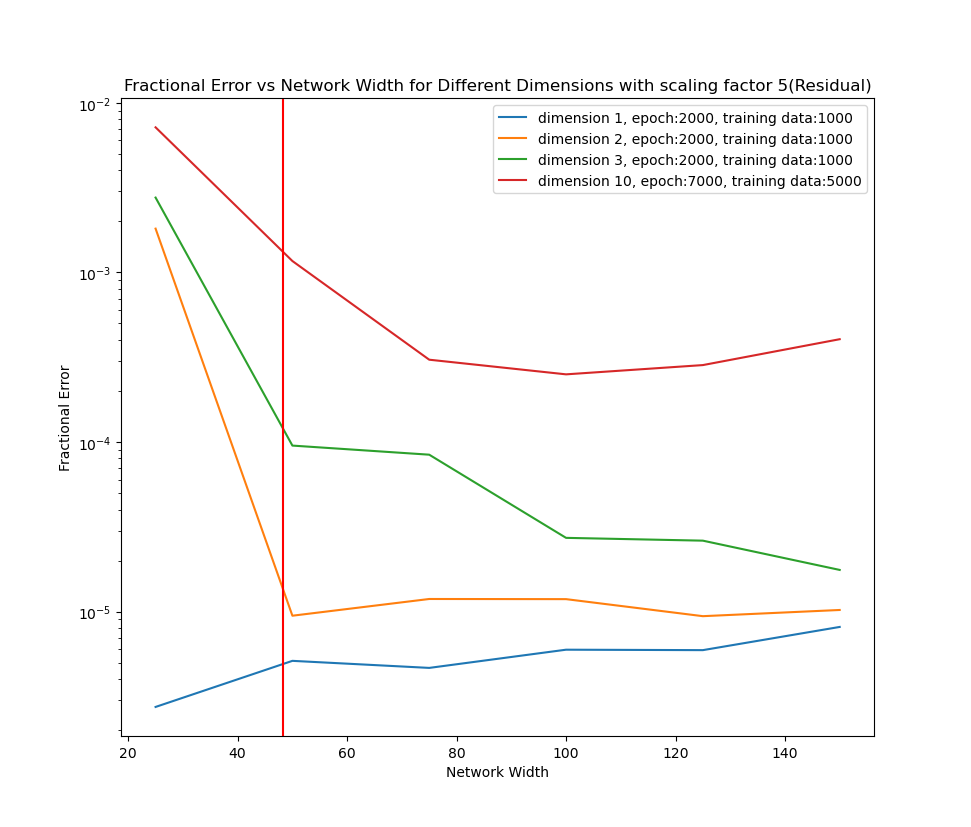}}
    }
  \caption{Fractional Error of Different Loss Functions with Multiplicative Regularizer $=5$. {\it Left}  : $\mathcal{\hat{\gR}}_{\rm energy, boundary-included, \lambda = 5}$ (\ref{Energyloss-outscale}) Lowest $\mathcal{Error}=2.50\times 10^{-3}(d=1)$ Highest $\mathcal{Error}=9.52\times 10^{-3}(d=10)$.  {\it Right}:  $\mathcal{\hat{\gR}}_{\rm residual, boundary-included, \lambda = 5}$ (\ref{Residualloss-scale}) Lowest $\mathcal{Error}=8.13\times 10^{-6}(d=1)$ Highest $\mathcal{Error}=4\times 10^{-4}(d=10)$.}
  \label{fig:fractional-error-d12310-with-scale}
\end{figure}

The left panel of Figure \ref{fig:fractional-error-d12310-with-scale} demonstrates that the fractional error of the variational energy method reaches approximately $10^{-2}$ across all dimensions after setting $\lambda = 5$. The right figure illustrates that as the network width increases, the fractional error of the residual method is around $10^{-5}$ in one, two, and three dimensions, and approximately $10^{-3}$ in 10 dimensions. \textit{Based on Figure \ref{fig:fractional-error-d12310-with-scale}, it can be observed that although both methods exhibit performance gains after scaling, the improvement in residual loss is more pronounced.}


In the setup in the previous segment, the model that satisfied the boundary conditions exhibited diminishing output values as the dimensionality increased. This resulted in a significant increase in error at higher dimensions. \textit{By adjusting the multiplicative regularizer to $5$, this experiment normalizes the empirical loss at the beginning of training to $\sim 1$.} 

\paragraph{Dimension Dependence of Performance of Boundary Condition Penalized Models - Variational Energy and Residual Loss}

This experiment is to yield observations regarding the performance of energy (Equation \ref{Energyloss-pen}) and residual (Equation \ref{Residualloss-pen}) based loss with including boundary penalty. 

\begin{figure}[htb!]
  \centering
  \subfigure{
    \includegraphics[width=3.2in,height = 2.8in]{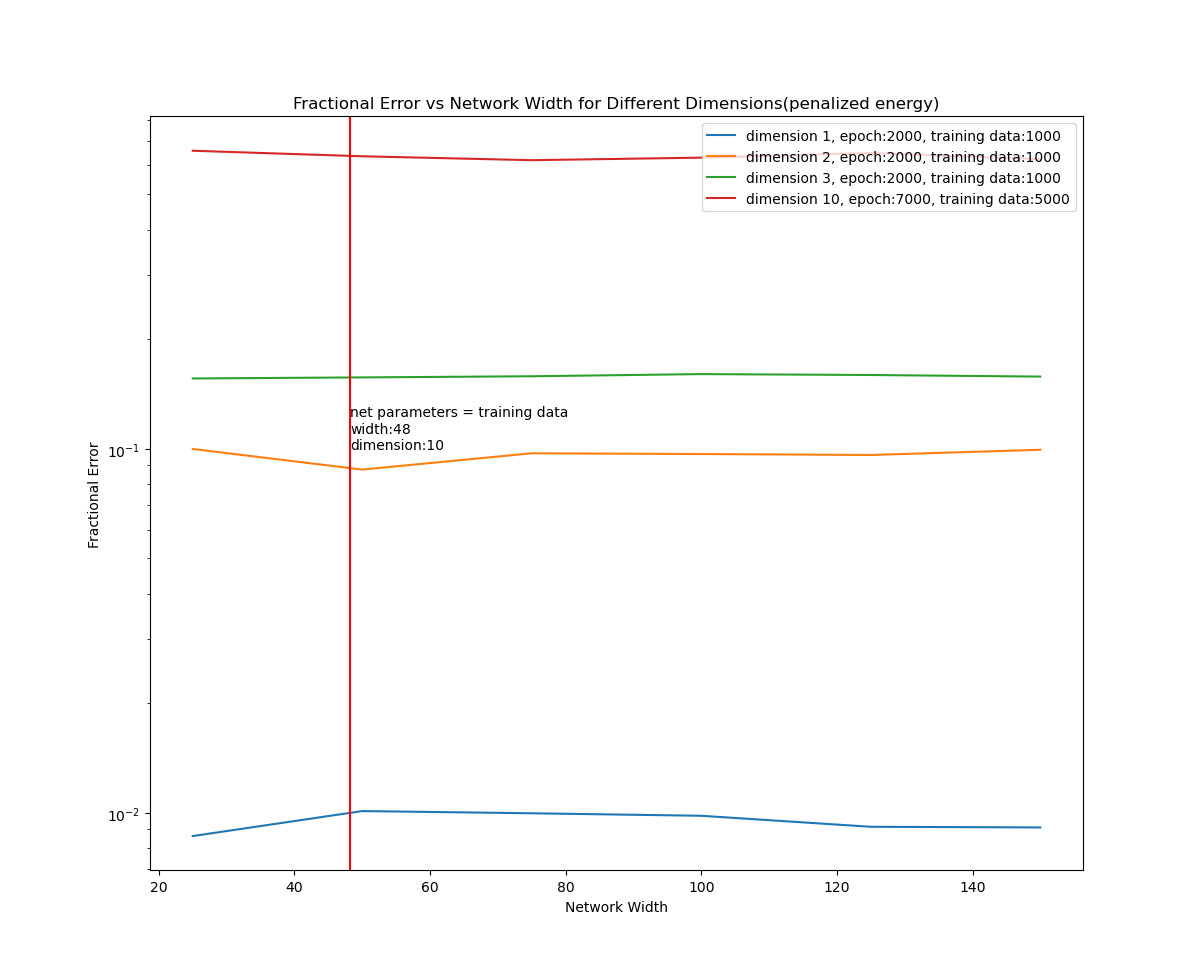}
    \includegraphics[width=3.2in,height = 2.8in]{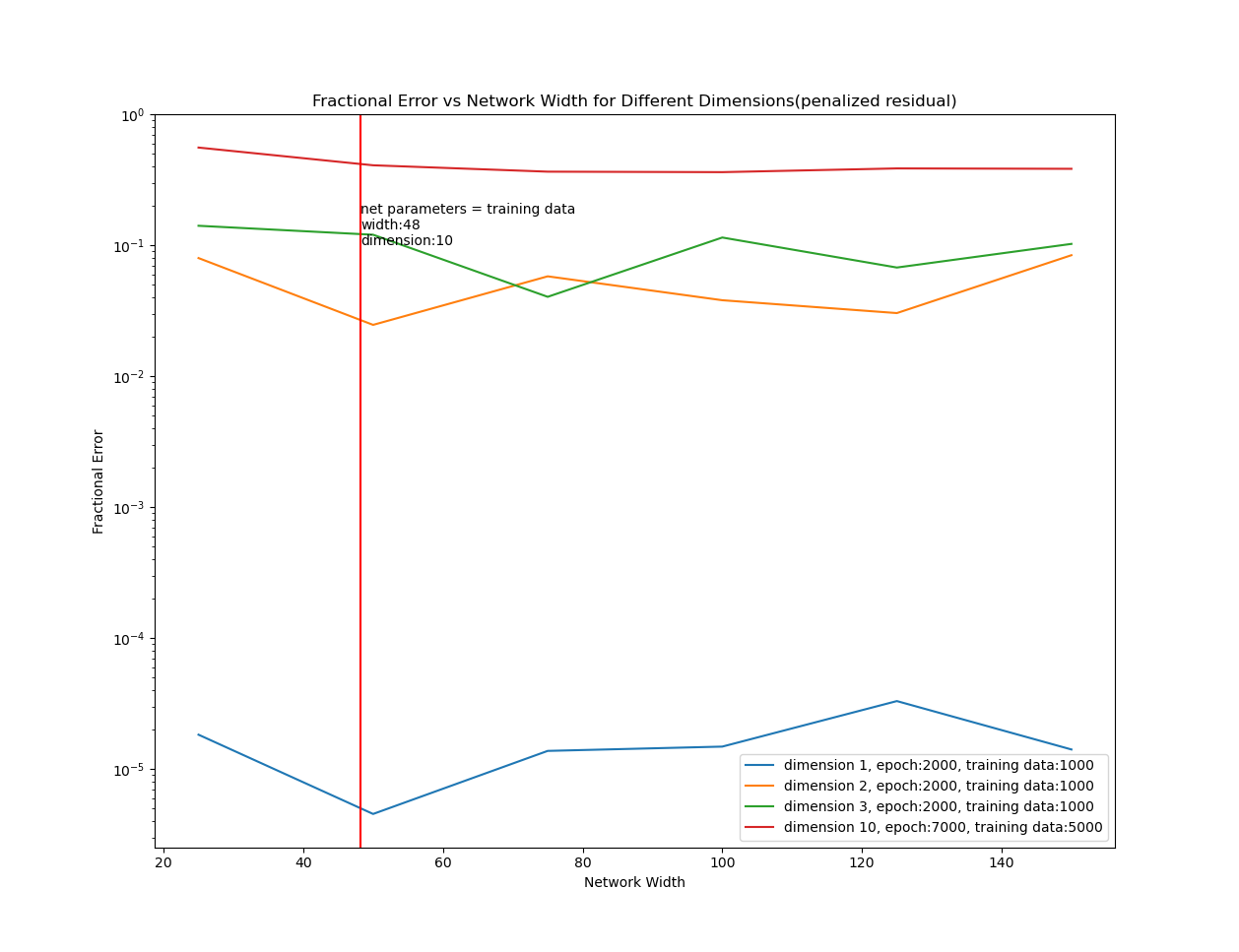}
    }
  \caption{Fractional Error of Different Loss Function (\(d = 1, 2, 3, 10\)).
  {\it Left}  : $\mathcal{\hat{\gR}}_{\rm energy, loss-with-boundary-penalty}$ (\ref{Energyloss-pen}) Lowest $\mathcal{Error}=9.11\times 10^{-3}(d=1)$ Highest $\mathcal{Error}=6.25\times 10^{-1}(d=10)$.  {\it Right}:  $\mathcal{\hat{\gR}}_{\rm residual, loss-with-boundary-penalty}$ (\ref{Residualloss-pen}) Lowest $\mathcal{Error}=1.41\times 10^{-5}(d=1)$ Highest $\mathcal{Error}=5.14\times 10^{-1}(d=10)$.}
  \label{fig:fractional-error-d12310-pen}
\end{figure}

The left panel of Figure \ref{fig:fractional-error-d12310-pen}, presents the energy method, which displays significant independence of performace with network widths. The error is approximately $10^{-2}$ in one dimension and as the dimensions increase to two and three, the error experiences a rise to about $10^{-1}$. The error exhibits an even more pronounced increase when evaluated in $10$ dimensions. On the other hand, the right figure showcases the residual method, which performs considerably better in dimension one, reaching a value of $10^{-5}$. 

The penalty terms in the loss function, responsible for satisfying the boundary conditions, necessitate sampling at the boundary. As the dimensionality increases, sampling complexity rises linearly, potentially contributing to the heightened error observed in higher dimensions. \textit{Overall, in this setting, the residual method demonstrates marginally superior performance compared to the energy method, particularly exhibiting strong performance in one dimension.}

\subsection{Reflective Summary of Solving the Poisson PDE via PINN Modifications}
\label{sum_poisson}

We recall the overview diagram \ref{loss_settings} of the models we have tested and we recall that therein we have identified six different settings to test. We emphasize that, to the best of our knowledge, there is currently no comprehensive theory available to help us decide which of these options is the most suitable or accurate. This lack of an unifying framework has led to various competing models, each with its own strengths and weaknesses. Our study aimed to take a step forward towards resolving this issue by systematically evaluating the performance of each of these six settings across different scenarios and dimensions. 


Table \ref{tab: Method compara} presents the quantitative summary comparison over all the six settings considered. The table clearly shows that $\rm Residual-b(\lambda=5)$ delivers superior results in almost all categories, surpassing the performance of other settings. Although $\mathcal{\hat{\gR}}_{\rm residual-b(\lambda=1)}$ performs slightly better in one dimension, the performance is quite similar considering orders of magnitude. 

\begin{table}[htpb!]
\centering
\caption{Fractional Error $\mathcal{Error}$ with Different Method, Width = 150.}
\label{tab: Method compara}
	\begin{tabular}{rcccccc}
		\toprule
		$d$ & $\mathcal{Error}_{\rm energy-p}$  & $\mathcal{Error}_{\rm energy-b(\lambda=1)}$ & $\mathcal{Error}_{\rm energy-b(\lambda=5)}$ & $\mathcal{Error}_{\rm residual-p}$ & $\mathcal{Error}_{\rm residual-b(\lambda=1)}$ & \textbf{\color{red}$\mathcal{Error}_{\rm residual-b(\lambda=5)}$}\\
		\hline
		1 &  $9.12\times 10^{-3}$  & $1.51\times 10^{-3}$ &  $2.53\times 10^{-3}$ & $1.41 \times 10^{-5}$ & $3.63 \times 10 ^{-6}$ &  \textbf{\color{red}$8.13 \times 10^{-6}$} \\
		\hline
		2 &  $9.94\times 10^{-2}$  & $3.21\times 10^{-3}$ & $2.73\times 10^{-3}$ & $8.42\times 10^{-2}$ & $1.51\times 10^{-3}$& \textbf{\color{red}$1.02 \times 10 ^{-5}$} \\
		\hline
		3 &  $1.58\times 10^{-1}$ & $7.51\times 10^{-3}$ & $3.63\times 10^{-3}$ & $1.03\times 10^{-1}$ & $1.11\times 10^{-2}$ & \textbf{\color{red}$1.76 \times10^{-5}$} \\
		\hline
		10 &  $6.25\times 10^{-1}$  &  $2.12\times 10^{-2}$ & $9.54\times 10^{-3}$ & $5.14\times 10^{-1}$ & $2.23\times 10^{-2}$ & \textbf{\color{red}$4.00\times 10^{-4}$} \\
		\midrule 
	\end{tabular}
\end{table}

\textit{We note that through this comparative analysis of the performance of six distinct loss function configurations in ten dimensions, this experiment identified an optimal combination: $\mathcal{\hat{\gR}}_{\rm residual, boundary-included-model, \lambda = 5}$ (\ref{Residualloss-scale})} which surpasses the setup of equation \ref{Energyloss-pen}, which follows the approach as described in \citep{e2017deep}.

\textit{But it has also been observed here that the loss function of equation ({$\mathcal{\hat{\gR}}_{\rm energy-p}$, \ref{Energyloss-pen}}), yields the most rapid convergence rate, albeit not achieving the optimal performance.} 

And it is noteworthy that both energy methods exhibit considerably faster training than the residual method.

In addition to exploring these six settings, this study also uncovered limitations when dealing with higher-dimensional problems. The experiment was conducted on the same Poisson equation (\ref{eqn: poisson equation}), extending it up to $50$ dimensions. The results indicated that training reliability consistently decreased as the problem's dimensionality increased. All variations of the energy loss function exhibited poor training performance, while the residual loss function required a considerably longer training time as the dimensionality grew.



%% file: Sections/Burger.tex
\section{A Study of PINN Modifications for the Burgers' PDE}
\label{Sec_Burger_exp}

\subsection{Boundary Condition Included Models Help with Burgers' PDE in 1+1 PDE} \label{Sec_1burger}

For $d=1$ in the Burgers' PDE, Equation \ref{BurgerPDE}, we use $u$ to represent $u_{1}$,
$$
\frac{\partial u}{\partial t}+u \frac{\partial u}{\partial x} =\nu \frac{\partial^{2} u}{\partial x^{2}}$$
\paragraph{Inviscid Burgers' PDE in $1+1$ Dimensions}
has been a subject of studies for a long time \citep{sinai1992statistics} and for $t>0$ we shall use the following exact solution for it \citep{salih2015inviscid}, 
$$u(x,t)  = \frac{\alpha x + \beta} {\alpha t+1}$$\\
In our experiments, we will choose $\alpha = 1$ and $\beta = 0$ in above and we choose the computational domain as $x, t \in [0,1]$. We define $u_{0}$ as the initial condition, and ${g}_{x,0}(t)$ and ${g}_{x,1}(t)$ as the boundary condition for $u$ when $x=0$ and $x=1$ and thus we are led to solve, 

\begin{equation}
\label{Burger1.inviscid}
\left\{
\begin{aligned}
    \frac{\partial u}{\partial t}+u \frac{\partial u}{\partial x}=0\\
    u_{0}=F(x)=x\\
        \left\{\begin{array}{l}
{g}_{x,0}(t)=0 \\
{g}_{x,1}(t)=\frac{1}{t+1}
\end{array}\right.
\end{aligned}
\right.
\end{equation}

For, ${\gN} : \R^2 \rightarrow \R$ being a neural net, we define the following models with which we shall try to obtain a solution for the above PDE. The standard PINN model being,
\[ {\rm model_{vanilla}}({x,t}) \coloneqq \hat{u}_{v} \coloneqq {\gN}(x, t) \]
The model with the boundary condition being built into it,
\[ {\rm model_{boundary-included}}({x,t}) \coloneqq \hat{u}_{b} \coloneqq {\gN}(x, t) \cdot x \cdot (1-x)  +  (1-x) \cdot {g}_{x,0}(t) + x{\cdot} {g}_{x,1}(t) \]
The model with the initial condition being built into it,
\[ {\rm model_{initial-included}}({x,t}) \coloneqq \hat{u}_{i} \coloneqq {\gN}(x, t) \cdot t  +  F(x) \cdot (1-t) \]

We denote the Lebesgue measures on the space-time domain $[0,1]^2$ as $\nu_1$ and that on the domain $t=0,x\in [0,1]$ as $\nu_2$ and that on the domain $\{0,1\} \times [0,1]$ as $\nu_3$. Hence it follows that the population risks for training each of the above models are as follows,  

\[ \operatorname{{\gR} _{\rm vanilla}}({u}_{v}) \coloneqq \left\|\frac{\partial {u}_{v}}{\partial t}+{u}_{v} \frac{\partial {u}_{v}}{\partial x}\right\|_{[0,1] \times[0,1], \nu_{1}}^{2}+\|{u}_{v}-F(x)\|_{t=0,[0,1], \nu_{2}}^{2}+\|{u}_{v}-g(t)\|_{\{0,1\} \times[0,1], \nu_{3}}^{2} \] 


\[ \operatorname{{\gR} _{\rm boundary-included}}({u}_{b}) \coloneqq \left\|\frac{\partial {u}_{b}}{\partial t}+{u}_{b} \frac{\partial {u}_{b}}{\partial x}\right\|_{[0,1] \times[0,1], \nu_{1}}^{2}+\|{u}_{b}-F(x)\|_{t=0,[0,1], \nu_{2}}^{2} \]


\[ \operatorname{{\gR} _{\rm initial-included}}({u}_{i}) \coloneqq \left\|\frac{\partial {u}_{i}}{\partial t}+{u}_{i} \frac{\partial {u}_{i}}{\partial x}\right\|_{[0,1] \times[0,1], \nu_{1}}^{2}+\|{u}_{i}-g(t)\|_{\{0,1\} \times[0,1], \nu_{3}}^{2} \]

In the above notation above it is understood that the symbol $g$ will get specified to the function $g_{x,0}$ or $g_{x,1}$ depending on the value of $x$ it is being evaluated at. 

In Table \ref{tab:Burger1.para}, we see that each of the models above have been tested at different values of the over-parameterization ratio - and the best performing scenario has been marked in red. 

\begin{table}[h]
\centering
\begin{tabular}{|c|c|c|}
  \hline
  \multicolumn{3}{|c|}{\text { 1D Burgers', epoch }= 30000, {learning rate}=0.0001, p = \#parameters} \\
  \hline
  Sub-Figures of Figure \ref{Fig.Burger1.trans}& \text { \#parameters } & \text { mesh size } \\
  \hline
  \text {(a) ~[$p \less n$]} & \makecell*[c]{\text 57} & \multirow{4}{*}{\makecell*[c]{\textbf{vanilla}:{ ~bulk mesh size}=1000,\text {initial conditions slice mesh size}=1000, \\ \text{boundary mesh size}=1000; \textcolor{red}{\textbf{boundary-included}:bulk mesh size = 1000,}\\  \textcolor{red}{\text{initial conditions slice mesh size} = 1000,\text {boundary mesh size} = 0,}\\ \textbf{initial-included}:{ bulk mesh size} = 1000,\text {initial mesh size} = 0, \\ \text {boundary mesh size} = $1000$($500$ for $x=1$ and $500$ for $x=0$) } }\\
  \cline{1-2}
  \text {(b) ~[$p\approx n$]} & \makecell*[c]{\text 3441} & \\ 
  \cline{1-2}
  \text {(c) ~[$ p\geq n$]}& \makecell*[c]{\text 81201} &\\ 
  \cline{1-2}
  \text {(d) ~[$p\gg n$]}& \makecell*[c]{\text 181801} & \\ 
  \hline
\end{tabular}
\caption{1D Inviscid Burgers' PDE Experiments with Different Number of Trainable Parameters and $n$ Being the Mesh Size Used for the Vanilla Model}
\label{tab:Burger1.para}
\end{table}

\begin{figure}[H]
    \centering  %
    \subfigure[Results for the 57 Parameters Net]{
        \centering
        \includegraphics[width=0.48\textwidth]{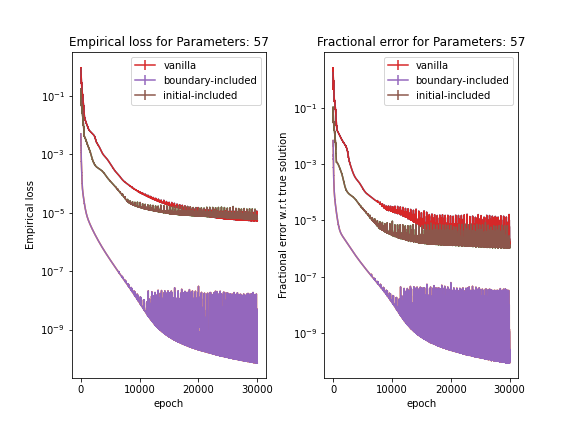}}
        \label{Burger1.trans.sub.1}
    \subfigure[Results for the 3441 Parameters Net]{
        \label{Burger1.trans.sub.2}
        \includegraphics[width=0.48\textwidth]{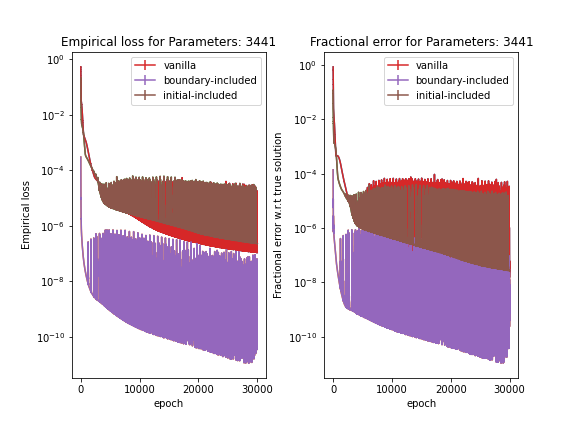}}
    \subfigure[Results for the 81201 Parameters Net]{
        \label{Burger1.trans.sub.3}
        \includegraphics[width=0.48\textwidth]{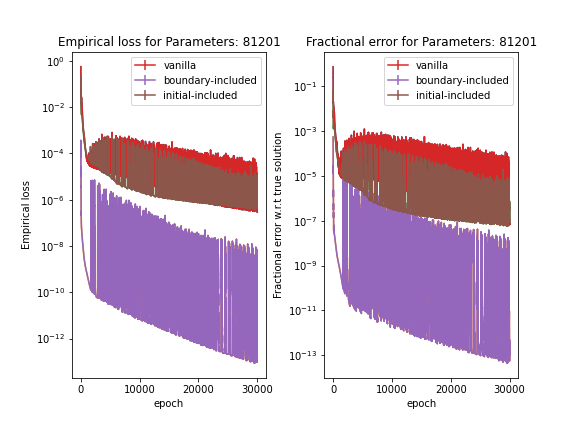}}
    \subfigure[Results for the 181801 Parameters Net]{
        \label{Burger1.trans.sub.4}
        \includegraphics[width=0.48\textwidth]{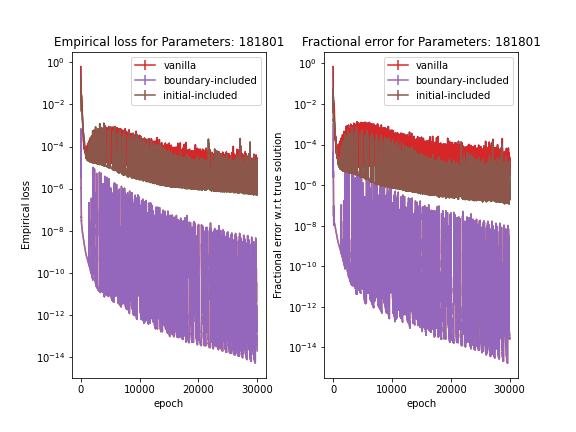}}
    \caption{Training Progress of the Different Models at Different Levels of Overparameterization When Solving the $1$D Inviscid Burgers' PDE Using $3\times 10^{4}$ Epochs, $10^{-4}$ Learning-Rate and a Space-Time Mesh Size of $3000$}
    \label{Fig.Burger1.trans}
\end{figure}

Thus from figure \ref{Fig.Burger1.trans}, we can conclude that \textit{no matter how large our number of parameters is, the boundary-included model always has the best performance.}

\paragraph{Viscid Burgers' PDE in $1+1$ Dimensions}
This is a  widely used PDE \citep{bendaas2018periodic} and in this segment, we shall test performances against the following exact solution for it, ~\citep{agarwal2011burgers}, 

$$u(x, t)=1-\tanh \frac{x-x_{c}-t}{2 \nu}$$

where $x_c$ is an arbitrary constant. We can choose the computational domain as $x, t \in [0,1]$ and define $u_{0}$ as the initial condition, and ${g}_{x,0}(t)$ and ${g}_{x,1}(t)$ as the boundary conditions for $u$ when $x=0$ and $x=1$. Then the above can be seen as a solution to the PDE, 

\begin{equation}
\label{Burger1.viscid}
\left\{
\begin{aligned}
\frac{\partial u}{\partial t}+u \frac{\partial u}{\partial x}=\nu \frac{\partial^2 u}{\partial x^2} \\
u_{0}=1-\tanh \frac{x-x_{c}}{2 \nu}\\
\left\{\begin{array}{l}
{g}_{x,0}(t)=1-\tanh \frac{x_{c}-t}{2 \nu}\\
{g}_{x,1}(t)=1-\tanh \frac{1-x_{c}-t}{2 \nu}
\end{array}\right.
\end{aligned}
\right.
\end{equation}

The model and loss function for the neural network will be the same as for the the 1D inviscid Burgers' PDE.

Tables \ref{tab:Burger1.nu.4}, \ref{tab:Burger1.nu.40} and \ref{tab:Burger1.nu.300} given below summarize the experimental results for solving the above PDE at different values of the kinematic viscosity parameter $\nu$ and different models (vanilla, boundary-included, and initial-included) -- as in the previous subsection. After a hyperparameter search the tests were decided to be reported for $3 \times 10^4$ epochs and at a learning rate of $10^{-4}$.  

For every value of $\nu$, we tested at different values of number of parameters for the nets, $57, ~3441$ and $181801$ - to cover various ranges of overparamatererization. The best performing models, for all the cases, have been summarized at the end in table \ref{tab:Burger1.nu.best.model}.


\clearpage 

\begin{table}[h]
\centering
\begin{tabular}{|c|c|c|}
  \hline
  \multicolumn{3}{|c|}{\text { 1D Burgers', epoch }=30000, {lr}=0.0001,\#parameters=57} \\
  \hline
  Sub-Figures of Figure \ref{Fig.Burger1.nu.4}& \text { $\nu$ } & \text { model } \\
  \hline
  \text {(a) Experiment1} & \makecell*[c]{\text {$\nu$}=0.01} & \multirow{6}{*}{\makecell*[c]{\textbf{vanilla}:{ bulk mesh size}=1000,\text {initial mesh size}=1000,\\ \text{boundary mesh size}=1000 \\ \textbf{boundary-included}:{ bulk mesh size} = 1000,\text {initial mesh size} = 1000, \\ \text {boundary mesh size} = 0,\\ \textbf{initial-included}:{ bulk mesh size} = 1000,\text {initial mesh size} = 0, \\ \text {boundary mesh size} = 1000 } } \\ 
  \cline{1-2}
  \text {(b) Experiment2} & \makecell*[c]{\text {$\nu$}=0.1} & \\ 
  \cline{1-2}
  \text {(c) Experiment3}& \makecell*[c]{\text {$\nu$}=0.5} &\\ 
  \cline{1-2}
  \text {(d) Experiment4}& \makecell*[c]{\text {$\nu$}=1} & \\ 
  \cline{1-2}
  \text {(e) Experiment5}& \makecell*[c]{\text {$\nu$}=2} &\\ 
  \cline{1-2}
  \text {(f) Experiment6}& \makecell*[c]{\text {$\nu$}=10} &\\ 
  \hline
\end{tabular}
\caption{1D Viscid Burgers' PDE Experiments at Different Values of $\nu$ and Number of Trainable Parameters $= 57$}
\label{tab:Burger1.nu.4}
\end{table}

\begin{table}[h]
\centering
\begin{tabular}{|c|c|c|}
  \hline
  \multicolumn{3}{|c|}{\text { 1D Burgers', epoch }=30000, {lr}=0.0001, \#parameters=3441} \\
  \hline
  Sub-Figures of Figure \ref{Fig.Burger1.nu.40}& \text { $\nu$ } & \text { model } \\
  \hline
  \text {(a) Experiment1} & \makecell*[c]{\text {$\nu$}=0.01} & \multirow{4}{*}{\makecell*[c]{\textbf{vanilla}:{ bulk mesh size}=1000,\text {initial mesh size}=1000,\\ \text{boundary mesh size}=1000 \\ \textbf{boundary-included}:{ bulk mesh size} = 1000,\text {initial mesh size} = 1000, \\ \text {boundary mesh size} = 0,\\ \textbf{initial-included}:{ bulk mesh size} = 1000,\text {initial mesh size} = 0, \\ \text {boundary mesh size} = 1000 } } \\ 
  \cline{1-2}
  \text {(b) Experiment2} & \makecell*[c]{\text {$\nu$}=0.1} & \\ 
  \cline{1-2}
  \text {(c) Experiment3}& \makecell*[c]{\text {$\nu$}=0.5} &\\ 
  \cline{1-2}
  \text {(d) Experiment4}& \makecell*[c]{\text {$\nu$}=1} & \\ 
  \cline{1-2}
  \text {(e) Experiment5}& \makecell*[c]{\text {$\nu$}=2} & \\ 
  \cline{1-2}
  \text {(e) Experiment6}& \makecell*[c]{\text {$\nu$}=10} & \\ 
  \hline
\end{tabular}
\caption{1D Viscid Burgers' PDE Experiments at Different Values of $\nu$ and Number of Trainable Parameters $= 3441$}
\label{tab:Burger1.nu.40}
\end{table}

\begin{table}[h]
\centering
\begin{tabular}{|c|c|c|}
  \hline
  \multicolumn{3}{|c|}{\text { 1D Burgers', epoch }=30000, {lr}=0.0001, \#parameters=181801} \\
  \hline
  Sub-Figures of Figure \ref{Fig.Burger1.nu.400}& \text { $\nu$ } & \text { model } \\
  \hline
  \text {(a) Experiment1} & \makecell*[c]{\text {$\nu$}=0.01} & \multirow{4}{*}{\makecell*[c]{\textbf{vanilla}:{ bulk mesh size}=1000,\text {initial mesh size}=1000,\\ \text{boundary mesh size}=1000 \\ \textbf{boundary-included}:{ bulk mesh size} = 1000,\text {initial mesh size} = 1000, \\ \text {boundary mesh size} = 0,\\ \textbf{initial-included}:{ bulk mesh size} = 1000,\text {initial mesh size} = 0, \\ \text {boundary mesh size} = 1000 } } \\ 
  \cline{1-2}
  \text {(b) Experiment2} & \makecell*[c]{\text {$\nu$}=0.1} & \\ 
  \cline{1-2}
  \text {(c) Experiment3}& \makecell*[c]{\text {$\nu$}=0.5} &\\ 
  \cline{1-2}
  \text {(d) Experiment4}& \makecell*[c]{\text {$\nu$}=1} & \\ 
  \cline{1-2}
  \text {(e) Experiment5}& \makecell*[c]{\text {$\nu$}=2} & \\ 
  \cline{1-2}
  \text {(f) Experiment6}& \makecell*[c]{\text {$\nu$}=10} & \\ 
  \hline
\end{tabular}
\caption{1D Viscid Burgers' PDE Experiments at Different Values of $\nu$ and Number of Trainable Parameters $= 181801$}
\label{tab:Burger1.nu.300}
\end{table}

\begin{figure}[H]
    \centering  %
    \subfigure[Results for $\nu$ = 0.01]{
        \label{Burger1.nu.sub.4.1}
        \includegraphics[width=0.46\textwidth]{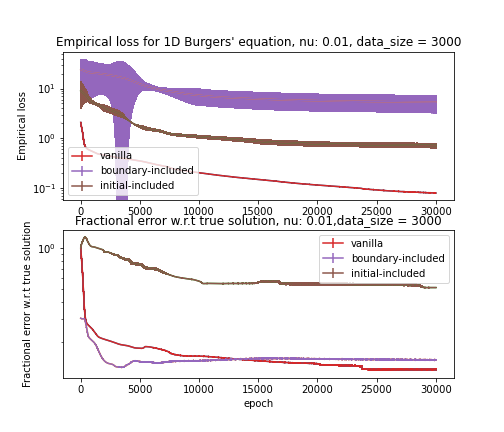}}
    \subfigure[Results for $\nu$ = 0.1]{
        \label{Burger1.nu.sub.4.2}       
        \includegraphics[width=0.46\textwidth]{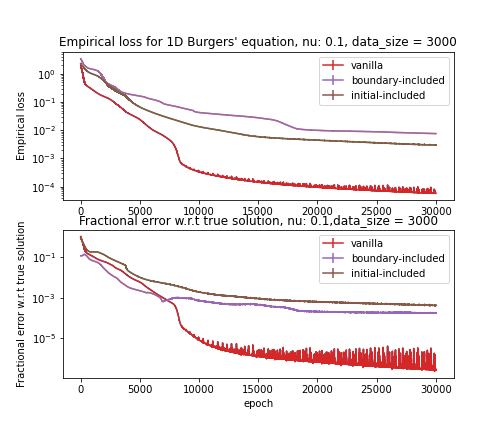}}
    \subfigure[Results for $\nu$ = 0.5]{
        \label{Burger1.nu.sub.4.3}
        \includegraphics[width=0.46\textwidth]{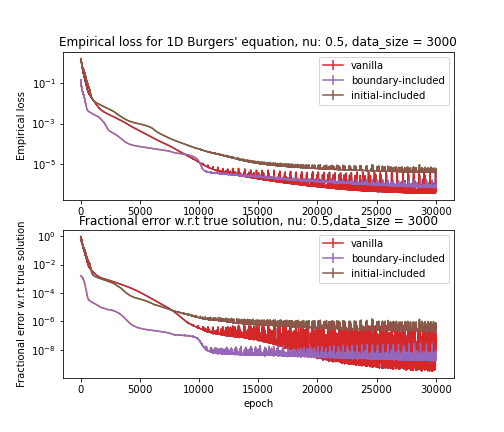}}
    \subfigure[Results for $\nu$ = 1]{
        \label{Burger1.nu.sub.4.4}
        \includegraphics[width=0.46\textwidth]{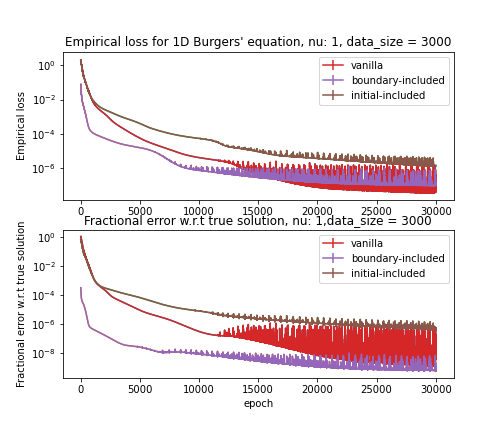}}
    \subfigure[Results for $\nu$ = 2]{
        \label{Burger1.nu.sub.4.5}
        \includegraphics[width=0.46\textwidth]{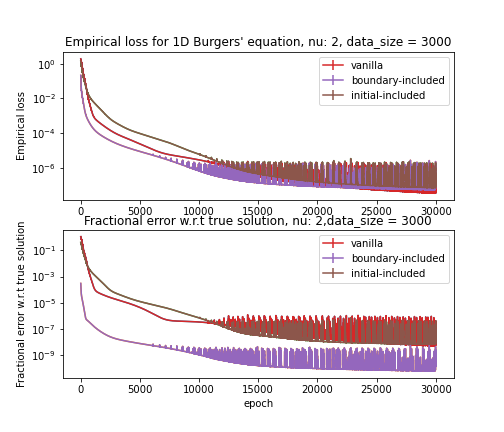}}
    \subfigure[Results for $\nu$ = 10]{
        \label{Burger1.nu.sub.4.6}
        \includegraphics[width=0.46\textwidth]{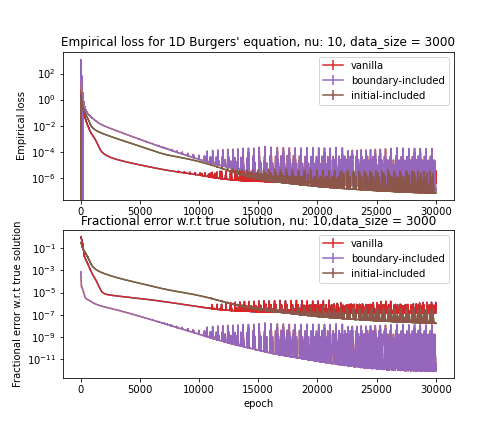}}
    \caption{Training Progress of the Different Models for Solving 1D Burgers' PDE Using $3 \times 10^4$ Epochs, $57$ Parameters, lr $= 0.0001$ and a Mesh Size of $3000$}
    \label{Fig.Burger1.nu.4}
\end{figure}

\begin{figure}[H]
    \centering  %
    \subfigure[Results for $\nu$ = 0.01]{
        \label{Burger1.nu.sub.40.1}
        \includegraphics[width=0.46\textwidth]{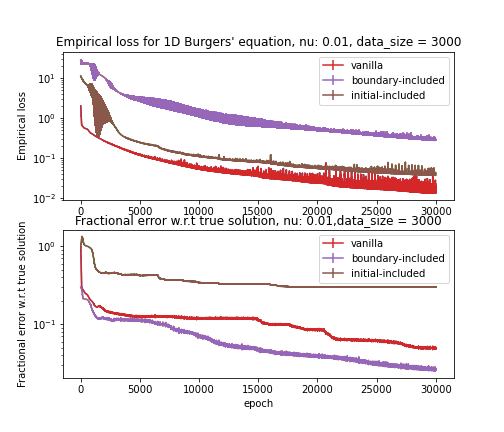}}
    \subfigure[Results for $\nu$ = 0.1]{
        \label{Burger1.nu.sub.40.2}
        \includegraphics[width=0.46\textwidth]{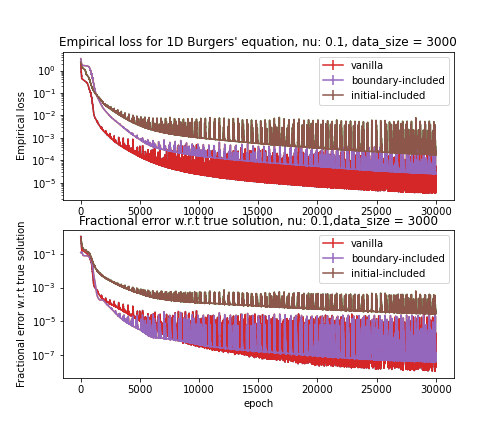}}
    \subfigure[Results for $\nu$ = 0.5]{
        \label{Burger1.nu.sub.40.3}
        \includegraphics[width=0.46\textwidth]{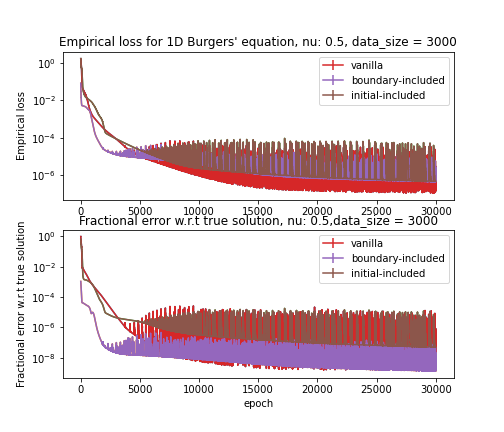}}
    \subfigure[Results for $\nu$ = 1]{
        \label{Burger1.nu.sub.40.4}
        \includegraphics[width=0.46\textwidth]{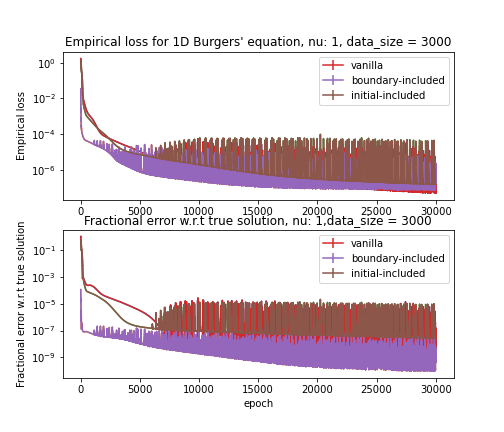}}
    \subfigure[Results for $\nu$ = 2]{
        \label{Burger1.nu.sub.40.5}
        \includegraphics[width=0.46\textwidth]{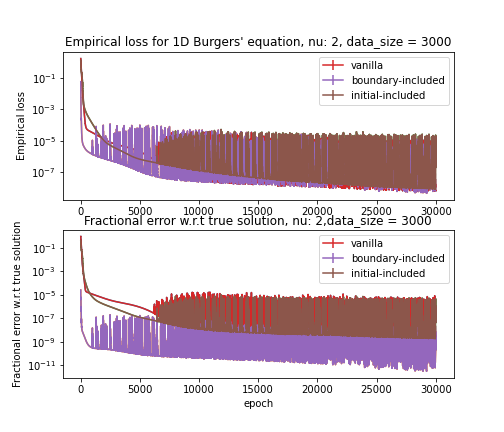}}
    \subfigure[Results for $\nu$ = 10]{
        \label{Burger1.nu.sub.40.6}
        \includegraphics[width=0.46\textwidth]{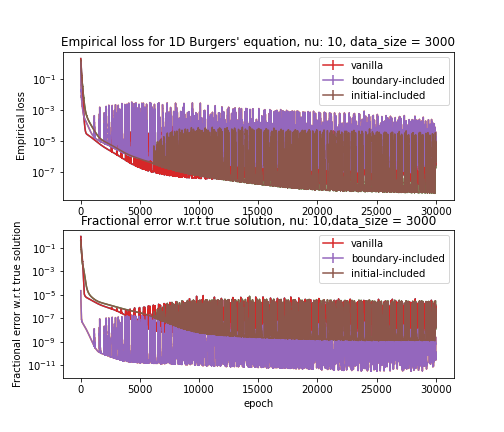}}
    \caption{Training Progress of the Different Models for Solving 1D Burgers' PDE Using $3 \times 10^4$ Epochs, $3441$ Parameters, lr $= 0.0001$ and Mesh Size $=3000$}
    \label{Fig.Burger1.nu.40}
\end{figure}

\begin{figure}[H]
    \centering  %
    \subfigure[Results for $\nu$ = 0.01]{
        \label{Burger1.nu.sub.400.1}
        \includegraphics[width=0.46\textwidth]{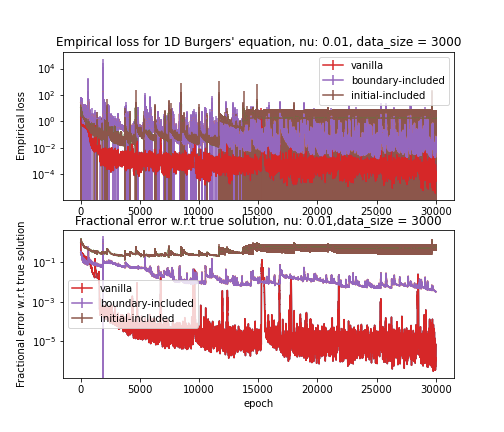}}
    \subfigure[Results for $\nu$ = 0.1]{
        \label{Burger1.nu.sub.400.2}
        \includegraphics[width=0.46\textwidth]{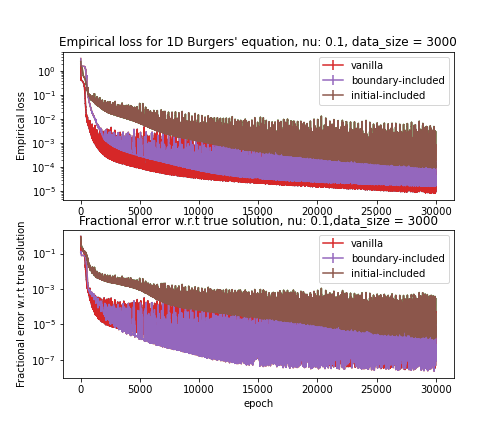}}
    \subfigure[Results for $\nu$ = 0.5]{
        \label{Burger1.nu.sub.400.3}
        \includegraphics[width=0.46\textwidth]{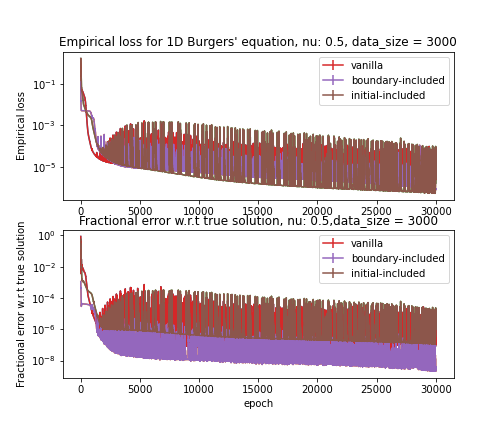}}
    \subfigure[Results for $\nu$ = 1]{
        \label{Burger1.nu.sub.400.4}
        \includegraphics[width=0.46\textwidth]{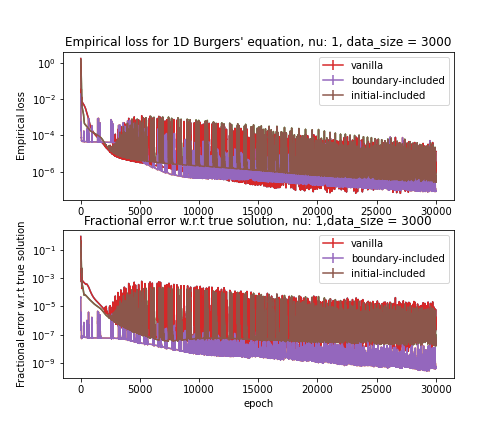}}
    \subfigure[Results for $\nu$ = 2]{
        \label{Burger1.nu.sub.400.5}
        \includegraphics[width=0.46\textwidth]{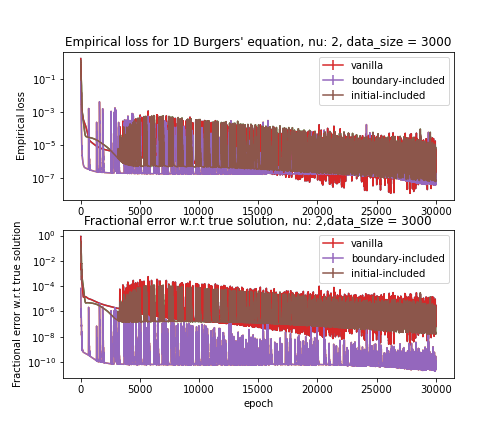}}
    \subfigure[Results for $\nu$ = 10]{
        \label{Burger1.nu.sub.400.6}
        \includegraphics[width=0.46\textwidth]{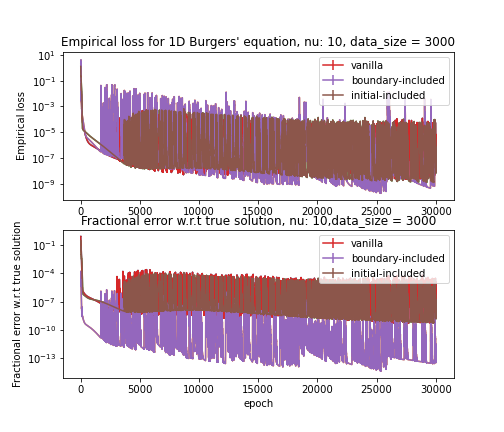}}      
    \caption{Training Progress of the Different Models for Solving 1D Burgers' PDE, Using $3 \times 10^4$ Epochs, $181801$ Parameters, lr $= 0.0001$ and Mesh Size $=3000$}
    \label{Fig.Burger1.nu.400}
\end{figure}

\begin{table}[h]
\centering
\begin{tabular}{|c|c|c|c|}
  \hline
  \multicolumn{4}{|c|}{\text { 1D Burgers', epoch}=30000, {lr}=0.0001, mesh size=3000(fractional error is given inside the brackets)} \\
  \hline
  \text { $\nu$ }& \text { \#parameters=57 } & \text { \#parameters=3441 } & \text { \#parameters=181801 }\\
  \hline
  \makecell*[c]{\text {$\nu$}=0.01} & vanilla($1.0 \times 10^{-1}$) &vanilla($3.1 \times 10^{-2}$) &vanilla ($5.2 \times 10^{-7}$)\\
  \hline
  \makecell*[c]{\text {$\nu$}=0.1} & vanilla($1.1 \times 10^{-7}$) &vanilla($7.1 \times 10^{-8}$) &boundary-included($9.6 \times 10^{-6}$)\\
  \hline
  \makecell*[c]{\text {$\nu$}=0.5} & vanilla($1.2 \times 10^{-8}$) &boundary-included($2.2 \times 10^{-9}$) &boundary-included($2.2 \times 10^{-9}$)\\
  \hline
  \makecell*[c]{\text {$\nu$}=1} & boundary-included($7.5 \times 10^{-10}$) &boundary-included($7.2 \times 10^{-10}$) &boundary-included($4.2 \times 10^{-10}$) \\
  \hline
  \makecell*[c]{\text {$\nu$}=2} & boundary-included($3.1 \times 10^{-10}$) &boundary-included($1.1 \times 10^{-9}$) &boundary-included($5.3 \times 10^{-11}$) \\
  \hline
  \makecell*[c]{\text {$\nu$}=10} & boundary-included($1.0 \times 10^{-12}$) &boundary-included($1.3 \times 10^{-10}$) &boundary-included($7.1 \times 10^{-13}$) \\
  \hline
\end{tabular}
\caption{Best Models Found for Solving the 1D Viscous Burgers' PDE at Different Values of $\nu$ and Neural Net Size}
\label{tab:Burger1.nu.best.model}
\end{table}

Table \ref{tab:Burger1.nu.best.model} summarizes the results from all our experiments for different values of viscosity parameter $\nu$ - and it reveals certain intricate insights.  {\em We realize that the boundary-included model (which was the best model at zero viscosity) continues to be the best performer for more viscous fluids and when the nets are large. But at low but non-zero viscosity it turns out that the standard PINN loss is still the best.}


\subsection{Boundary Condition Included Models Help with Burgers' PDE in 2+1 PDE}\label{Sec_2burger}
In this section, we consider Burgers' PDE, equation \ref{BurgerPDE} at $d = 2$ and $\nu = 0$. We will use $u$ to represent $u_{1}$, $v$ to represent $u_{2}$ and the governing equations can be re-written as, 
$$
\left\{\begin{array}{l}
u_{t}+u u_{x}+v u_{y}=0 \\
v_{t}+u v_{x}+v v_{y}=0
\end{array}\right.$$

The above has an interesting exact solution \citep{zhu2010numerical} with a finite time blow-up and it can be seen to correspond to the following setup on the computational domain $x, y\in [0,1]$ and $t \in[0,0.5]$ -- where we have defined ${g}_{x,0}(y, t)$ and ${g}_{x,1}(y, t)$ as the boundary conditions for $u$ at $x=0, 1$, ${g}_{y,0}(x, t)$ and ${g}_{y,1}(x, t)$ as the boundary conditions for $v$ at $y=0,1$ and $u_{0}$ and $v_{0}$ as the initial conditions for the two velocity fields. 

\begin{equation}
\label{burger2.inviscid}
\left\{\begin{array}{l}
u_{t}+u u_{x}+v u_{y}=0 \\
v_{t}+u v_{x}+v v_{y}=0
\end{array}\right.,
\left\{\begin{array}{l}
u_{0}= x+y \\
v_{0}= x-y
\end{array}\right.,
\left\{\begin{array}{l}
{g}_{x,0}(y, t)=\frac{y}{1-2\cdot t^2}\\
{g}_{x,1}(y, t)=\frac{1+y-2\cdot t}{1-2\cdot t^2}\\
{g}_{y,0}(x, t)=\frac{x}{1-2\cdot t^2}\\
{g}_{y,1}(x, t)=\frac{x-1-2\cdot t}{1-2\cdot t^2}
\end{array}\right.
\end{equation}

\paragraph{Models for Solving Two-dimensional Inviscid Burgers' PDE} Towards solving the above PDE, we consider training a neural net $\gN:\R^3 \rightarrow \R^2$, with output coordinates labeled as $({\gN_{u}},{\gN_{v}})$. Using this net we define the following  three kinds of models analogous to the previous example, 

The model where initial and boundary conditions are trained for,
\[ {\rm model_{vanilla-u}}({x,y,t}) \coloneqq \hat{u}_{v} \coloneqq {\gN_{u}}(x,y, t) \]
\[ {\rm model_{vanilla-v}}({x,y,t}) \coloneqq \hat{v}_{v} \coloneqq {\gN_{v}}(x,y, t)  \]

The model where only the boundary condition is trained for,
\[ {\rm model_{boundary-included-u}}({x,y,t}) \coloneqq \hat{u}_{b} \coloneqq {\gN_{u}}(x,y, t) \cdot x \cdot (1-x)  +  (1-x) \cdot {g}_{x,0}(y,t) + x \cdot {g}_{x,1}(y,t)\]
\[ {\rm model_{boundary-included-v}}({x,y,t}) \coloneqq \hat{v}_{b} \coloneqq {\gN_{v}}(x,y, t) \cdot y \cdot (1-y)  +  (1-y) \cdot {g}_{y,0}(x,t) + y \cdot {g}_{y,1}(x,t) \]

The model where the initial condition is trained for,
\[ {\rm model_{initial-included-u}}({x,y,t}) \coloneqq \hat{u}_{i} \coloneqq {\gN_{u}}(x,y, t) \cdot t  +  2 \cdot (\frac{1}{2}-t) \cdot u_{0}(x,y) \]
\[ {\rm model_{initial-included-v}}({x,y,t}) \coloneqq \hat{v}_{i} \coloneqq {\gN_{v}}(x,y, t) \cdot t  +  2 \cdot (\frac{1}{2}-t) \cdot v_{0}(x,y) \]

\paragraph{Population Risks for Two-dimensional Inviscid Burgers' PDE} \label{sec.2D.loss}

Let $ {u} $ and $ {v} $ be the predicted solutions for the two coordinates of the velocity. Correspondingly we define the PDE population risks, ${\gR}_{1}$  and $ {\gR}_{2}$ as follows, 
\begin{equation}
\label{Burger2-loss1}
\begin{aligned}
{\gR}_{1} &= \left\|\frac{\partial {u}}{\partial t}+{u} \frac{\partial {u}}{\partial x} + {v} \frac{\partial {u}}{\partial y}\right\|_{[0,1]^2 \times[0,1], \nu_{1}}^{2},\
{\gR}_{2} &= \left\|\frac{\partial {v}}{\partial t}+{u} \frac{\partial {v}}{\partial x} + {v} \frac{\partial {v}}{\partial y}\right\|_{[0,1]^2 \times[0,1], \nu_{1}}^{2}
\end{aligned}
\end{equation}

In the above $\nu_1$ is a measure on the whose space-time domain $[0,1]^3$. Similarly corresponding to a measure $\nu_2$ on $[0,1]^2$ (one interval being space and the other being time), we define ${\gR}_{3}$  and $ {\gR}_{4} $ corresponding to violation of the boundary conditions,
\begin{equation}
\label{Burger2-loss2}
\begin{aligned}
{\gR} _{3} &= \left\| {u} - {g}_{0,x}(y,t)\right\|_{\{0\} \times[0,1]\times[0,1], \nu_{2}}^{2} + \left\| {u} - {g}_{x,1}(y,t)\right\|_{\{1\} \times[0,1]\times[0,1], \nu_{2}}^{2}
\\
{\gR} _{4} &= \left\|{v} - {g}_{y,0}(x,t)\right\|_{[0,1]\times\{0\} \times[0,1], \nu_{2}}^{2} + \left\|{v} - {g}_{y,1}(x,t)\right\|_{[0,1]\times\{1\} \times[0,1], \nu_{2}}^{2}
\end{aligned}
\end{equation}

Similarly corresponding to a measure $\nu_3$ on the spatial volume $[0,1]^2$ we define ${\gR} _{5} $ and ${\gR} _{6}$ corresponding to the violation of initial conditions,
\begin{equation}
\label{Burger2-loss3}
{\gR} _{5} = \left\| {u} - u_{0}\right\|_{[0,1]^2 ,t=0, \nu_{3}}^{2},\
{\gR} _{6} = \left\|{v} - v_{0}\right\|_{[0,1]^2 ,t=0, \nu_{3}}^{2}
\end{equation}

Thus, in terms of the models and the risks defined above we consider algorithmically minimizing the following $3$ risks, 
$${\gR} _{\rm vanilla}=\left({\gR} _{1}({u}_{v})+{\gR}_{2}({v}_{v})\right)+\left({\gR} _{3}({u}_{v})+{\gR} _{4}({v}_{v})\right)+\left({\gR} _{5}({u}_{v})+{\gR} _{6}({v}_{v})\right)
$$
$${\gR} _{\rm boundary-included}=\left({\gR}_{1}({u}_{b})+{\gR}_{2}({v}_{b})\right)+\left({\gR}_{5}({u}_{b})+{\gR}_{6}({v}_{b})\right)
$$
$${\gR} _{\rm initial-included}=\left({\gR}_{1}({u}_{i})+{\gR}_{2}({v}_{i})\right)+\left({\gR}_{3}({u}_{i})+{\gR}_{4}({v}_{i})\right)
$$
\paragraph{Summary of Results for Solving the Two-Dimensional Inviscid Burgers' PDE with a Finite-Time Blow-Up}
In Table \ref{tab:Burger2.para}, we have listed all the $9$ experimental settings that have been tried corresponding to $3$ distinct ranges of under/overparameterization and the $3$ distinct risks as stated above. 

\begin{table}[h]
\centering
\begin{tabular}{|c|c|c|}
  \hline
  \multicolumn{3}{|c|}{\text { 2D Burgers', epoch}=30000, {lr}=0.001, p=\#parameters} \\
  \hline
  Sub-Figures of Figure \ref{Fig.Burger2.trans}& \text { \#parameters } & \text { mesh size } \\
  \hline
  \text {(a) ~[$p < n$]} & \makecell*[c]{\text 66} & \multirow{3}{*}{\makecell*[c]{\textbf{vanilla}:{ bulk mesh size}=800,\text {initial mesh size}=800,\\ \text{boundary mesh size}=800;\\ \textcolor{red}{\textbf{boundary-included}:{ bulk mesh size}=800,\text {initial mesh size}=800,}\\ \textcolor{red}{\text {boundary mesh size}=0;} \textbf{initial-included}:{ bulk mesh size} = 800,\\ \text {initial mesh size} = 0,  \text {boundary mesh size} = 800 } } \\ 
  \cline{1-2}
  \text {(b) ~[$p\approx n$]} & \makecell*[c]{\text 1794} & \\ 
  \cline{1-2}
  \text {(c) ~[$ p\geq n$]}& \makecell*[c]{\text 20802} &\\ 
  \hline
\end{tabular}
\caption{2D Inviscid Burgers' PDE Experiments with Different Settings for the Number of Trainable Parameters $p$ and $n$ Being the Mesh Size Used for the Vanilla Model. Note That in All Cases Above the Boundary Mesh Size of $800$ Comes from a Uniformly at Random Sampling of $200$ Points at Each of the Surfaces at $x=0,1$ and $y=0,1$}
\label{tab:Burger2.para}
\end{table}

\begin{figure}[H]
    \centering  %
    \subfigure[Results for the 66 Parameters Net]{
        \label{Burger2.trans.sub.1}
        \includegraphics[width=0.48\textwidth]{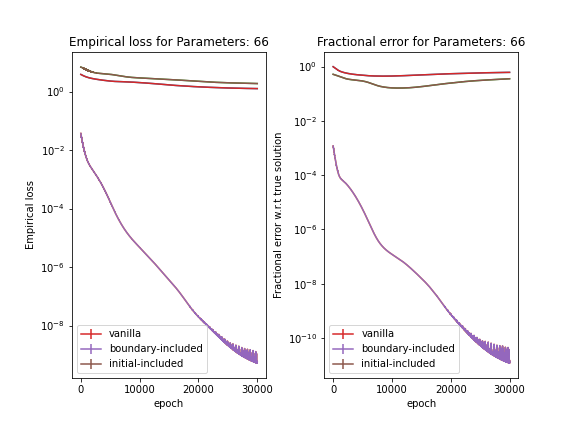}}
    \subfigure[Results for the 1794 Parameters Net]{
        \label{Burger2.trans.sub.2}
        \includegraphics[width=0.48\textwidth]{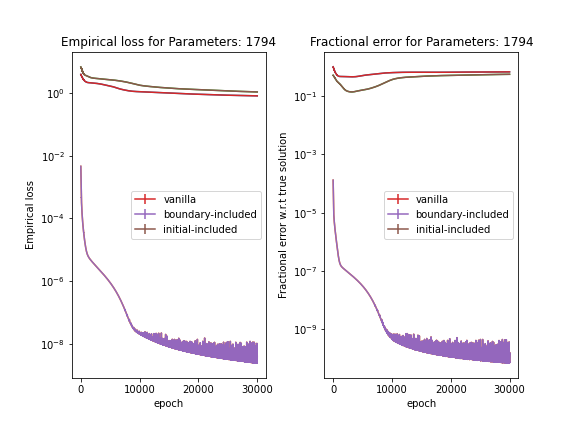}}
    \subfigure[Results for the 20802 Parameters Net]{
        \label{Burger1.trans.sub.3}
        \includegraphics[width=0.5\textwidth]{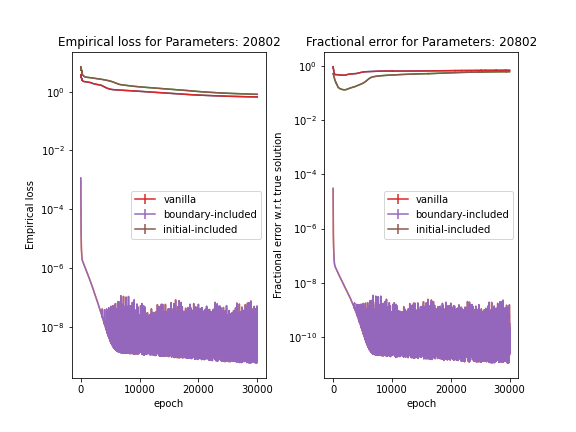}}

    \caption{Training Progress of the Different Models at Different Levels of Overparameterization When Solving the $2$D Inviscid Burgers' PDE, Using $3\times 10^{4}$ Epochs, $10^{-3}$ Learning-Rate and a Space-Time Mesh Size of $3000$}
    \label{Fig.Burger2.trans}
\end{figure}

From the above table we can conclude that  -- similar to that of 1D inviscid Burgers case -- even now the \textit{boundary-included model has the best performance.}


\subsection{Initial Condition Included Models Help with Burgers' PDE in 3+
1 PDE}\label{Sec_3burger}

The three-dimensional viscid Burgers' PDE is a partial differential equation in fluid dynamics that describes the motion of an incompressible fluid. For here,d = 3 in Burgers' PDE \ref{BurgerPDE}.
We use $u$ to represent $u_{1}$, $v$ to represent $u_{2}$, and $w$ to represent $u_{3}$,
\begin{equation}
\label{Burger3.viscid}
{\begin{array}{rl}
\frac{\partial u}{\partial t}+u \frac{\partial u}{\partial x}+v \frac{\partial u}{\partial y}+w \frac{\partial u}{\partial z} &= \nu\left(\frac{\partial^{2} u}{\partial x^{2}}+\frac{\partial^{2} u}{\partial y^{2}}+\frac{\partial^{2} u}{\partial z^{2}}\right) \\
\frac{\partial v}{\partial t}+u \frac{\partial v}{\partial x}+v \frac{\partial v}{\partial y}+w \frac{\partial v}{\partial z} &= \nu\left(\frac{\partial^{2} v}{\partial x^{2}}+\frac{\partial^{2} v}{\partial y^{2}}+\frac{\partial^{2} v}{\partial z^{2}}\right) \\
\frac{\partial w}{\partial t}+u \frac{\partial w}{\partial x}+v \frac{\partial w}{\partial y}+w \frac{\partial w}{\partial z} &= \nu\left(\frac{\partial^{2} w}{\partial x^{2}}+\frac{\partial^{2} w}{\partial y^{2}}+\frac{\partial^{2} w}{\partial z^{2}}\right)
\end{array}}
\end{equation}
On a spatial domain $\Omega$ and for $t>0$ we shall use the following exact solution for 3D viscid Burgers' PDE, \citep{shukla2016modified}, as our test case (where $\nu=\frac{1}{\operatorname{Re}}$  is the kinematic viscosity),
\[ \begin{array}{l}
{u}(x, y, z, t)=-\frac{2}{\mathrm{Re}} \left(\frac{1+ \cos (x) \sin (y) \sin (z) \exp (-t)}{1+x+\sin (x) \sin (y) \sin (z) \exp (-t)}\right) \\
{v}(x, y, z, t)=-\frac{2}{\mathrm{Re}} \left(\frac{\sin (x) \cos (y) \sin (z) \exp (-t)}{1+x+\sin (x) \sin (y) \sin (z) \exp (-t)}\right)\\
{w}(x, y, z, t)=-\frac{2}{\mathrm{Re}} \left(\frac{\sin (x) \sin (y) \cos (z) \exp (-t)}{1+x+\sin (x) \sin (y) \sin (z) \exp (-t)}\right)
\end{array}
\]
We decided to compare the precise answer to gauge how well my model predicted the bounds. We got the starting and boundary conditions required to train and test my model by applying the precise solution at the boundary locations.\\
However, the same problem, we can not directly get the boundary condition and initial condition from this paper, and we know that using the exact solution in this way is often done in deep learning-based approaches to solving partial differential equations. We can choose the computational domain as $x, y, z, t \in [0,1]$. \\
We will define $u_{0}$, $v_{0}$ and $w_{0}$ is the initial condition, and then we can get the initial condition for this Three-dimensional viscid Burgers' PDE experiment,
\begin{align*} 
&u_{0}(x, y, z)=-\frac{2}{\mathrm{Re}}\left(\frac{1+\cos (x) \sin (y) \sin (z)}{1+x+\sin (x) \sin (y) \sin (z) }\right) ,
v_{0}(x, y, z)=-\frac{2}{\mathrm{Re}} \left(\frac{\sin (x) \cos (y) \sin (z)}{1+x+\sin (x) \sin (y) \sin (z)}\right) \\
&w_{0}(x, y, z)=-\frac{2}{\mathrm{Re}} \left(\frac{\sin (x) \sin (y) \cos (z)}{1+x+\sin (x) \sin (y) \sin (z)}\right), (x,y,z) \in \Omega
\end{align*}
We will define ${g}_{x,0}(y,z,t)$ and ${g}_{x,1}(y,z,t)$ is the boundary condition for $u$ when $x=0$ and $x=1$, ${g}_{y,0}(x,z,t)$ and ${g}_{y,1}(x,z,t)$ is the boundary condition for $v$ when $y=0$ and $y=1$, ${g}_{z,0}(x,y,t)$ and ${g}_{z,1}(x,y,t)$ is the boundary condition for $w$ when $z=0$ and $z=1$, so for the boundary condition for this experiment,
\begin{align*}
{g}_{x,0}(y, z, t)=-\frac{2}{\mathrm{Re}} \left(1+  \sin (y) \sin (z) \exp (-t)\right), &~{g}_{x,1}(y, z, t)=-\frac{2}{\mathrm{Re}} \left(\frac{1}{2+\sin (1) \sin (y) \sin (z) \exp (-t)}\right)\\
{g}_{y,0}(x, z, t)=-\frac{2}{\mathrm{Re}} \left(\frac{\sin (x) \sin (z) \exp (-t)}{1+x}\right), &~{g}_{y,1}(x, z, t)=-\frac{2}{\mathrm{Re}} \left(\frac{\sin (x) \cos (1) \sin (z) \exp (-t)}{1+x+\sin (x) \sin (1) \sin (z) \exp (-t)}\right)\\
{g}_{z,0}(x, y, t)=-\frac{2}{\mathrm{Re}} \left(\frac{\sin (x) \sin (y)  \exp (-t)}{1+x}\right), &~{g}_{z,1}(x, y, t)=-\frac{2}{\mathrm{Re}} \left(\frac{\sin (x) \sin (y) \cos (1) \exp (-t)}{1+x+\sin (x) \sin (y) \sin (1) \exp (-t)}\right)
\end{align*}

\paragraph{Model for Solving the Three-Dimensional Viscid Burgers' P.D.E.}Towards solving the above PDE, we consider training a neural net $\gN:\R^4 \rightarrow \R^3$, with output coordinates labeled as $({\gN_{u}},{\gN_{v}},{\gN_{w}})$. Using this net we define the following  three kinds of models analogous to the previous example.\\
Model for training on initial and boundary,$${\rm model_{vanilla-u}}({x,y,z,t}) \coloneqq \hat{u}_{v} \coloneqq {\gN_{u}}(x,y,z, t)  $$
$${\rm model_{vanilla-v}}({x,y,z,t}) \coloneqq \hat{v}_{v} \coloneqq {\gN_{v}}(x,y,z, t)  $$
$${\rm model_{vanilla-w}}({x,y,z,t}) \coloneqq \hat{w}_{v} \coloneqq {\gN_{w}}(x,y,z, t)  $$
Model for training on the boundary,
$${\rm model_{boundary-included-u}}({x,y,z,t}) \coloneqq \hat{u}_{b} \coloneqq {\gN_{u}}(x,y,z, t) \cdot x \cdot (1-x)  +  (1-x) \cdot g_{x,0}(y,z,t) + x \cdot g_{x,1}(y,z,t)$$
$${\rm model_{boundary-included-v}}({x,y,z,t}) \coloneqq \hat{v}_{b} \coloneqq {\gN_{v}}(x,y,z, t) \cdot y \cdot (1-y)  +  (1-y) \cdot g_{y,0}(x,z,t) + y \cdot g_{y,1}(x,z,t) $$
$${\rm model_{boundary-included-w}}({x,y,z,t}) \coloneqq \hat{w}_{b} \coloneqq {\gN_{w}}(x,y,z, t) \cdot z \cdot (1-z)  +  (1-z) \cdot g_{z,0}(x,y,t) + z \cdot g_{z,0}(x,y,t) $$
Model for training on the initial,
$${\rm model_{initial-included-u}}({x,y,z,t}) \coloneqq \hat{u}_{i} \coloneqq {\gN_{u}}(x,y,z, t) \cdot t  +  \cdot (1-t) \cdot u_{0}(x,y,z)$$
$${\rm model_{initial-included-v}}({x,y,z,t}) \coloneqq \hat{v}_{i} \coloneqq {\gN_{v}}(x,y,z, t) \cdot t  +  \cdot (1-t) \cdot v_{0}(x,y,z)$$
$${\rm model_{initial-included-w}}({x,y,z,t}) \coloneqq \hat{w}_{i} \coloneqq {\gN_{w}}(x,y,z, t) \cdot t  +  \cdot (1-t) \cdot w_{0}(x,y,z)$$

\paragraph{Loss for Our Models for the Three-dimensional Viscid Burgers' PDE}

 ${\gR}{1}$,${\gR}{2}$ and $ {\gR}{3} $ that are parts of the population loss function correspond to the evaluations of differential operators,
\begin{equation}
\label{burger3.loss1}
\begin{aligned}
{\gR}_{1} &= \left\|\frac{\partial {u}}{\partial t}+{u} \frac{\partial {u}}{\partial x} + {v} \frac{\partial {u}}{\partial y} + {w} \frac{\partial {u}}{\partial z} - \nu\left(\frac{\partial^{2} {u}}{\partial x^{2}}+\frac{\partial^{2} {u}}{\partial y^{2}}+\frac{\partial^{2} {u}}{\partial z^{2}}\right) \right\|_{[0,1]^3 \times[0,1], \nu_{1}}^{2}
\\
{\gR}_{2} &= \left\|\frac{\partial {v}}{\partial t}+{u} \frac{\partial {v}}{\partial x} + {v} \frac{\partial {v}}{\partial y} + {w} \frac{\partial {v}}{\partial z} - \nu\left(\frac{\partial^{2} {v}}{\partial x^{2}}+\frac{\partial^{2} {v}}{\partial y^{2}}+\frac{\partial^{2} {v}}{\partial z^{2}}\right) \right\|_{[0,1]^3 \times[0,1], \nu_{1}}^{2}
\\
{\gR}_{3} &= \left\|\frac{\partial {w}}{\partial t}+{u} \frac{\partial {w}}{\partial x} + {v} \frac{\partial {w}}{\partial y} + {w} \frac{\partial {w}}{\partial z} - \nu\left(\frac{\partial^{2} {w}}{\partial x^{2}}+\frac{\partial^{2} {w}}{\partial y^{2}}+\frac{\partial^{2} {w}}{\partial z^{2}}\right) \right\|_{[0,1]^3 \times[0,1], \nu_{1}}^{2}
\\
\end{aligned}
\end{equation}
$ {\gR}_{4}$,${\gR}_{5}$  and $ {\gR}_{6} $ correspond to the evaluation of boundary conditions,
\begin{equation}
\label{burger3.loss2}
\begin{aligned}
{\gR}_{4} &= \left\| {u} - {g}_{x,0}(y,z,t)\right\|_{\{0\} \times[0,1]^2 \times[0,1], \nu_{2}}^{2} + \left\| {u} - {g}_{x,1}(y,z,t)\right\|_{\{1\} \times[0,1]^2 \times[0,1], \nu_{2}}^{2}
\\
{\gR}_{5} &= \left\|{v} - {g}_{y,0}(x,z,t)\right\|_{[0,1] \times\{0\} \times[0,1] \times[0,1], \nu_{2}}^{2} + \left\|{v} - {g}_{y,1}(x,z,t)\right\|_{[0,1] \times\{1\} \times[0,1] \times[0,1], \nu_{2}}^{2}
\\
{\gR}_{6} &= \left\| {w} - {g}_{z,0}(x,y,t)\right\|_{[0,1]^2 \times \{0\} \times[0,1], \nu_{2}}^{2} + \left\| {w} - {g}_{z,1}(x,y,t)\right\|_{[0,1]^2 \times \{1\} \times[0,1], \nu_{2}}^{2}
\end{aligned}
\end{equation}
$ {\gR}_{7} $,${\gR}_{8}$ and $ {\gR}_{9} $ correspond to the evaluation of initial conditions,
\begin{equation}
\label{burger3.loss3}
\begin{aligned}
{\gR}_{7} &= \left\| {u} - u_{0}\right\|_{[0,1]^3 ,t=0, \nu_{3}}^{2},\
{\gR}_{8} &= \left\|{v} - v_{0}\right\|_{[0,1]^3 ,t=0, \nu_{3}}^{2},\
{\gR}_{9} &= \left\| {w} - w_{0}\right\|_{[0,1]^3 ,t=0, \nu_{3}}^{2}
\end{aligned}
\end{equation}
where $ {u} $,$ {v} $ and $ {w} $ are approximate solutions predicted as an output of DNN. \\
Loss for training on initial and boundary, {\bf ${u}_{v}$, ${v}_{v}$ and ${w}_{v}$ are the prediction for vanilla model},
$${\gR}_{\rm vanilla}=\left({\gR}_{1}({u}_{v})+{\gR}_{2}({v}_{v})+{\gR}_{3}({w}_{v})\right)+\left({\gR}_{4}({u}_{v})+{\gR}_{5}({v}_{v})+{\gR}_{6}({w}_{v})\right)+\left({\gR}_{7}({u}_{v})+{\gR}_{8}({v}_{v})+{\gR}_{9}({w}_{v})\right)
$$
Loss for training on the boundary, {\bf ${u}_{b}$, ${v}_{b}$ and ${w}_{b}$ are the prediction for boundary-included model},
$${\gR}_{\rm boundary-included}=\left({\gR}_{1}({u}_{b})+{\gR}_{2}({v}_{b})+{\gR}_{3}({w}_{b})\right)+\left({\gR}_{7}({u}_{b})+{\gR}_{8}({v}_{b})+{\gR}_{9}({w}_{b})\right)
$$
Loss for training on the initial, {\bf ${u}_{i}$, ${v}_{i}$ and ${w}_{i}$ are the prediction for initial-included model},
$${\gR}_{\rm initial-included}=\left({\gR}_{1}({u}_{i})+{\gR}_{2}({v}_{i})+{\gR}_{3}({w}_{i})\right)+\left({\gR}_{4}({u}_{i})+{\gR}_{5}({v}_{i})+{\gR}_{6}({w}_{i})\right)
$$
\paragraph{Performances of the Different Models at Different Values of Viscosity}
The two tables (Table \ref{tab:Burger3.nu.4} and Table \ref{tab:Burger3.nu.40}) show the results of experiments conducted on the 3D viscid Burgers' PDE with different values of $\nu$ and different models. 
For the hyper-parameters here, a fixed number of epochs (10000), learning rate
(0.001), mesh size (3000), and the sizes of the neural network, we can design experiments to investigate how changing the value of $\nu$ affects the model's performance. It is to be noted that the boundary
mesh will have $166$ points for $x=0,1$, $166$ points for $y=0,1$ and $170$ points for $z = 0,1$.

\begin{table}[h]
\centering
\begin{tabular}{|c|c|c|}
  \hline
  \multicolumn{3}{|c|}{\text { 3D Burgers', epoch }=10000, {lr}=0.001, \#parameters=75} \\
  \hline
  Sub-Figures of Figure \ref{Fig.Burger3.nu.4}& \text { $\nu$ } & \text { model } \\
  \hline
  \text {(a) Experiment 1} & \makecell*[c]{\text {$\nu$}=0.01} & \multirow{6}{*}{\makecell*[c]{\textbf{vanilla}:{ bulk mesh size}=1000,\text {initial mesh size}=1000, \text{boundary mesh size}=1000 \\ \textbf{boundary-included}:{ bulk mesh size} = 1000,\text {initial mesh size} = 1000, \\ \text {boundary mesh size} = 0,\\ \textbf{initial-included}:{ bulk mesh size} = 1000,\text {initial mesh size} = 0, \\ \text {boundary mesh size} = 1000 } } \\ 
  \cline{1-2}
  \text {(b) Experiment 2} & \makecell*[c]{\text {$\nu$}=0.1} & \\ 
  \cline{1-2}
  \text {(c) Experiment 3}& \makecell*[c]{\text {$\nu$}=0.5} &\\ 
  \cline{1-2}
  \text {(d) Experiment 4}& \makecell*[c]{\text {$\nu$}=1} & \\ 
  \hline
\end{tabular}
\caption{3D Viscid Burgers' PDE Experiments with Different Values of $\nu$ and Number of Trainable Parameters $=75$}
\label{tab:Burger3.nu.4}
\end{table}
\begin{table}[h]
\centering
\begin{tabular}{|c|c|c|}
  \hline
  \multicolumn{3}{|c|}{\text { 3D Burgers', epoch }=10000, {lr}=0.001, \#parameters=3603} \\
  \hline
  Sub-Figures of Figure \ref{Fig.Burger3.nu.40}& \text { $\nu$ } & \text { model } \\
  \hline
  \text {(a) Experiment 1} & \makecell*[c]{\text {$\nu$}=0.01} & \multirow{6}{*}{\makecell*[c]{\textbf{vanilla}:{ bulk mesh size}=1000,\text {initial mesh size}=1000, \text{boundary mesh size}=1000 \\ \textbf{boundary-included}:{ bulk mesh size} = 1000,\text {initial mesh size} = 1000, \\ \text {boundary mesh size} = 0,\\ \textbf{initial-included}:{ bulk mesh size} = 1000,\text {initial mesh size} = 0, \\ \text {boundary mesh size} = 1000 } } \\ 
  \cline{1-2}
  \text {(b) Experiment 2} & \makecell*[c]{\text {$\nu$}=0.1} & \\ 
  \cline{1-2}
  \text {(c) Experiment 3}& \makecell*[c]{\text {$\nu$}=0.5} &\\ 
  \cline{1-2}
  \text {(d) Experiment 4}& \makecell*[c]{\text {$\nu$}=1} & \\ 
  \hline
\end{tabular}
\caption{3D Viscid Burgers' PDE Experiments with Different Values of $\nu$ and Number of Trainable Parameters $= 3603$}
\label{tab:Burger3.nu.40}
\end{table}

\begin{figure}[H]
    \centering  %
    \subfigure[Results for $\nu$ = 0.01]{
        \label{Burger3.nu.sub.4.1}
        \includegraphics[width=0.48\textwidth]{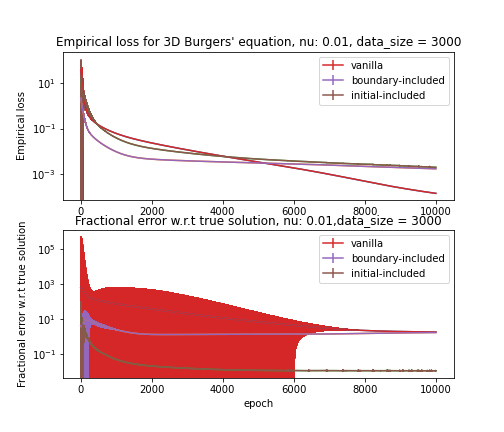}}
    \subfigure[Results for $\nu$ = 0.1]{
        \label{Burger3.nu.sub.4.2}
        \includegraphics[width=0.48\textwidth]{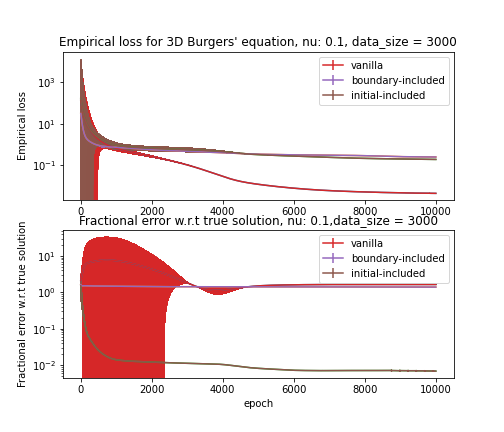}}
    \subfigure[Results for $\nu$ = 0.5]{
        \label{Burger3.nu.sub.4.3}
        \includegraphics[width=0.48\textwidth]{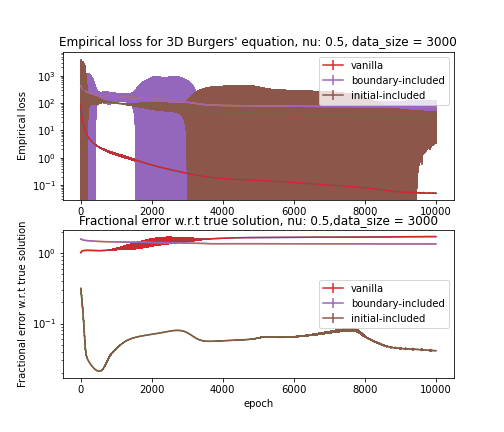}}
    \subfigure[Results for $\nu$ = 1]{
        \label{Burger3.nu.sub.4.4}
        \includegraphics[width=0.48\textwidth]{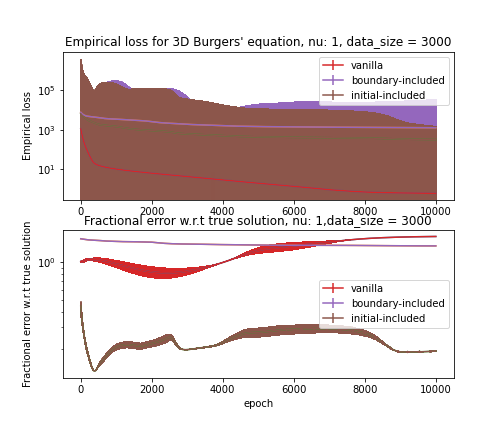}}
    \caption{Results for Experiments with 3D Burgers' PDE for Different $\nu$ and \#parameters $=75$.}
    \label{Fig.Burger3.nu.4}
\end{figure}

\begin{figure}[H]
    \centering  %
    \subfigure[Results for $\nu$ = 0.01]{
        \label{Burger3.nu.sub.40.1}
        \includegraphics[width=0.48\textwidth]{Images/Burger-Images/3Dburger/40/loss_epoch_3D_30000_0-01_0.0001_3000.png}}
    \subfigure[Results for $\nu$ = 0.1]{
        \label{Burger3.nu.sub.40.2}
        \includegraphics[width=0.48\textwidth]{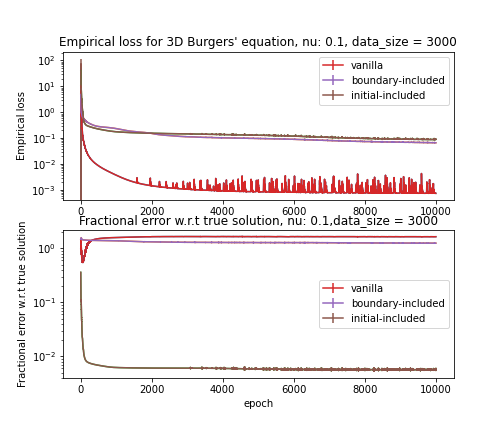}}
    \subfigure[Results for $\nu$ = 0.5]{
        \label{Burger3.nu.sub.40.3}
        \includegraphics[width=0.48\textwidth]{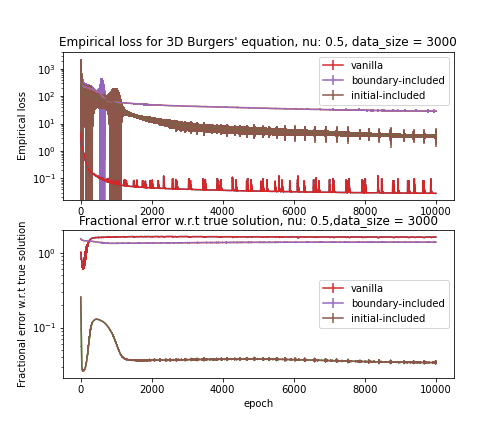}}
    \subfigure[Results for $\nu$ = 1]{
        \label{Burger3.nu.sub.40.4}
        \includegraphics[width=0.48\textwidth]{Images/Burger-Images/3Dburger/40/loss_epoch_3D_30000_1_0-0001_3000.png}}
    \caption{Results for Experiments with 3D Burgers' PDE for Different $\nu$ and \#parameters $=3603$.}
    \label{Fig.Burger3.nu.40}
\end{figure}

\begin{table}[h]
\centering
\begin{tabular}{|c|c|c|}
  \hline
  \multicolumn{3}{|c|}{\text { 3D Burgers', epoch }=10000, {lr}=0.001(inside the brackets is error)} \\
  \hline
  \text { $\nu$ }& \text { \#parameters=75 } & \text { \#parameters=3603 } \\
  \hline
  \makecell*[c]{\text {$\nu$}=0.01} & initial-included($1.1 \times 10^{-2}$) &initial-included($9.8 \times 10^{-3}$)\\
  \hline
  \makecell*[c]{\text {$\nu$}=0.1} & initial-included($6 \times 10^{-3}$) &initial-included($5.1 \times 10^{-3}$) \\
  \hline
  \makecell*[c]{\text {$\nu$}=0.5} & initial-included($4 \times 10^{-2}$) &initial-included($3.4 \times 10^{-2}$) \\
  \hline
  \makecell*[c]{\text {$\nu$}=1} & initial-included($1.9 \times 10^{-1}$) &initial-included($1.8 \times 10^{-1}$) \\
  \hline
\end{tabular}
\caption{Errors of the Best Model for Solving 3D Viscous Burgers' PDE for Different Values of $\nu$ and Number of Parameters }
\label{tab:Burger3.nu.best.model}
\end{table}

From the two tables, we can observe that as the viscosity coefficient $\nu$ increases, the fractional error will increase, which indicates a decrease in model performance. This trend is consistent for both values of the parameter tested.\\
For the 3D viscous Burgers' PDE, different from 1D viscous Burgers' PDE, we can see that the initial-included mode configuration consistently performs better than the other two models, vanilla and boundary-included.\\ 
Moreover, we can see that the error decreases when the parameter number is increased from $75$ to $3603$, for all values of $\nu$. However, the progress is not very significant and the result suggests that the model's performance is not very sensitive to changes in the width of the neural network.

\subsection{Reflective Summary of Solving the Burgers' PDE via PINN Modifications} 
The following tree represents all the models tried in this segment. In brackets we have stated the kind of model that performed the best for the corresponding experiment.

\begin{forest}
  for tree={
    grow'=0,
    parent anchor=east,
    child anchor=west,
    edge path={
      \noexpand\path[\forestoption{edge}]
      (!u.parent anchor) -- +(5mm,0) |- (.child anchor)\forestoption{edge label};
    },
    base=bottom,
    anchor=west,
    align=left,
    l sep=7mm,
    s sep=3mm,
  }
      [Inviscid Burgers' 
        [1D Inviscid Burgers'
            [{\#parameters=57(boundary-included)}]
            [{\#parameters=3441(boundary-included)}]
            [{\#parameters=81201(boundary-included)}]
            [{\#parameters=181801(boundary-included)}]
        ]
        [2D Inviscid Burgers'
            [{\#parameters=66}(boundary-included)]
            [{\#parameters=1794}(boundary-included)]
            [{\#parameters=20802}(boundary-included)]
        ]]  
\end{forest}

\begin{forest}
  for tree={
    grow'=0,
    parent anchor=east,
    child anchor=west,
    edge path={
      \noexpand\path[\forestoption{edge}]
      (!u.parent anchor) -- +(5mm,0) |- (.child anchor)\forestoption{edge label};
    },
    base=bottom,
    anchor=west,
    align=left,
    l sep=7mm,
    s sep=3mm,
  }
      [\ \ Viscid Burgers'
        [1D Viscid Burgers'
            [{\#parameters=57 with different $\nu$ (boundary-included)}]
            [{\#parameters=3441 with different $\nu$ (boundary-included)}]
            [{\#parameters=81201 with different $\nu$ (boundary-included)}]
            [{\#parameters=181801 with different $\nu$ (boundary-included)}]
        ]
        [3D Viscid Burgers'
            [{\#parameters=75 with different $\nu$ (initial-included)}]
            [{\#parameters=3603 with different $\nu$(initial-included)}]
        ]
    ]  
\end{forest}



From above we note that for \textit{3D viscid Burgers' PDE, the best performing models are the initial-included ones, unlike at lower dimensions.} From the all the data given in this appendix we can also note that \textit{for the 1D viscous Burgers' PDE, as the viscosity increases, the accuracy increases. But for the 3D viscous Burgers' PDE, when the viscosity increases, the accuracy of solving the PDE decreases.}

We believe that both the patterns uncovered above suggest exciting directions for future research. 




%% file: Sections/KdV.tex
\section{A Study of PINN Modifications for the KdV PDE} \label{Sec_KdV_exp}

We attempt to numerically solve the KdV by depth $4$ feedforward neural nets mapping, $\gN : \mathbb{R}^2 \xrightarrow{} \mathbb{R}$ as, ${\gN}(x, t) := {\rm A}_{\rm 4}(\sigma{\rm A}_3({\sigma}{\rm A}_2({\sigma}{\rm A}_{\rm 1}(x, t))))$. But we shall study various choices of (uniform) width to probe the effect of having the model be under- or over- parameterized w.r.t the number of collocation points.  

The boundary and initial condition included modes we consider for the solution, in terms of the above net are as follows, 

\begin{equation}\label{KdV_models}
\begin{aligned}
    &f_{\rm {vanilla}}(x, t) \coloneqq f_v(x,t) := {\gN}(x, t), \\
    &f_{\rm {boundary-included}}(x, t) \coloneqq f_b(x,t) := {\gN(x, t)} \cdot \frac{- a + x}{b - a} \cdot \frac{b - x}{b - a} + \frac{b - x}{b - a} \cdot g_{a}(a, t) + \frac{- a + x}{b - a} \cdot g_{b}(b, t), \\
    &f_{\rm {initial-included}}(x, t) \coloneqq f_i(x,t) := {\gN(x, t)} \cdot \frac{t^2}{t^2 + q} + \frac{q}{t^2 + q} \cdot u_0(x, 0),\\
\end{aligned}
\end{equation}

where $[a, b] = \Omega$ defined in equation \ref{KdV} and $q$ in $f_{\rm {initial-included}}$ would eventually be chosen to be a very small number. It is clear from the above that $f_{\rm {boundary-included}}(x, t)$ is designed to automatically satisfy the boundary conditions while $f_{\rm {initial-included}}(x, t)$ satisfies the initial conditions as designed. And our experiments shall be designed to decide which of these perform the best.  For these models, the corresponding empirical risk functions they are trained on are,

\begin{equation} \label{KdV_empirical_risk}
    \begin{aligned}
        &{\rm \hat{\gR}_{vanilla}} \coloneqq \frac{1}{\mathcal{D}_u}\sum_{p=1}^{\mathcal{D}_u}|\pdv{f_{v}}{t} + \pdv[3]{f_{v}}{x} + 6f_{v}\pdv{f_{v}}{x}|^2_p + \frac{1}{\mathcal{D}_b}\sum_{q=1}^{\mathcal{D}_b}|f_{v} - g|^2_{q} + \frac{1}{\mathcal{D}_0}\sum_{r=1}^{\mathcal{D}_0}|f_{v} - u_0|^2_{r}, \\
        &{\rm \hat{\gR}_{boundary-included}} \coloneqq \frac{1}{\mathcal{D}_u}\sum_{p=1}^{\mathcal{D}_u}|\pdv{f_{b}}{t} + \pdv[3]{f_{b}}{x} + 6f_{b}\pdv{f_{b}}{x}|^2_p + \frac{1}{\mathcal{D}_0}\sum_{r=1}^{\mathcal{D}_0}|f_{b} - u_0|^2_{r}, \\
        &{\rm \hat{\gR}_{initial-included}} \coloneqq \frac{1}{\mathcal{D}_u}\sum_{p=1}^{\mathcal{D}_u}|\pdv{f_{i}}{t} + \pdv[3]{f_{i}}{x} + 6f_{i}\pdv{f_{i}}{x}|^2_p + \frac{1}{\mathcal{D}_b}\sum_{q=1}^{\mathcal{D}_b}|f_{i} - g|^2_{q}, 
    \end{aligned}
\end{equation}

where $\mathcal{D}_u$, $\mathcal{D}_b$, $\mathcal{D}_0$ denote the number of points set sampled from the space-time bulk, the boundary and the $t=0$ spatial slice. The notation of $|\pdv{f_{v}}{t} + \pdv[3]{f_{v}}{x} + 6f_{v}\pdv{f_{v}}{x}|^2_{p}$ is short-hand for  $\left ( \left ( \pdv{f_{v}}{t}  + \pdv[3]{f_{v}}{x} + 6f_{v}\pdv{f_{v}}{x} \right ) \mid_{(x_p,t_p)} \right )^2$ where $(x_p,t_p)$ is the $p-$th point sampled in the space-time bulk. And similarly for the other terms. 

All nets shall use the sigmoid activation function and the neural net shall be initialized via the  Xavier normalization method and optimized via the Adam optimizer with a learning rate of $10^{-4}$ and no explicit regularization shall be applied.

\subsection{Solving for a 1 Soliton Solution of the KdV}\label{Sec_1Soliton}


The computational domain is selected as $(x,t) \in [0,1]^2$, equivalent to setting $a=0$ and $b=1$. Furthermore, $q=10^{-9}$ is determined for $f_i$ in equation \ref{KdV_models} via hyperparameter optimization. According to the past work \citep{soliton_cambridge}, a valid $N-$soliton solution arises from an initial condition of $u_0(x) = N(N+1)\sech^2(x)$.
\[ u_0(x) = 2\sech^2(x) \] 

\[
 g(x, t)=\left\{
    \begin{aligned}
    u(0, t), &x=0 \\
    u(1, t), &x=1, 
    \end{aligned}
    \right.
\]
   
where $u$ denotes the exact single soliton solution \ref{1_soliton_sol}. Recall that boundary conditions need to be set to solve in the finite domain and we can reverse engineer them from the known solitonic solution against which we shall be testing our performances.

To assess the risk functions in equation \ref{KdV_empirical_risk}, we randomly sampled 500 points in the space-time bulk, allocating 125 points each at $x=0$ and $x=1$, and 250 points at $t=0$. The fractional error relative to the true solution was calculated using a uniform sample of 10,201 points within the domain. It's crucial for a thorough accuracy test and generalization that the testing set size significantly surpasses the training set size, as chosen in this study. To account for statistical variations, each model underwent multiple training cycles, specifically three iterations in this study. The variance in behaviors is represented alongside average performance metrics.



\subsubsection{Results}



\begin{table}[H]
    \centering
    \begin{tabular}{|c|c|c|c|c|c|}
		\hline
            \multicolumn{6}{|c|}{\text { 1-Soliton Solution of KdV (Epoch =60000, Repeats=3, Test Mesh Size=10201)}} \\
            \hline
		Number of Parameters $(p)$ & Mesh Size $(n)$ & $p/n \approx$ & Model & Empirical Risk & Fractional Error \\
		\hline
		\multirow{3}*{57} & \multirow{12}*{\shortstack{$\mathcal{D}_u = 500$,\\$\mathcal{D}_b = \mathcal{D}_0 = 250$}} & \multirow{3}*{0.05} & vanilla & $1.20\times 10^{-2}$ & $1.05 \times 10^{-2}$  \\
            \cline{4-6}
                &  & & boundary-included & $8.04 \times 10^{-3}$ & $3.24 \times 10^{-2}$  \\
            \cline{4-6}
                &  & & \textcolor{red}{initial-included} & $1.24 \times 10^{-5}$ & \textcolor{red}{$1.39 \times 10^{-6}$}  \\
            \cline{1-1}
            \cline{3-6}
                \multirow{3}*{541} &   & \multirow{3}*{0.5} & vanilla & $4.06 \times 10^{-5}$ & $2.50 \times 10^{-4}$ \\
            \cline{4-6}
                &  & & boundary-included & $7.27 \times 10^{-4}$ & $2.29 \times 10^{-2}$  \\
            \cline{4-6}
                &  & & initial-included & $4.51 \times 10^{-6}$ & $6.22 \times 10^{-6}$  \\
            \cline{1-1}
            \cline{3-6}
                \multirow{3}*{1009} &   & \multirow{3}*{1} & vanilla & $1.58 \times 10^{-4}$ & $8.57 \times 10^{-3}$ \\
            \cline{4-6}
                &  & & boundary-included & $1.77 \times 10^{-4}$ & $8.99 \times 10^{-3}$  \\
            \cline{4-6}
                &  & & initial-included & $3.72 \times 10^{-6}$ & $6.60 \times 10^{-5}$  \\
            \cline{1-1}
            \cline{3-6}
                \multirow{3}*{1981} &   & \multirow{3}*{2} & vanilla & $1.29 \times 10^{-5}$ & $3.00 \times 10^{-4}$ \\
            \cline{4-6}
                &  & & boundary-included & $2.89 \times 10^{-5}$ & $1.53 \times 10^{-3}$  \\
            \cline{4-6}
                &  & & initial-included & $3.49 \times 10^{-6}$ & $1.63 \times 10^{-5}$  \\
            \hline
	\end{tabular}
    \caption{Comparison of Results of the Different PINN Models Attempting on the 1-Soliton KdV Solution}
    \label{tab:KdV1}
\end{table}




\begin{figure}[H]
    \centering  %
    \subfigure[Results for the 54 Parameters Net]{
        \label{KdV1.sub.1}
        \includegraphics[width=0.44\textwidth]{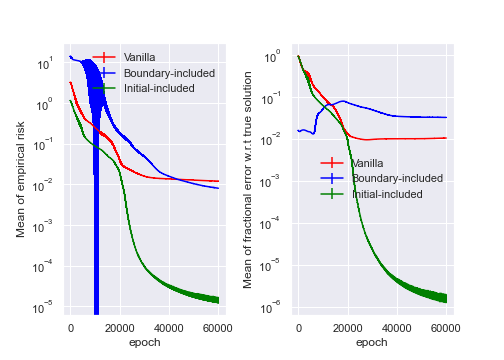}}
    \subfigure[Results for the 541 Parameters Net]{
        \label{KdV1.sub.2}
        \includegraphics[width=0.44\textwidth]{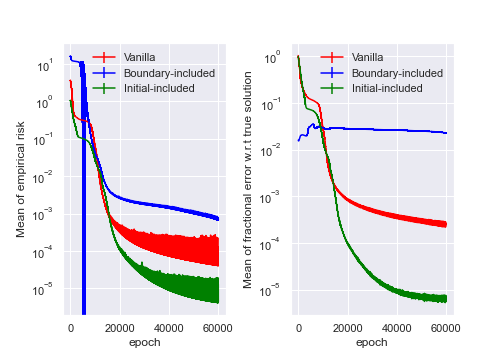}}
    \subfigure[Results for the 1009 Parameters Net]{
        \label{KdV1.sub.3}
        \includegraphics[width=0.44\textwidth]{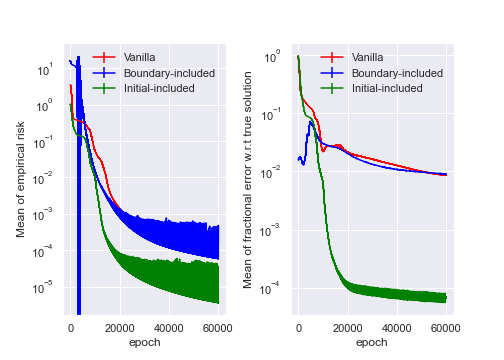}}
    \subfigure[Results for the 1981 Parameters Net]{
        \label{KdV1.sub.4}
        \includegraphics[width=0.44\textwidth]{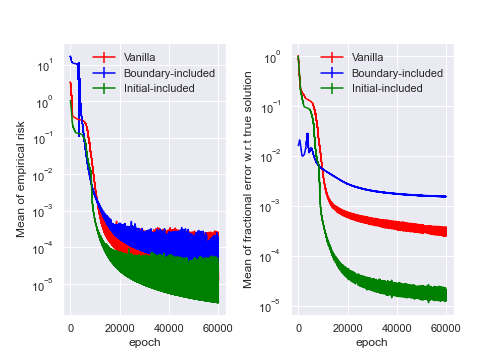}}
    \caption{Computational Results of 1-Soliton Solution KdV with Different Size Neural Nets}
    \label{Fig.KdV_models}
\end{figure}

We conducted a fair comparison by uniformly training each model for 60,000 epochs. Table \ref{tab:KdV1} presents the empirical risk and fractional error concerning the true solution at the final epoch for each model, differing by neural nets and $p/n$ ratios. Here, $p$ represents the model's trainable parameters, and $n$ indicates collocation points.

The table shows 12 neural nets trained, with the minimum fractional error (highlighted in red) being $1.39 \times 10^{-6}$. Interestingly, this was achieved by the initial-condition-included model with the smallest $p/n$ ratio magnitude, models incorporating initial conditions consistently outperformed others.

Figure \ref{Fig.KdV_models}  gives a visualization of the trends in results across epochs, for training these different models - along with the error bars.

A consistent feature observed across all plots - {\em that is not captured by the table above} - is that for running the training below $5000$ epochs, the boundary-included model (blue) exhibits the optimal performance. Thus we realize that these modified models can have interesting trade-offs between accuracy and time. Though, the initial-included model's performance improves rapidly in the later epochs and despite this model's higher variance, prolonged training times yield significant performance improvements. 



\begin{figure}[H]
    \centering
    \includegraphics[width=1\textwidth,height=0.8\textwidth]{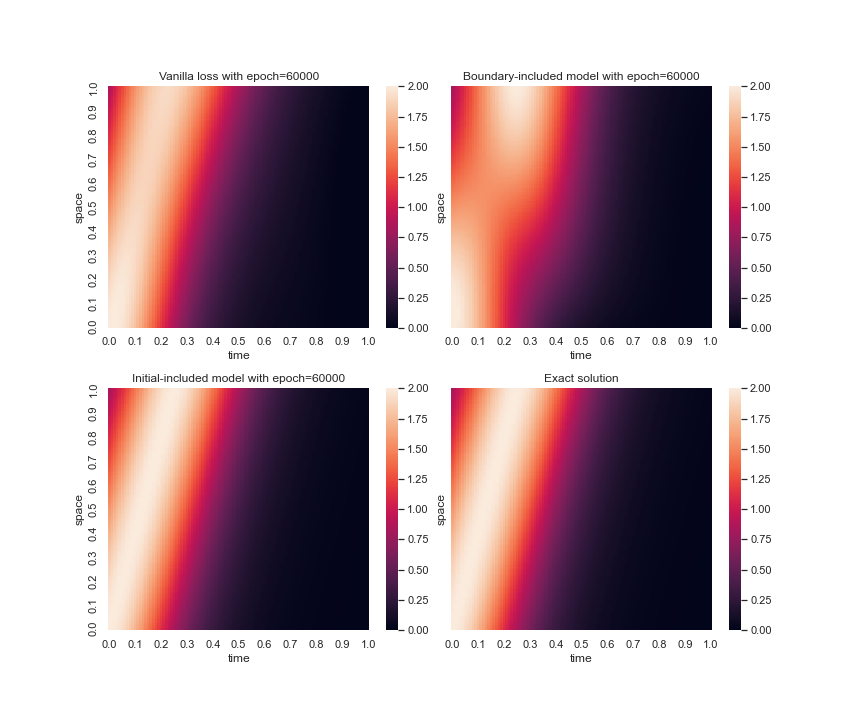}
    \caption{Heat Map of the 1-Soliton Solution Using 3 Models with 57 Parameters and the Exact Solution: Left Top: Vanilla Model; Right Top: Boundary-included Model; Left Bottom: Initial-included Model; Right Bottom: Exact Solution}
    \label{KdV1.heatmap}
\end{figure}

In summary, Figure \ref{KdV1.heatmap} visually represents the three PINN models' ability to simulate the KdV equation for a single soliton solution.




\subsection{Solving for a 2 Soliton Solution of the KdV}\label{Sec_2Soliton}


We recall the $2-$soliton solution in equation \ref{2_soliton_sol}, that we shall test against and we choose a larger computational domain of $(x,t) \in [-5,5]\times [-1,1]$ for this test than in the $1-$soliton test in the previous subsection. Also, a larger set of collocation points are sampled i.e $4000$ points in the space-time bulk, $2000$ points at $t=0$, $1000$ points each at $x= \pm 5$. Also, a larger size of the testing mesh has been chosen i.e $301 \times 201$ points uniformly distributed within the domain. We continue with the choice of $q=10^{-9}$ for the $f_i$ model in equation \ref{KdV_models} and the empirical risk functions continue to be as given in equation \ref{KdV_empirical_risk}.









As required for the setup in equation \ref{KdV_empirical_risk}, we can infer the required initial condition and the boundary conditions from the true solution equation \ref{2_soliton_sol}  as,

\[ u_0(x) = 6\sech^2(x), t=0 \]
\[
g(x, t)=\left\{
    \begin{aligned}
    &u(-5, t), &x=-5 \\
    &u(5, t), &x=5, 
    \end{aligned}
    \right.
\]

\begin{figure}[H]
    \centering  %
    \subfigure[Results for the 417 Parameters Net]{
        \label{KdV2.sub.1}
        \includegraphics[width=0.44\textwidth]{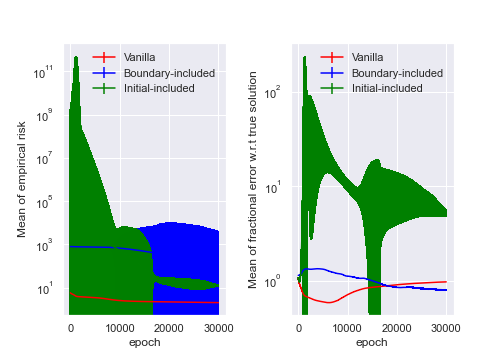}}
    \subfigure[Results for the 2109 Parameters Net]{
        \label{KdV2.sub.2}
        \includegraphics[width=0.44\textwidth]{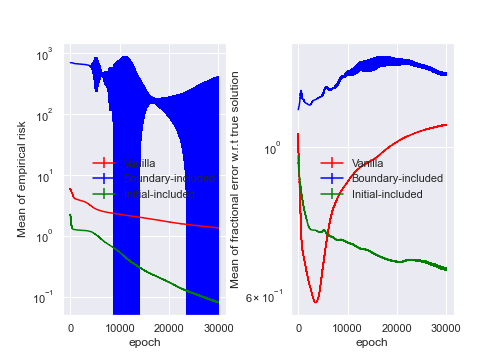}}
    \subfigure[Results for the 4137 Parameters Net]{
        \label{KdV2.sub.3}
        \includegraphics[width=0.44\textwidth]{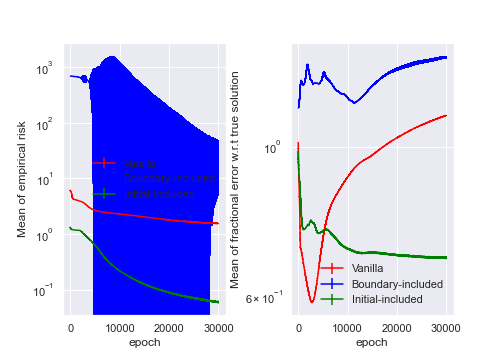}}
    \subfigure[Results for the 10221 Parameters Net]{
        \label{KdV2.sub.4}
        \includegraphics[width=0.44\textwidth]{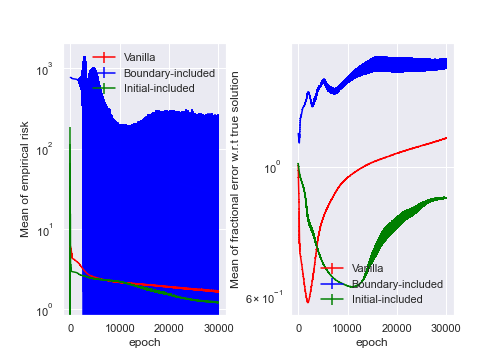}}

    \caption{Computational Results of 2-Soliton Solution KdV with Different Size Neural Nets}
    \label{Fig.KdV2_models}
\end{figure}
\paragraph{Results}

As in the previous section we again checked all three models at various values of $p/n$ and we report the results in table \ref{tab:KdV2} and the dynamics of training is shown in figure \ref{Fig.KdV2_models}. Note that all plots of performance metrics show the variance with respect to four repeats of the training. 


\begin{table}[H]
    \centering
    \begin{tabular}{|c|c|c|c|c|c|}
		\hline
            \multicolumn{6}{|c|}{\text { 2-Soliton Solution of KdV (Epoch = 30000, Repeats = 3, Test Mesh Size=60000)}} \\
            \hline
		Number of Parameters (p) & Mesh Size(n) & p / n $\approx$ & Model & Empirical Risk & Fractional Error \\
		\hline
		\multirow{3}*{417} & \multirow{12}*{\shortstack{$\mathcal{D}_u = 2000$,\\$\mathcal{D}_b = \mathcal{D}_0 = 1000$}} & \multirow{3}*{0.1} & vanilla & 2.05 & $9.74 \times 10^{-1}$ \\
            \cline{4-6}
                &  & & boundary-included & $1.44e \times 10^{2}$ & $7.98 \times 10^{-1}$  \\
            \cline{4-6}
                &  & & initial-included & $1.25e \times 10^{1}$ & 5.16  \\
            \cline{1-1}
            \cline{3-6}
                \multirow{3}*{2109} &   & \multirow{3}*{0.5} & vanilla & 1.35 & 1.08  \\
            \cline{4-6}
                &  & & boundary-included & $9.70 \times 10^{11}$ & 1.29 \\
            \cline{4-6}
                &  & & \textcolor{red}{initial-included} & $8.04 \times 10^{-2}$ & \textcolor{red}{$6.56 \times 10^{-1}$} \\
            \cline{1-1}
            \cline{3-6}
                \multirow{3}*{4137} &   & \multirow{3}*{1} & vanilla & 1.54 & 1.11 \\
            \cline{4-6}
                &  & & boundary-included & $2.71 \times 10^{1}$ & 1.36  \\
            \cline{4-6}
                &  & & initial-included & $5.84 \times 10^{-2}$ & $6.84 \times 10^{-1}$  \\
            \cline{1-1}
            \cline{3-6}
                \multirow{3}*{10221} &   & \multirow{3}*{2.5} & vanilla & 1.65 & 1.12 \\
            \cline{4-6}
                &  & & boundary-included & $2.76 \times 10^{1}$ &  1.50 \\
            \cline{4-6}
                &  & & initial-included & 1.21 &  $8.84 \times 10^{-1}$ \\
            \hline
	\end{tabular}
    \caption{Comparion of Results of the Different PINN Models Attempting to Solve for the 1-Soliton KdV Solution}
    \label{tab:KdV2}
\end{table}

Of the $12$ different neural networks trained, in Table \ref{tab:KdV2}, the figure highlighted in red represents $6.56 \times 10^{-1}$ is the best performance and that was again obtained from the initial-included model and at $p/n <1$- as for the single-soliton experiment.

\begin{figure}[H]
    \centering
    \includegraphics[width=1\textwidth,height=0.65\textwidth]{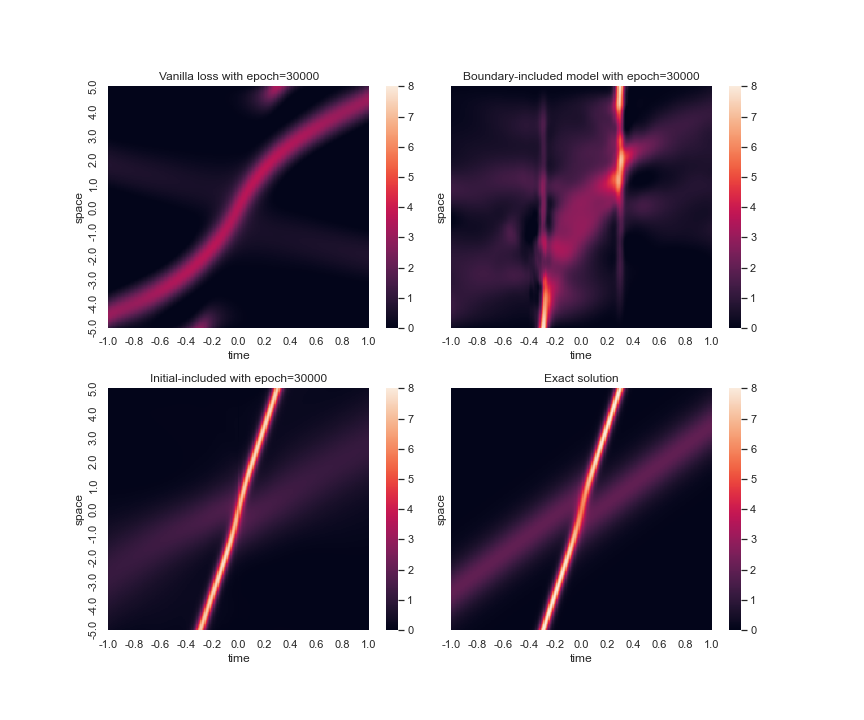}
    \caption{Heat Map of the 2-Soliton Solution Obtained Using All the Models with 2109 Parameters. Left Top: Vanilla Model; Right Top: Boundary-included Model; Left Bottom: Initial-included Model; Right Bottom: Exact Solution}
    \label{KdV2.heatmap}
\end{figure}

As in the experiments with the $1-$soliton solution in figure  \ref{Fig.KdV2_models} we can see a accuracy-time trade-off show up - that when run for less epochs the standard PINN loss (red) does work the best. However, eventually, the initial-included model, emerges as the best model for all but the smallest neural net size in Figure \ref{KdV2.sub.1} - where the boundary-included model dominated the performance. 

All figures reveal an intriguing pattern where the fractional error of the standard PINN loss/vanilla model initially decreases before rapidly increasing post $5000$ epochs. Despite a continual decrease in empirical risk, the final fractional error often surpasses the initial error. {\em This pattern suggests a potential overfitting tendency in the standard PINN loss, emphasizing the need for meticulous early-stopping implementation for reliable use.}

In summary, the results of our experiments with the double-soliton solution have been encapsulated in Figure \ref{KdV2.heatmap}, which visualizes the movement of the two solitons within the spatial domain over time. The heatmaps for the three PINN models make it vivid as to how much of an advantage the model $f_i$ has over others.



\subsection{Solving for a 3 Soliton Solution of the KdV}\label{Sec_3Soliton}


For solving for the 3-soliton solution of the KdV PDE we choose as the computational domain, $(x,t) \in [-4,4]\times [-0.5,0.5]$. $18000$ collocation points were randomly chosen in the space-time bulk while $914$ points were sampled from the $t=0$ spatial slice and $457$ points each from the boundaries at $x= \pm 4$. Further, for computing the fractional error w.r.t the true solution another $102400$ points were chosen uniformly from the space-time bulk. By hyperparameter search we choose $q=10^{-4}$ in the $f_{\rm {initial-included}}$. And the empirical risk functions continue to be as defined in the equation \ref{KdV_empirical_risk}.


We recall the 3-soliton solution, $u$, that was given in equation \ref{3_soliton_sol} from which one can reverse engineer the following initial  and boundary conditions, 
\[ u_0(x) = 12\sech^2(x), t=0  \]
\[
g(x, t)=\left\{
    \begin{aligned}
    &u(-4, t), x=-4 \\
    &u(4, t), x=4, 
    \end{aligned}
    \right.
\]
As opposed to our previous KdV experiments, $\sin$ activation functions were used here and the step-length of the Adam optimizer was set to $10^{-3}$, a larger value than earlier. All training was repeated three times to measure variance and the error bars obtained have been reported in the figures below. 

\paragraph{Results}
We compare our results against the data reported for a similar $3-$soliton experiment \citep{hu2022XPINNs}. We employed a neural architecture of depth $5$ and uniform width $32$ and thus used ~$12\%$ less parameters than the size that can be estimate of the report of a similar $3-$soliton experiment \citep{hu2022XPINNs}. For a fair comparison against them in the following table we record the lowest achieved errors by our models till the $5000^{th}-$epoch -- although the figures below show performance improvement well beyond that.

\begin{table}[H]
    \centering
    \begin{tabular}{|c|c|c|c|}
		\hline
            \multicolumn{4}{|c|}{\text { 3-Soliton Solution of KdV (Epoch =5000, Repeat=3, Test Size=102400)}} \\
            \hline
		Model & Sample Size & \#Parameters & Fractional Error \\
		\hline
		model \citep{hu2022XPINNs} & \multirow{4}*{\shortstack{Bulk:18000\\t=0: 914\\x $\in \partial \Omega$: 914}} & 4000 & $6.9 \times 10^{-1}$  \\
            \cline{1-1}
            \cline{3-4}
            vanilla &  & \multirow{3}*{\text 3500} & $6.0 \times 10^{-1}$ \\
            \cline{1-1}
            \cline{4-4}
            boundary-included &  &  & $4.1 \times 10^{-1}$\\
            \cline{1-1}
            \cline{4-4}
            \textcolor{red}{initial-included} &  & & \textcolor{red}{$4.3 \times 10^{-2}$}  \\
            \hline
	\end{tabular}
    \caption{Table of the Results of the 3-Soliton Solution for Different Models}
    \label{tab:KdV3}
\end{table}

Table \ref{tab:KdV3} indicates that our optimal model (highlighted in red) significantly surpasses the former work \citep{hu2022XPINNs} in fractional error relative to the true solution, achieved with a smaller network. This underscores the significance of judicious model/function class selection in training.

\begin{figure}[H]
    \centering  %
    \subfigure[Training Progress With the Different Models]{
        \label{KdV3.sub.1}
        \includegraphics[height=0.5 \textwidth]{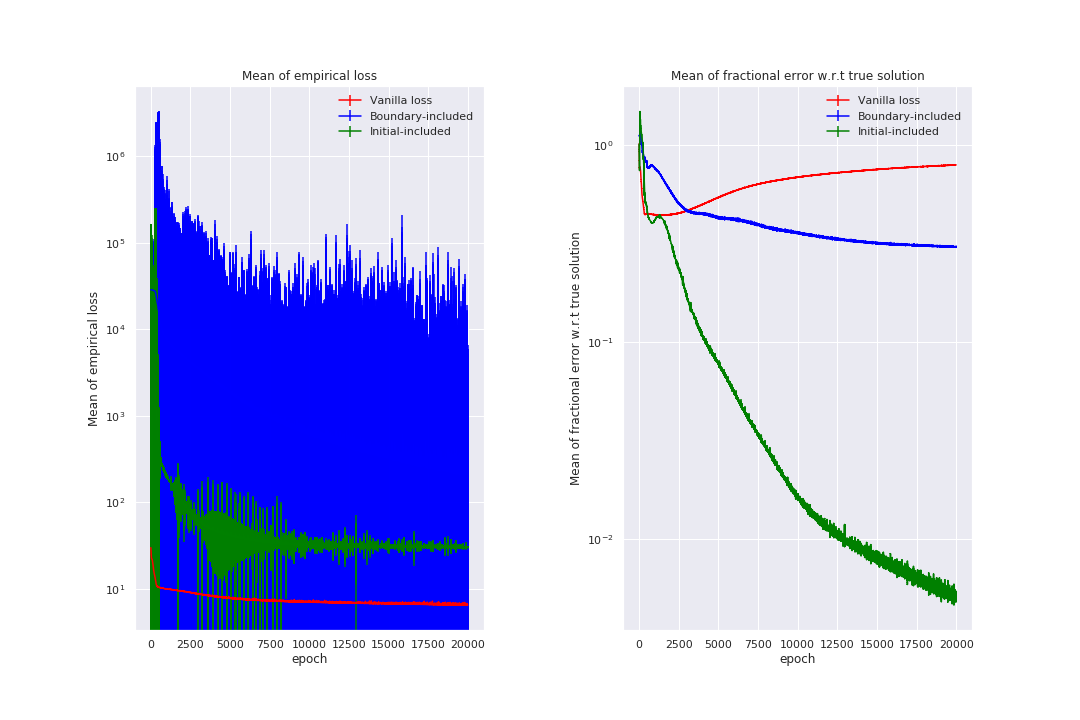}}
    \subfigure[Heatmap Respresentation of the Solutions Obtained by the Different Models Compared Against the True Solution (Bottom-Right)]{
        \label{KdV3.sub.2}
        \includegraphics[width=0.9\textwidth]{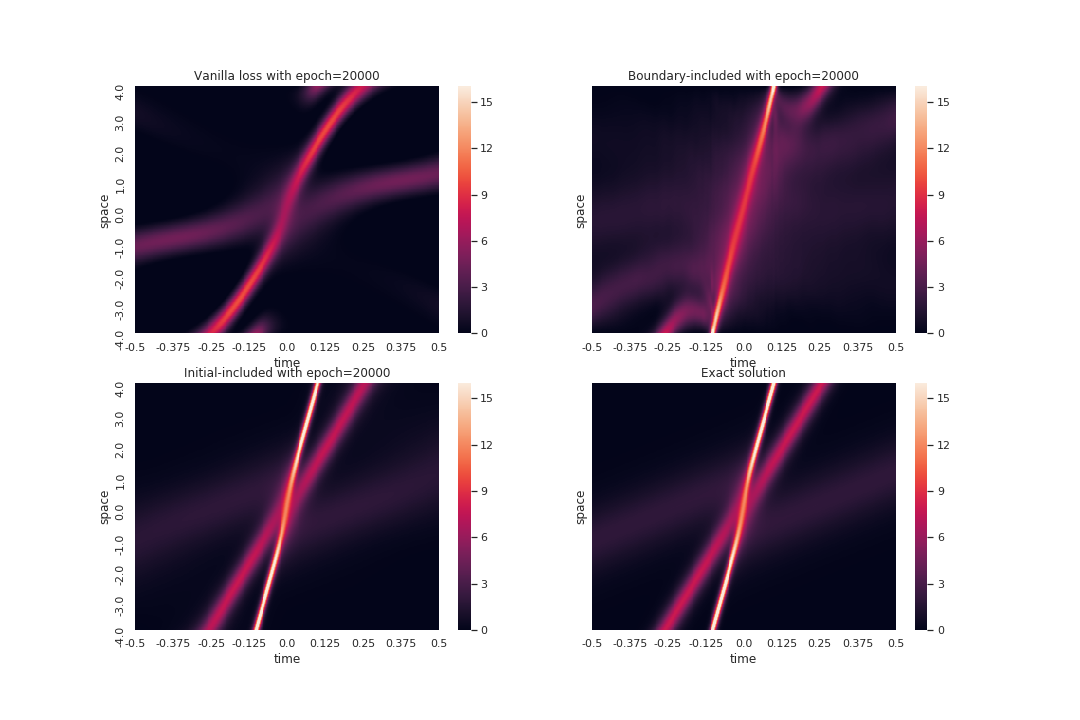}}

    \caption{Results of Solving for the 3-Soliton Solution of KdV}
    \label{Fig.KdV3_models}
\end{figure}

Figure \ref{Fig.KdV3_models}, provides clear visualization of the superiority of the initial-included model ($f_{\rm {initial-included}}$ in equation \ref{KdV_models})'s performance as opposed to the alternatives.